\documentclass[aos]{imsart}

\RequirePackage{amsthm,amsmath,amsfonts,amssymb, algorithm, mathtools}
\RequirePackage[authoryear]{natbib}%% uncomment this for author-year citations
\RequirePackage[colorlinks,citecolor=blue,urlcolor=blue]{hyperref}
\usepackage{comment}
\RequirePackage{graphicx}
\RequirePackage{subcaption}

\startlocaldefs
\theoremstyle{plain}

\newtheorem{theorem}{Theorem}[section]
\newtheorem{lemma}[theorem]{Lemma}
\newtheorem{prop}[theorem]{Proposition}
\newtheorem{corollary}[theorem]{Corollary}
\theoremstyle{definition}

\newtheorem{assumption}[theorem]{Assumption}

\def\bb{\boldsymbol}

\def\Var{\operatorname{Var}}

\endlocaldefs

\begin{document}

\begin{frontmatter}
\title{Transfer Learning with Distance Covariance for Random Forest: Error Bounds and an EHR Application}
%\title{A sample article title with some additional note\thanksref{t1}}
\runtitle{Transfer Learning with Distance Covariance for Random Forest}
%\thankstext{T1}{A sample additional note to the title.}

\begin{aug}
%%%%%%%%%%%%%%%%%%%%%%%%%%%%%%%%%%%%%%%%%%%%%%%
%% Only one address is permitted per author. %%
%% Only division, organization and e-mail is %%
%% included in the address.                  %%
%% Additional information such as            %%
%% identifying the corresponding author must %%
%% be included in in the Acknowledgments     %%
%% section if necessary.                     %%
\author[A]{\fnms{Chenze}~\snm{Li} 
%%\thanks{[\textbf{Corresponding author indication should be put in the Acknowledgment section if necessary.}]}
\ead[label=e1]{li.13074@osu.edu}}
%%\author[B]{\fnms{Second}~\snm{Author}\ead[label=e2]{second@somewhere.com}\orcid{0000-0000-0000-0000}}
\and
\author[A]{\fnms{Subhadeep}~\snm{Paul}\ead[label=e2]{paul.963@osu.edu}}
%%%%%%%%%%%%%%%%%%%%%%%%%%%%%%%%%%%%%%%%%%%%%%
%% Addresses                                %%
%%%%%%%%%%%%%%%%%%%%%%%%%%%%%%%%%%%%%%%%%%%%%%
\address[A]{Department of Statistics,
The Ohio State University %\textbf{[Additional affiliations should be put in the Acknowledgments section]}
\printead[presep={ ,\ }]{e1,e2}}

%%\address[B]{Department, University or Company Name \textbf{[Additional affiliations should be put in the Acknowledgments section]}\printead[presep={,\ }]{e2,e3}}
\end{aug}

\begin{abstract}
 We propose a method for transfer learning in nonparametric regression using a random forest (RF) with distance covariance-based feature weights, assuming the unknown source and target regression functions are sparsely different. Our method obtains residuals from a source domain-trained Centered RF (CRF) in the target domain, then fits another CRF to these residuals with feature splitting probabilities proportional to feature-residual sample distance covariance. We derive an upper bound on the mean square error rate of the procedure as a function of sample sizes and difference dimension, theoretically demonstrating transfer learning benefits in random forests. A major difficulty for transfer learning in random forests is the lack of explicit regularization in the method. Our results explain why shallower trees with preferential selection of features lead to both lower bias and lower variance for fitting a low-dimensional function. We show that in the residual random forest, this implicit regularization is enabled by sample distance covariance. In simulations, we show that the results obtained for the CRFs also hold numerically for the standard RF (SRF) method with data-driven feature split selection. Beyond transfer learning, our results also show the benefit of distance-covariance-based weights on the performance of RF when some features dominate. Our method shows significant gains in predicting the mortality of ICU patients in smaller-bed target hospitals using a large multi-hospital dataset of electronic health records for 200,000 ICU patients.
\end{abstract}

\begin{keyword}[class=MSC]
\kwd[Primary ]{62G08}
\kwd[; secondary ]{62G20}
\kwd{62P10}
\end{keyword}

\begin{keyword}
\kwd{Transfer learning}
\kwd{Random forest}
\kwd{Tree-based methods}
\kwd{Machine learning}
\kwd{Distance covariance}
\end{keyword}

\end{frontmatter}
%%%%%%%%%%%%%%%%%%%%%%%%%%%%%%%%%%%%%%%%%%%%%%
%% Please use \tableofcontents for articles %%
%% with 50 pages and more                   %%
%%%%%%%%%%%%%%%%%%%%%%%%%%%%%%%%%%%%%%%%%%%%%%
%\tableofcontents

\section{Introduction}

Random forest \citep{breiman2001random, biau2012analysis} is one of the most empirically successful machine learning methods. In particular, random forests (RFs) have been shown to outperform almost all classifiers available at the time in large-scale comparative studies for the classification problem \citep{fernandez2014we}. The RF method and related tree-based XGBoost typically outperform modern deep learning in small to moderate-sized structured tabular data \citep{grinsztajn2022tree,kossen2021self}. Further, in a very large-scale empirical evaluation, \cite{mcelfresh2023neural} found that tree-based methods empirically outperform deep learning methods in tabular data when the feature distributions are skewed or heavy-tailed or the data has other types of irregularities.  %Therefore, RF continues to be an important statistical method for modern Machine Learning (ML) and Artificial Intelligence (AI) applications. 
The method is quite suitable for nonparametric regression as it has been both empirically and theoretically shown to be successful in estimating unknown functional relationships between a response and a set of features \citep{biau2012analysis,chi2022asymptotic,scornet2015consistency,klusowski2021sharp,klusowski2024large,wager2018estimation}.

In ML and AI, ``transfer learning'' and ``fine-tuning'' refer to transferring knowledge or model parameters from a vast source dataset to a new target domain problem where we may only have limited data \citep{zhuang2020comprehensive,day2017survey,torrey2010transfer,pan2009survey}. Typically, it is assumed that the domains or problem setups are somewhat related, but aspects of the domains differ. 
The difference in outcome model across the domains is thought to be less ``complex'' compared to the difficulty of learning the models (e.g., they differ in only a small subset of features in a high-dimensional model), justifying the use of ``pre-trained'' source domain models for learning a new target domain model. This circumvents the need for large-scale training data from the new domain, which is typically needed to train complex models \citep{tripuraneni2020theory}. Despite the importance of transfer learning in modern ML, and the success of random forest as a leading ML method, not much research exists for transfer learning with random forest \citep{gu2022transfer,segev2016learn}.

Recently, several authors have theoretically studied the transfer learning problem and obtained statistical error rates. However, most of the results are available only for the linear model. For example, \citet{bastani2021predicting,li2022transfer,takada2020transfer} studied the estimation and prediction error for transfer learning with a high-dimensional penalized regression (Lasso) method under the assumption that outcome model parameters only differ for a few predictors (also called posterior drift). Further results for the linear model with various types of heterogeneity between the source and the target population were obtained in \cite{chang2024heterogeneous,he2024transfusion}. These papers considered differing marginal distributions of covariates (covariate shift) \citep{he2024transfusion}, and heterogeneous transfer learning where feature spaces differ \citep{chang2024heterogeneous} in addition to posterior drift. Beyond the regression problem, transfer learning has been statistically studied in the generalized linear model \citep{tian2022transfer,li2023estimation} and graph estimation with the Gaussian graphical model problem \citep{li2023transfer}. A comprehensive survey on statistical approaches and theoretical results for transfer learning can be found in the recent article by \cite{zhu2025recent}.

However, most existing transfer learning methods for general machine learning with unknown functional relationships do not provide statistical error rates \citep{zhuang2020comprehensive}.  
Some recent works have explored the problem of transfer learning with nonparametric regression using local polynomial regression \citep{cai2024transfer}, nonparametric least squares \citep{schmidt2024local}, and kernel regression \citep{wang2023minimax}. The authors in \cite{cai2024transfer} develop a statistical theory for transfer learning under the assumption that the difference in the functions in the source and target domains is smooth and at most a polynomial function of the features. Assuming only covariate shift and no posterior drift, \cite{schmidt2024local} developed a method and statistical theory for transfer learning with nonparametric least squares.  Despite significant interest and rapid growth in the literature on the transfer learning problem, to our knowledge, no work exists that analyzes the statistical error rates of transfer learning with the random forest method. While  \cite{gu2022transfer,segev2016learn} proposed methods for transfer learning in random forests, neither approach contains statistical theory or guarantees on error rates. A related problem of federated learning with random forests was investigated in \cite{xiang2024transfer}, where the authors, similar to \cite{gu2022transfer}, use the predictions from random forests trained at other sites as additional features for training a new random forest.

There is a wide array of theoretical results known in the literature for various versions of random forest. These results vary by assumptions on the particular random forest version used and asymptotic or nonasymptotic nature of the results (See \cite{biau2016random,scornet2025theory} for comprehensive reviews on theoretical results for random forest). The standard random forest (SRF) method is a complex algorithm due to its many data-driven components, which include different bootstrap subsamples of training data for each tree, a random subset of features at each split in each tree, an empirical criterion-driven choice of feature and location of split in a greedy algorithm, and pruning trees to reduce overfitting. Therefore, a statistical theoretical analysis of the full method is quite difficult, and researchers have focused on theoretically understanding various components and special cases of the method. Nevertheless, these analyses of some versions of random forest have led to remarkable insight into the working of random forest and tree-based methods in general.  In a seminal paper, \cite{biau2012analysis} analyzed and proved nonasymptotic upper bounds on the MSE of a special case of the method called ``centered random forest'' (CRF), which was first introduced by \cite{breiman2004consistency}. The CRF uses the same training data for all trees, the feature to split at each node is selected randomly with predefined probabilities, and each tree is split until a fixed depth. More recently, \cite{klusowski2021sharp} further improved the rate in the MSE bound for this method. For the SRF, \cite{scornet2015consistency} proved consistency of estimating the mean function with an additional assumption of an additive regression model. More recently, \cite{klusowski2024large} showed consistency can be achieved even in the high-dimensional case, when the number of feature grows exponentially as sample size.

We develop a new transfer learning method for nonparametric regression using CRFs aided by distance covariance \citep{szekely2007measuring,li2012feature} and theoretically study the procedure's error rate. Our method first fits a CRF to the source domain data with uniform weights for all features. We predict the response in the target domain using the source-trained random forest and obtain residuals. Now, we divide the target domain data into two parts. In one of the parts, we use the data to calculate sample distance covariances of the residuals with each of the features. Then, using the second part of the data, we fit another centered random forest, treating the residuals as the response, which we call the residual random forest. The vector of probabilities for splitting along each of the features at each split (which we call weights) in this residual random forest is proportional to the distance covariances estimated in the other half of the data. We apply the new methods to improve the prediction accuracy of ICU outcomes for patients in smaller-bed target hospitals using source data from large hospitals with a large multi-hospital EHR dataset of more than 200,000 ICU patients from 335 ICU units in 208 hospitals.

Our theoretical results show that the method is beneficial when the discrepancy between the source and target domains only involves a subset of the features and the source domain has many more samples than the target domain. Suppose $n_t$ and $n_s$ denote the sample sizes in the target and source domains, $d$ denotes the number of features, and $l$ denotes the number of features involved in the domain difference function. With certain assumptions and simplifications, the error rate in our transfer learning procedure decreases as $O(n_t^{-1/l})$, as opposed to the rate for the target domain only CRF being $O(n_t^{-1/d})$ with $l<<d$, provided $n_s >> n_t$. 

Our results also shed light on the mechanism of implicit regularization in the residual random forest in the target domain, enabled by fitting shallower trees with the help of distance covariance. A major difficulty of designing transfer learning in random forests is the lack of explicit regularization that enables the calibration stage of transfer learning in penalized linear regression \citep{bastani2021predicting,li2022transfer,tian2022transfer}, or nonparametric kernel regression \citep{wang2023minimax}. Instead, the regularization in random forests is ``implicit", enabled through limiting the tree depths and decorrelating trees by selecting a subset of features randomly at each split. Shallower trees do not overfit, but are more biased since they are unable to approximate a complex function well, which would generally lead to inferior performance. However, due to a lack of sample size in the target domain, we are forced to fit shallow trees, resulting in low performance of the target data only random forest. In our transfer learning procedure, our first idea is to only fit the \textit{residual random forest} with shallower trees in the target domain. If the true residual function is low-dimensional, then shallow trees with depth $O(\log_2 n_t)$ suffice. The source domain RF has much deeper trees with depth scaling as $O(\log_2 n_s)$. This provides the implicit regularization needed for better error rates. 
However, we may have high variance in the residual RF in the target domain due to not knowing the set of related features and having too many unrelated features. Our second idea is to use distance covariance as a soft feature screening method to preferentially split along certain features in our shallow trees in the residual random forest. A key theoretical result we show is that under the assumption that the true function only depends on the set of features in a set $S^C$, if we can guarantee the feature selection probability vector $\mathbf{p}_n=(p_{n1},\ldots,p_{nd})$ is such that all related features are chosen with probability at least a constant $c>0$ (i.e., $p_{nj} \geq c>0$ for $j \in S^C$), and if for a subset $S_{\alpha} \subset S$ of irrelevant, we can guarantee that $p_{nj} \leq n^{-\alpha}$ for some $\alpha>0$, then we have simultaneously lower squared bias and lower variance. We show that
our sample DC-based splitting leads to those guarantees and consequently have lower variance in addition to lower bias.

To our knowledge, this is the first procedure for transfer learning with random forest that has statistical error rate guarantees. Moreover, while related work such as \cite{cai2024transfer} restricted the difference function to be polynomial, we do not restrict the function class of the difference function beyond Lipschitz-type requirements. Transfer learning in other nonparametric regression methods, such as kernel regression in \cite{wang2023minimax}, also fundamentally differs from our method and theoretical results, since in kernel regression, one can explicitly regularize by the norm in the RKHS space similar to $\ell_1$ regularization in transfer learning for linear models as in \cite{bastani2021predicting,li2022transfer}.

Our use of distance covariance learnt in a split sample for determining the vector of probabilities for splitting the features may also be of interest for the random forest procedure in general. We theoretically show that for the features independent of the difference function between source and target domains, the population distance covariances with the residual random variable in the target data are small. We further show that their sample counterparts well approximates the population measures with high probability. Therefore, using the sample distance covariances as splitting probabilities for the features amounts to a form of soft feature screening, which enables a faster rate of convergence in the mean squared error of the residual random forest on the target data. This procedure, therefore, answers a question raised by \cite{klusowski2021sharp} on how to estimate the set of ``active'' features in CRF and presents an alternative procedure to the decision stumps procedure in \cite{klusowski2021nonparametric}.

\section{Transfer learning with random forest}
We consider the general nonparametric regression problem with an unknown functional relationship between the response and the predictors. Suppose we have $d$ features or predictors $X_1, \ldots, X_d$ each taking values in the set $[0,1]$, which are available in both the source and target domains. We use the superscript ``(s)'' to denote the source domain and the superscript ``(t)'' to denote the target domain. Accordingly, in the source domain, we observe $\bb X_{i}^{(s)} = \{X_{i1}^{(s)}, \ldots, X_{id}^{(s)}\} \in [0,1]^d$ and corresponding response $Y_i^{(s)} \in \mathbb{R}$ for $i = 1, \ldots, n_s$. In the target domain, we observe $\bb X_{i}^{(t)} = \{X_{i1}^{(t)}, \ldots, X_{id}^{(t)}\} \in [0,1]^d$ and corresponding response $Y_i^{(t)} \in \mathbb{R}$ for $i=1, \ldots, n_t$. We assume the data generating models in the source and target domains are
    \[
    Y_{i}^{(s)} = f_s(\bb X_{i}^{(s)}) + \epsilon_{i}^{(s)}, \quad     Y_{i}^{(t)} = f_t(\bb X_{i}^{(t)}) + \epsilon_{i}^{(t)},
    \]
    with $\epsilon_{i}^{(s)}, \epsilon_{i}^{(t)}$ are i.i.d. random variables. Both $f_s$ and $f_t$ are unknown regression functions. Therefore, while we assume the response in both domains is a function of the same set of features or covariates, the functional relationships differ between the domains. We assume $n_s >>n_t$ i.e., the sample available in the source domain is vastly higher than the sample available in the target domain. Let $D_s$ and $D_t$ denote the training data in the source and the target domain, respectively.
    We make the following assumptions on the predictors $\bb X_i$, the error terms $\epsilon_i$ and the functions $f_s, f_t$. We use the notation $\|f\|_{\infty}$ to denote $\sup_x |f(x)|$ over the domain of $f$. These assumptions are modification of the ones made in \cite{biau2012analysis,klusowski2021sharp}.

\begin{assumption}
    In both source and target domains, we assume $\mathbf{X}$ is uniformly distributed on $[0,1]^d$. The random errors $\{\varepsilon_i^{(s)}\}_{1 \leq i \leq n_s}$ and $\{\varepsilon_i^{(t)}\}_{1 \leq i \leq n_t}$ are i.i.d. Sub-Gaussian random variables and $\mathrm{Var}(Y_i^{(s)} \mid \mathbf{X}_i^{(s)}) = \sigma_s^2, \mathrm{Var}(Y_i^{(t)} \mid \mathbf{X}_i^{(t)}) = \sigma^2_t$, for some positive constants $\sigma^2_s, \sigma^2_t$.
    \label{assmp1}
\end{assumption}

\begin{assumption}
    The regression functions $f_s(\cdot), f_t(\cdot)$ are bounded and L-Lipschitz  functions on the domain $[0,1]^d$, i.e., \[ \| f_s\|_\infty \leq M_s, \quad \| f_t\|_\infty \leq M_t \quad \max(\max_j\|\partial_j f_s\|_\infty,\max_j\|\partial_j f_t\|_\infty) < L,\]
    for constants $M_s, M_t, L$. Further the difference function $R(\cdot):=f_t(\cdot)- f_s(\cdot)$ is L-Lipschitz function  on $[0,1]^l$ involving $l$ features and is independent of the remaining $d-l$ features, where $l << d$. Concretely, let $\mathbf{X}_{.,k}$ denote the $k$th feature, and then we define,
\[
\mathcal{I}_R = \left\{k: \mathbf{X}_{.,k}  \text{ is independent of  } R( \mathbf{X} ), \, k=1,\ldots,  d\right\}.
\]
Then $|\mathcal{I}^C_R|=l<<d$. 
\label{assmp2}
\end{assumption}

Our goal in this paper is to develop methodology to estimate the function $f_t$ in the target domain. However, given the limitation of data in the target domain, it might be beneficial to transfer knowledge from the source domain with vast amount of data provided the functions $f_s$ and $f_t$ are somewhat related. The Assumption \ref{assmp2} above relates the functions $f_t$ and $f_s$ and posits that the difference between the functions is a low-dimensional function, i.e., a function of a small number of features. This assumption can be related and contrasted with the assumption for transfer learning in nonparametric regression problems in \cite{cai2024transfer}, who assumed the difference $f_s-f_t$ is a polynomial function (of all features). In contrast, we let the difference function be an arbitrary function of $l$ out of $d$ features with $l<<d$ and is a ``sparse'' function in that sense.  Note that our transfer learning setup is also different from the feature representation-based transfer learning paradigm considered in \cite{tripuraneni2020theory}, where the source and target ``tasks'' are assumed to share a low-dimensional representation function $h(x)$, even though the tasks may differ. Instead, we consider the same or a related task, but do not assume the existence of any shared representation of the features.

\subsection{Review of random forest and centered random forest}
In this section, we review the random forest and associated methods following the review articles of \cite{scornet2025theory,biau2016random}.
The random forest is an ensemble method that predicts a response by aggregating predictions from several trees, each of which is grown with some randomization, for example, based on a bootstrapped version of the training data or by selecting a random subset of features as candidate features at each split.
Using the notations in \cite{klusowski2021sharp,scornet2025theory}, in any domain suppose we observe dataset $D$ that consists of $n$ i.i.d pairs of observations ($\mathbf{X}_i,Y_i$) with  $\mathbf{X}_i \in \mathbb{R}^d$ and $Y_i \in \mathbb{R}$. We let $T(\mathbf{X},\Theta_m,D)$ denote the prediction at a new observation $\mathbf{X}$ of a decision tree trained on data $D$ and randomization parameter $\Theta_m$ that controls random decisions in the tree formation. Mathematically, $T(\mathbf{X},\Theta_m,D)$ is the average of all observations in the training sample $D$ that fall in the same random partition as $X$. Let $t(\mathbf{X}, \Theta_m,D)$ denote the random partition in the tree that contains $\mathbf{X}$. Then
\[
T(\mathbf{X},\Theta_m,D) = \frac{\sum_{i=1}^n Y_i \mathbf{1}(\mathbf{X}_i \in t)}{\sum_{i=1}^n \mathbf{1} (\mathbf{X}_i \in t)} \mathbf{1} \left(\sum_{i=1}^n \mathbf{1} (\mathbf{X}_i \in t) \neq 0\right).
\]
A random forest consists of an average of $M$ randomized decision trees
\[\hat{Y}(\mathbf{X},\Theta, D) = \frac{1}{M} \sum_{m=1}^M T(\mathbf{X}, \Theta_m, D),
\]
where the parameters $\Theta_m$ that govern the randomization mechanism are realizations (sample) of a random variable $\Theta$. It is customary in the literature to consider the ``infinite'' or ``expected'' random forest, which is given by \[
\hat{Y}(\mathbf{X}, D):= \mathbb{E}_\Theta\left[\hat{Y}(\mathbf{X}, \Theta, D)\right]\]
where the expectation is with respect to the random variable $\Theta$, and is conditional on $\mathbf{X}$ and
$D$ \citep{klusowski2021sharp}. This is justified by considering the case when $M \to \infty$, that is, we consider an ensemble of a large number of trees \citep{scornet2016asymptotics}. 

For our theoretical development, we work with the centered random forest (CRF), where the individual trees are formed purely randomly and not in a data-driven way.  In a centered random forest, each tree is grown with the following two steps \citep{klusowski2021sharp}: (1)  after initializing the root as $[0,1]^d$, at each node we select one feature $j$ out of $d$ features with \textit{pre-decided} probabilities of selection given by the sequence $\mathbf{p}=\{p_1,\ldots, p_d\}$, (2) then we split the node exactly at the \textit{midpoint} of the feature selected in the previous step. We repeat these two steps a \textit{fixed} $[\log _2 k_n]$ times. This gives us a tree with $k_n$ leaves. Accordingly, the centered random forest object is completely characterized by the training data $D$ and the sequence of weights for each feature $\mathbf{p}$. Intuitively, even though both the direction (feature to split) and the point of splits are random, the \textit{predictive power} of centered random forest comes from the fact that at each leaf node we take the average of training data points which are ``similar'' to the test data points $\mathbf{X}$. The probability of the selection vector $\mathbf{p}$ is an important part of our transfer learning strategy, as we can control how the feature space is partitioned using this vector.

\subsection{Proposed transfer learning method}
With the assumptions made in the previous section, we propose the following three-stage transfer learning method to estimate the function $f_t$, using the target data $D_t$ and transferring appropriate information from the source data $D_s$.

\begin{longlist}
    \item \textbf{Step 1 (source domain):} train a centered random forest using the training sample from the source domain $D_s:= (Y^{(s)}, \mathbf{X}^{(s)})$. Let us denote $\hat{Y}(\mathbf{X},D_s)$ as the random forest obtained in the source domain.
    \item \textbf{Step 2 (residualize in the target domain):} In the target domain we obtain the prediction $\hat{Y}_{i}^{(s,t)} = \hat{Y}(\mathbf{X}_{i}^{(t)},D_s)$ using the source model on target feature vectors. Then we obtain the residual, $\tilde{Y}_{i}^{(t)} = Y_{i}^{(t)}- \hat{Y}_{i}^{(s,t)}$.
    \item \textbf{Step 3 (fit the calibrated model in target domain):} We fit another random forest model in the target domain on these residuals. Call the modified training sample in the target domain as $\tilde{D}_t = (\tilde{Y}_t,\mathbf{X}_t)$. Split the data into two parts, $\tilde{D}_t'$ and $\tilde{D}_t''$. The centered random forest is fitted to the data $\tilde{D}_t'$ with feature weights for predictor $j$ proportional to distance covariance between $\tilde{Y}_t$ and $\mathbf{X}_j$.  These sample distance covariances are estimated from the second split of the data $\tilde{D}_t''$. Call this random forest $\hat{Y}( \mathbf{X},\tilde{D}_t')$. 
 \item \textbf{Step 4 (add to obtain final predictions)} Obtain the final predictions as 
    \[\hat{Y}(\mathbf{X},(D_s,\tilde{D}_t)) = \hat{Y}(\mathbf{X},D_s) + \hat{Y}(\mathbf{X},\tilde{D}_t').
    \]
\end{longlist}

The method is summarized in Algorithm \ref{alg:transfer-rf}. For estimating the sample distance covariance, we use the computationally fast implementation of the unbiased sample covariance estimator from \cite{huo2016fast}. In particular the probabilities of selecting the features is given by $
      p_j \;=\;\frac{\widehat{\mathrm{dCov}}\!\bigl(\mathbf{X}_{.,j},Y\bigr)}
                   { \sum_{k=1}^d \widehat{\mathrm{dCov}}\!\bigl(\mathbf{X}_{.,k},Y\bigr)}$, where $\widehat{\mathrm{dCov}}$ is the sample distance covariance. We remind the reader that distance covariance is always non-negative. We assume the sum of those in the denominator is positive so that the ratios are well-defined and are non-negative. This is justified since if all the sample distance covariances $p_j$s are 0s, then we will not fit the residual random forest anymore.

\begin{algorithm}[h]
\caption{Centered Random Forest with Distance Covariance (CRF-DCOV)}
\label{alg:centered-rf}
\begin{longlist}
  \item \textbf{Initialize.}  Start the root node with $[0,1]^d$.
  \item \textbf{Feature sampling.}  At each node, select feature $j\in\{1,\dots,d\}$ with probability
    \[
      p_j \;=\;\frac{\widehat{\mathrm{dCov}}\!\bigl(\mathbf{X}_{.,j},Y\bigr)}
                   { \sum_{k=1}^d \widehat{\mathrm{dCov}}\!\bigl(\mathbf{X}_{.,k},Y\bigr)},
    \]
    where $\widehat{\mathrm{dCov}}$ is the sample distance covariance obtained from an independent copy of the sample.
  \item \textbf{Split.}  { Split the selected feature at the midpoint.}
  \item \textbf{Recurse.}  {Repeat steps (ii)–(iii) for exactly $\bigl\lceil \log_2 k_n \bigr\rceil$ times.}
\end{longlist}
\end{algorithm}

\begin{algorithm}[h]
\caption{Three‐Stage Transfer Learning via Centered RF (TLCRF)}
\label{alg:transfer-rf}
\begin{longlist}
  \item \textbf{Source‐domain modeling.}\quad 
    Given source data 
    $\displaystyle D_s= (Y^{(s)}, \mathbf{X}^{(s)})$, train 
    a CRF $\hat{Y}(\mathbf{X};D_s)$ using uniform weights for all features.
  \item \textbf{Residualization in target.}\quad 
    For each target point $\mathbf{X}_{t}^{(t)}$, compute the source prediction
    \[
      \hat{Y}_{i}^{(s,t)} = \hat{Y}(\mathbf{X}_{i}^{(t)},D_s),
    \]
    then form residuals 
    \[
    \tilde{Y}_{i}^{(t)} = Y_{i}^{(t)}- \hat{Y}_{i}^{(s,t)}.
    \]
  \item \textbf{Target‐domain calibration.}\quad 
    Let  $\tilde{D}_t = (\tilde{Y}_t,\mathbf{X}^{(t)})$. Divide the data into two parts $\widetilde{D}_t',\widetilde{D}_t''$.
    Train a second CRF $\hat{Y}(\mathbf{X},\widetilde{D}_t'),$ using Algorithm~\ref{alg:centered-rf}, with weights proportional to (distance covariance between residual and feature)
    \(\widehat{\mathrm{dCov}}(\mathbf{X}_{.,j},\tilde{Y})\) obtained from $\tilde{D}_t''$. Then produce the final prediction as,
    \[\hat{Y}(\mathbf{X},(D_s,\tilde{D}_t)) = \hat{Y}(\mathbf{X},D_s) + \hat{Y}(\mathbf{X},\tilde{D}_t').
    \]
\end{longlist}
\end{algorithm}

\section{Theoretical results}

In this section, our goal is to obtain a non-asymptotic upper bound for
the Mean Squared Error (MSE) in the target domain, which is, \(
\textstyle \mathbb{E}\big[(\hat{Y}(\mathbf{X},(D_s,\tilde{D}_t)) - f_t(\mathbf{X}))^2\big]
\). In doing so, we also clarify why the methodology works. Since our algorithm fits two centered random forest models, first to the response in the source data (source CRF) and then to the residuals in the target data (residual CRF), the overall MSE is controlled by the MSE errors of these two models. 
%Our results, therefore, analyze the MSE of these two models separately. 
While the source CRF uses uniform weights for the features as in \cite{klusowski2021sharp}, the residual CRF in the target domain requires special treatment since the weights the features receive $\bb p_t$ are determined through sample distance covariance computed in an independent sample through sample splitting $(D_t'')$. 

Note that since the predictions $\hat{Y}_{i}^{(s,t)}$ are based on the random forest trained in a different dataset (source data), they remain i.i.d across $i$ given our assumption of i.i.d feature vectors $\mathbf{X}_i^{(t)}$. Consequently, the residuals, $\tilde{Y}_{i}^{(t)}$ also satisfy the assumption of being i.i.d  across observations.

The following proposition shows the MSE error of the first-step model in the source domain. Since we run the usual centered random forest in the source domain, the result in \cite{klusowski2021sharp} is applicable to obtain an upper bound on the error rate in the source domain.

\begin{prop}[Centered random forests, \cite{klusowski2021sharp}]
\label{klusowski}
Assume a CRF is fit to the source data with uniform probabilities of selection, $p_j^{(s)}=\frac{1}{d},\ j=1,\cdots, d$. Under Assumption \ref{assmp1} and Assumption \ref{assmp2},
\begin{align*}
\mathbb{E}\bigl[(\hat{Y}(\mathbf{X}, D_s) - f_s(\mathbf{X}))^2 \bigr]
& \le
d \sum_{j=1}^d \|\partial_j f_s\|_{\infty}^2 \, k_{n_s}^{2 \log_2 (1 - 1/2d)}\\
& +
\frac{12 \sigma_s^2 k_{n_s}}{n_s}
\frac{(8d^{1/2})^d}{\sqrt{\log_2^{\,d - 1}(k_{n_s}) }}
+
M_s^2 \, e^{ - n_s / (2 k_{n_s} ) }.
\end{align*}

\noindent Further, if we define $z_s := \frac{2 \log_2(1 - 1/2d)}{2 \log_2(1 - 1/2d) - 1}$, and choose $
k_{n_s} = c \bigl( n_s \bigl( \log_2^{\,d - 1} n_s \bigr)^{1/2} \bigr)^{1 - z_s}$, 
with some constant $c > 0$ independent of $n_s$, then, given $(p_j^{(s)})_{1 \le j \le d}$, there exists a constant $C > 0$ not depending on $n_s$, such that
\[
\mathbb{E}\bigl[(\hat{Y}(\mathbf{X}, D_s) - f_s(\mathbf{X}))^2 \bigr]
\le
C \Bigl( n_s \bigl( \log_2^{\,d - 1} n_s \bigr)^{1/2} \Bigr)^{-z_s}.
\]
\end{prop}

Therefore, the central analysis in this paper focuses on the residual CRF. To understand how the residual CRF effectively reduces both the squared bias and variance, we first establish a general result on random forest if the function $f(\mathbf{X})$ only depends on a subset of the features (i.e., independent of the other ``unrelated'' features), and we have a good estimate of what those features that $f(\mathbf{X})$ depends on are. The result has general applicability beyond transfer learning and justifies various feature screening methods that are often applied prior to running the random forest method. In our method, we propose using distance covariance to estimate the set of features on which $f(\mathbf{X})$ depends.

We start with the following technical lemma, which is a modification of a lemma in \cite{klusowski2021sharp}.
\begin{lemma}
\label{lemma1}
Let $\mathbf{X}=(X_1,\dots,X_k)$ be a random vector that follows a multinomial distribution with $m$ trials and parameters $(p_{m1},\dots,p_{mk})$, with none of them being exactly 0. Suppose that $mp_{mj}\to 0, \forall j=k_1+1,\cdots, k$ when $m\to\infty$. Let $\mathbf{X}'=(X'_1,\dots,X'_k)$ be an independent copy of $\mathbf{X}$. Then, 
\begin{equation}
\mathbb{E}\Bigl[2^{-\frac12\sum_{j=1}^{k}\lvert X_j - X'_j\rvert}\Bigr]
< \frac{8^k}{\sqrt{m^{k_1-1}\,p_{m1}\cdots p_{mk_1}}}.
\end{equation}
\label{lemmamult}
\end{lemma}
The main difference of this lemma from that of \cite{klusowski2021sharp} is that we show when $mp_{mj} \to 0$ for some classes $j$, the denominator in the upper bound in Lemma (\ref{lemmamult}) only involves the product over the other class probabilities and not on those $p_{mj}$s.

The next result is one of our key theorems, which provides a bound on the MSE of the error rate if the set of features that the response depends on is approximately correctly selected. We assume the true function $f$ that relates $Y$ with $\mathbf{X}$ is low-dimensional and depends only on the features in a set $S^C$ (and therefore does not depend on the features in the complementary set $S$). Suppose we can guarantee that the weights we put on the features which are truly related to the response (denoted as the set $S^C$) is bounded below by a constant $c>0$ while we have a subset $S_{\alpha} \subset S$ such that the weights on features in $S_{\alpha}$ are bounded above by $n^{-\alpha}$ for some constant $\alpha>0$. Then the next theorem shows the MSE of a CRF fit with the sequence of weights with this property has simultaneously lower squared bias and lower variance.

\begin{theorem}
\label{rate2}
Assume a dataset $D = \left\{\mathbf{X}_i, Y_i\right\}_{i=1}^n$ where $Y_i = f(\mathbf{X}_i)+\epsilon_i$. Assume $\mathbf{X}_i$ is uniformly distributed on $[0,1]^d$ and, for all $\mathbf{x}\in\mathbb{R}^d$,
$
\sigma^2(\mathbf{x})
= \Var\bigl(Y_i \mid \mathbf{X}_i =  \mathbf{x}\bigr)
\;\equiv\;\sigma^2$,
for some positive constant~$\sigma^2$. We run the CRF method on the dataset $D$ with probabilities of features $(p_{nj})_{1\leq j \leq d}$ and depth of trees being  $\lceil \log_2 k_{n} \rceil$. Suppose $f(\cdot)$ is an $l = \left| \mathcal{S}^C\right|$ dimensional $L$-Lipschitz function bounded by a positive constant $M$ and depends only on the features in the set $\mathcal{S}^C$. Assume $p_{nj} \geq c > 0 ,\forall j\in \mathcal{S}^C$ and denote the set $\mathcal{S}_\alpha=\left\{j: p_{nj}\leq n^{-\alpha}\right\}$ for some $\alpha>0$, such that $S_{\alpha} \subset S$. Let
 $\hat{f}_Y(\mathbf{X}; D)$ is the prediction of $f(\mathbf{X})$ with input $\mathbf{X}$ and $\bar{f}(\mathbf{X}, D) := \mathbb{E}[\hat{f}_Y(\mathbf{X}, D) \mid \mathbf{X}_{1}, \ldots, \mathbf{X}_{n}, \mathbf{X}]
$. 
When $n$ is large enough, 
\begin{align*}
\mathbb{E}\left[(\hat{f}_{Y}(\mathbf{X}, D) - f(\mathbf{X}))^2\right]\leq &|\mathcal{S}^C| \sum_{j \in \mathcal{S}^C} \left\| \partial_j f \right\|_{\infty}^2 k_n^{2 \log_2 (1 - p_{nj} / 2)} + M^2 e^{-n/(2 k_n)} \\
\qquad &+ \frac{12\sigma^2 k_{n}\,8^{d}}{n \sqrt{\displaystyle\prod_{j\in\mathcal{S}_{\alpha}^C}p_{nj}\;\bigl(\log_2 k_{n}\bigr)^{|\mathcal{S}_{\alpha}^C|-1}}}
\end{align*}
\end{theorem}
Note that in the above upper bound the first two quantities correspond to the squared bias, $\mathbb{E}[\bar{f}(\mathbf{X},D)-f(\mathbf{X}))^2]$, and the third quantity corresponds to the variance, $\mathbb{E}[(\hat{f}_Y(\mathbf{X},D)-\bar{f}(\mathbf{X},D))^2]$. Also note that the variance term involves $S_{\alpha}^C$ while the bias term involves $S^C$. Examining the proof, we note that the squared bias nicely adapts to the sparsity of the function, with the sum involving only features in $S^C$ and the scaling factor in front being $|S^C|$, irrespective of the assumption of $p_{nj} \geq c$ on $S^C$. The assumption $p_{nj} \geq c>0$ for all $j$ in the set $S^C$ ensures that the exponent of $k_n$ (which we remind the readers is typically of $O(n)$) remains negative for all $j \in S^C$. Therefore, the consequence of this assumption not holding is a higher rate in the squared bias. With the assumptions placed on $p_{nj}$'s for $j \in S^C$, the squared bias is lower than what we would have if we were fitting a non-sparse function. The third expression of the bound corresponds to the variance term which is bound using Lemma \ref{lemmamult}. This bound is also lower than what we would obtain if we did not have $p_{nj}\log_2 k_n \to 0$ for all $j \in S_{\alpha}$. This can be seen by noting that each component in the product in the denominator is a fraction, and having more terms in the product only decreases the denominator making the variance higher.

The above theorem shows that if one can obtain the weights on the features that guarantee non-vanishing probabilities on the set of related features $S^C$ and vanishing probabilities on at least a subset of unrelated features $S_{\alpha}$, then the upper bound's dependence on the number of features becomes favorable. Since we expect the true difference function $R(\mathbf{X})$ to be a ``low-dimensional'' function, the residual function is also approximately low-dimensional (we show this formally in a lemma below). Therefore,  for fitting the residual CRF in the target domain, if we can choose the weights so that the weights for features that are truly related to the residual function are high and the weights for the unrelated features are low, satisfying the conditions of the theorem, we will have a favorable upper bound. However, a challenge here is that we do not know which features are truly related to the residual function. Moreover, the residuals we obtain in the target domain are sample versions or realizations of the true difference function and hence at best are noisy versions of the difference function.

The following key lemma provides a first justification of our use of distance covariance between the residual random variable and features for weighing relevant features for modeling the residuals (also called debiasing or calibration in the literature on transfer learning). The lemma shows that if some features $\mathbf{X}_{.,j}$ are independent of the \textit{true} difference function $R(\mathbf{X})$, then the \textit{population} distance covariance $\mathcal{V}^2$ between the residual random variable obtained in the target domain using the fitted CRF from source domain and those features is small.
\begin{lemma}
\label{dcov0}
    Let $\tilde{Y}(\mathbf{X}^{(t)}, D_s)$ denote the residual random variable in the target domain obtained with predictions from the CRF trained in the source data. If $\mathbf{X}_{.j}$ which is the  $j$-th dimension of $\mathbf{X}$ is independent of the difference function $R(\mathbf{X})=f_t(\mathbf{X})-f_s(\mathbf{X})$, i.e., $\mathbf{X}_{.j}\in \mathcal{I}_R$, then we have the following bound on the population distance covariance:
        \begin{align*}  \mathcal{V}^2\left(\tilde{Y}(\mathbf{X}^{(t)}, D_s), \mathbf{X}_{.j}\right)\leq \tilde{C}_s\left(n_s(\log_2^{d-1}n_s)^{\frac{1}{2}}\right)^{-z_s/2},
\end{align*}
for some constant $\tilde{C}_s$.
\end{lemma}

Note that we can decompose the residual from the prediction of source trained CRF in the target domain for a generic data point $(Y^{(t)},\mathbf{X}^{(t)})$ as,  
\begin{align*}
    \tilde{Y}(\mathbf{X}^{(t)}, D_s) & = Y_t - \hat{Y}(\mathbf{X}^{(t)},D_s) \\
    & =R(\mathbf{X}^{(t)}) +\epsilon^{(t)}+ f_s(\mathbf{X}^{(t)}) - \hat{Y}(\mathbf{X}^{(t)}, D_s) \\
    & = R(\mathbf{X}^{(t)}) +\epsilon^{(t)} + \delta_s(\mathbf{X}^{(t)}), 
\end{align*}
where $\delta_s(\mathbf{X}^{(t)}) = f_s(\mathbf{X}^{(t)}) - \hat{Y}(\mathbf{X}^{(t)}, D_s)$ is the error in approximating the true function we would have if the data point $(Y^{(t)},\mathbf{X}^{(t)})$ was from the source domain. Therefore, the residual random variable is the difference function contaminated with two noise terms. 
 In the proof in Appendix, we show that $\mathcal{V}^2\left(\tilde{Y}(\mathbf{X}^{(t)}, D_s), \mathbf{X}^{(t)}_{.,j}\right)$ is upper bounded by a constant multiple of $\sqrt{E[(\delta_s(\mathbf{X}^{(t)}))^2]}$ when $X_{.,j} \in \mathcal{I}_R$, and then applying Proposition \ref{klusowski} we obtain the bound in the lemma.

However this result was about the \textit{population} distance covariance between the residual random variable and the feature $X_{.,j}$. The Proposition 1.2 in the Supplement, which is an adaptation of the main result in \cite{li2012feature}, shows that the sample distance covariance approximates the population distance covariance well.  Note that since $\mathbf{X}$ follows a uniform distribution and $f(\cdot)$ is a bounded function, $f(\mathbf{X})$ is bounded. It is easy to see that any bounded random variable is a sub-Gaussian random variable, and the sum of two independent sub-Gaussian random variables is also a sub-Gaussian random variable. Therefore, $Y$ is also a sub-Gaussian random variable. Hence, the conditions in \cite{li2012feature} are satisfied and the main results are applicable in our context.

The next two theorems show that the sample distance covariances between the residuals and features can be effectively used as the weights. 
For brevity of notation, we denote the population distance covariance between residuals in the target domain $\tilde{Y}(\mathbf{X}^{(t)}, D_s)$, and the $k$th feature $\mathbf{X}_{.,k}^{(t)}$ as $\omega_k= \mathcal{V}^2\left(\tilde{Y}(\mathbf{X}^{(t)}, D_s),\mathbf{X}_{.,k}^{(t)} \right)$. Further, let $\hat{\omega}_k$ denote its sample counterpart obtained as $\hat{\omega}_k =\hat{\mathcal{V}}^2\left(\tilde{Y}(\mathbf{X}^{(t)}, D_s), \mathbf{X}_{.,k}^{(t)}\right) $. Further recall, $\mathcal{I}_R$ denotes the set of features $\mathbf{X}$ that are independent of the function $R(\mathbf{X})$. Our Assumption \ref{assmp2} on sparse differences imply, $\left|\mathcal{I}_R\right|>>\left|\mathcal{I}_R^c\right|$, which means that a majority of features are independent with $R(\mathbf{X})$. We make a new assumption here on $\omega_k$s for the features in $\mathcal{I}_R^C$.

\begin{assumption}
    Define $\omega_k= \mathcal{V}^2\left(\tilde{Y}(\mathbf{X}^{(t)}, D_s),\mathbf{X}_{.,k}^{(t)} \right)$. Then, $\min_{k \in \mathcal{I}^C_R} \omega_k \geq cn_t^{-\alpha}$ for some $c>0$ and $0\leq \alpha<1/2$.
    \label{minomega}
\end{assumption}
This assumption is mild and states that the population distance covariance between the residual vector and the features which are related to it is bounded away from 0.

\begin{theorem}
\label{relateddcov}
Define $\hat{\Omega} = \sum_{k=1}^{d}\hat{\omega}_k, \ \Omega = \sum_{k=1}^{d}\omega_k$, and   $\omega_{max}:=\max_{j\in \mathcal{I}_R^C}\omega_j$ and assume all of them are positive numbers. Assume $|\mathcal{I}_R^c|<|\mathcal{I}_R|$ and assume the Assumption \ref{minomega}.  Then, for any $0 < \eta < 1/2-\alpha$, there exist positive constants $c_1 > 0$ and $c_2 > 0$ such that
\begin{align*}
\mathbb{P}\left(\max_{j\in \mathcal{I}_R^c}\left| \frac{\hat{\omega}_j}{\hat{\Omega}} - \frac{\omega_j}{\Omega} \right| \geq cn_t^{-\alpha}\right) &  \leq O\left(|\mathcal{I}_R|\left[\exp(-c_1\left(\tfrac{\Omega}{|\mathcal{I}_R|}\right)^2n_t^{1-2(\eta + \alpha)}+n_t \exp(-c_2 n_t^{\eta})\right]\right). 
\end{align*}

\end{theorem}

The above theorem shows that for the features which are related to (dependent on) $R(\mathbf{X})$, the ratio of sample distance covariance to the sum of sample distance covariances well approximates the ratio of population distance covariance to the sum of the population distance covariances.  
Informally, this result justifies our use of the ratio of the sample distance covariance to the sum of the sample distance covariances as feature weights in our random forest method.

\begin{theorem}
\label{targetprob} 
    Assume the Assumption \ref{minomega}, and  $\sum_{j=1 }^d\omega_j = \Omega \geq M>0$, for some constant $M$ that does not depend on $n_t$, and define $\omega_{max}^R:=\max_{j\in \mathcal{I}_R}\omega_j$.
Then for any $0 < \eta<1/2-\alpha$, and \(\tilde{c}n_t^{-\tilde{\alpha}} = c n_t^{-\alpha} + \left(\tilde{C}_s/\Omega\right)\big(n_s(\log_2^{d-1} n_s)^{\tfrac12}\big)^{-z_s/2}\) , there exists constants $c_1, c_2$, s.t.
\begin{align*}
\mathbb{P}\left(\max_{j \in \mathcal{I}_R} \frac{\hat{\omega}_j}{\hat{\Omega}}\leq \tilde{c}n_t^{-\tilde{\alpha}}\right)\geq 1-O\left(|\mathcal{I}_R| \left[\exp\left(-c_1 \left(\tfrac{\Omega}{|\mathcal{I}_R|}\right)^2 n_t^{1-2(\alpha+\eta)}\right) + n_t \exp(-c_2 n_t^{\eta})\right]\right)
\end{align*}
\end{theorem}

The above theorem shows that with high probability, the weights for the unrelated features $p_{nj}^t$ are bounded by a constant multiple of $n_t^{-\tilde{\alpha}}$. We have also seen in Theorem \ref{relateddcov} that for the related covariates, the sample covariance ratios well approximate the population covariance ratio. Therefore, these two results, along with Assumption \ref{minomega} guarantees the conditions on $p_{nj}$ over the sets $S^C$ and $S_{\alpha}$ needed in Theorem \ref{rate2}.

We are now ready to state our \textit{main results of the paper} (presented over Theorem \ref{main}, Corollary \ref{cormain}, and the remarks after that),  which provide a nonasymptotic upper bound for our transfer learning with the CRF method.

\begin{theorem}
\label{main}
    If $f_s(\cdot). f_t(\cdot)$ and $R(\cdot)$ satisfies assumptions \ref{assmp1} and \ref{assmp2}, and
    $$p_{nj}^{(s)} \;=\;\frac{1}{d},\quad p_{nj}^{(t)} \;=\;\frac{\hat{\omega}_j^t}{\hat{\Omega}_t},\quad j=1,\cdots, d, 
    $$
    where $p_{nj}^{(s)}, p_{nj}^{(t)}$ are the probabilities of selecting features in the CRFs trained on $D_s$ and $\tilde{D}_t$ respectively, $\hat{\omega}_j^t$ is the sample distance covariance based on an independent copy $\tilde{D}^\prime_t$ of $\tilde{D}_t$, and $\hat{\Omega}_t = \sum_{j=1}^d \hat{\omega}_j^t$. 
    For given $\alpha > 0$, and for any $0 < \eta<1/2-\alpha$, conditional on $(p_{nj}^t)_{j=1, \cdots,d}$ when $n_t$ is large enough, the MSE error bound of TL-CRF is 
    \begin{align*}
        \mathbb{E}\left[(\widehat{Y}(\mathbf{X}) - f_t(\mathbf{X}))^2\right] &\leq C_R\Bigg(|\mathcal{I}_R^c| \sum_{j\in\mathcal{I}_R^c  } \left\| \partial_j R \right\|_{\infty}^2 k_{n_t}^{2 \log_2 (1 - p_{nj}^t / 2)} + M_t^2 e^{-n_t/(2 k_{n_t})}\\
&\qquad \quad + \frac{12 \sigma_t^2 k_{n_t}}{n_t} \frac{8^{d}}{
    \sqrt{
        \prod_{j \in \mathcal{I}_R^c} p_{nj}^t \times \log_2^{|\mathcal{I}_R^c| - 1}(k_{n_t})
    }
}\Bigg)\\ 
&+ C_s(4k_{n_t}+2)\Bigg(d \sum_{j=1}^d \left\| \partial_j f_s \right\|_{\infty}^2 k_{n_s}^{2 \log_2 (1 - p_{nj}^s / 2)} + M_s^2 e^{-n_s/(2 k_{n_s})}.\\
&\quad  + \frac{12 \sigma_s^2 k_{n_s}}{n_s} \frac{8^{d}}{
    \sqrt{
        \prod_{j=1}^d p_{nj}^s \times \log_2^{d - 1}(k_{n_s})
    }
}\Bigg).
    \end{align*}
\end{theorem}

The following corollary simplifies the result for ease of understanding the dependence of the MSE on key model parameters, making appropriate assumptions on the depths of the trees.

\begin{corollary}
\label{cormain}
From the results of Theorem \ref{main}, if $p_{nj}^t$ is defined as in Theorem \ref{main}, then there exists $ \ p_{\epsilon_t} > 0$ independent of $n_t$, s.t.
 $$
 p_{\epsilon_t}\leq \min_{j\in \mathcal{I}_R^c}p_{nj}^t
 $$
 with high probability.
 
 \noindent Further, if 
  $r(p) := \frac{2 \log_2 (1-p/2)}{2 \log_2 (1-p/2) - 1}$, and define $r_s = r(1/d), r_t = r(p_{\epsilon_t})$, and select 
 \begin{align}
k_{n_s} &= C_s\left(n_s \left(\log_2^{d-1} n_s\right)^{1/2}\right)^{1 - r_s}, \nonumber \\
k_{n_t} &= C_t\left(
\left(n_t \left(\log_2^{|\mathcal{I}_R^c|-1} n_t\right)^{1/2}\right)^{-1}+
        \left(n_s \left(\log_2^{d-1} n_s\right)^{1/2}\right)^{-r_s}
\right)^{r_t-1}
\label{knt}
\end{align}
where $C_s$ and $C_t$ are constants independent of $n_s$ and $n_t$. Then, given $(p_{nj}^t)_{1\leq j\leq d}$,
\begin{align*}
    \mathbb{E}\left[\left(\hat{Y}(\mathbf{X},(D_s,\tilde{D}_t)) - f_t(\mathbf{X})\right)^2\right]
    \leq  \tilde{C}\left(\left(n_t (\log_2^{|\mathcal{I}_R^c|-1} n_t)^{\tfrac{1}{2}}\right)^{-1} + \left(n_s(\log_2^{d-1}n_s)^{\tfrac{1}{2}}\right)^{-r_s}\right)^{r_t}
\end{align*}
\end{corollary}

\vspace{0.3in}

Examining the proof of this corollary, we have the following remarks. 

Define 
$$
h(n_s, n_t) = \frac{\left(n_s(\log_2^{d-1}n_s)^{1/2}\right)^{r_s}}{n_t(\log_2^{|\mathcal{I}_R^c|-1}n_t)^{1/2}}.
$$

When $\lim_{n_s, n_t \to \infty}h(n_s,n_t) > 0$, then from Equation \ref{knt} in Corollary \ref{cormain}, we have, $k_{n_t} = C_t\left(n_t(\log_2^{|\mathcal{I}_R^c|-1}n_t)^{1/2}\right)^{1-r_t}$, and consequently, 
$$
\mathbb{E}\left[\left(\hat{Y}(\mathbf{X},(D_s,\tilde{D}_t)) - f_t(\mathbf{X})\right)^2\right]\leq C\left(n_t(\log_2^{|\mathcal{I}_R^c|-1}n_t)^{1/2}\right)^{-r_t},
$$
for some large enough generic constant $C>0$. In order to understand and compare the rate further we consider a further simplification.
 When the population distance covariances $(\omega_j^t)_{j\in \mathcal{I}_R^c}$ are approximately the same and $\lim_{n_s, n_t \to \infty}h(n_s,n_t) > 0$ then in Corollary \ref{cormain}, the depth of the trees would be
 \[
 k_{n_t} = O\left(n_t\left(\log_2^{|\mathcal{I}_R^c|-1}n_t\right)^{1/2}\right)^{1-\bar{r}_t}
 \]
and the rate in the upper bound would be
$$
O\left(n_t\left(\log_2^{|\mathcal{I}_R^c|-1}n_t\right)^{1/2}\right)^{-\bar{r}_t}
$$
where
$$\bar{r}_t := \frac{2\,\log_2\bigl(1 - |\mathcal{I}_R^c|^{-1}/2\bigr)}%
{2\,\log_2\bigl(1 - |\mathcal{I}_R^c|^{-1}/2\bigr) - 1}
= \frac{1}{|\mathcal{I}_R^c|\log 2 + 1}\,(1 + \delta_t)\,.$$
and $\delta_t$ are some positive quantity that decreases to zero as $|\mathcal{I}_R^c|$ become large.

This rate is lower than the rate of \textit{random forest trained on the target data only}, indicating the advantage of transfer learning in this case. Note the error rate for the random forest trained on the \textit{target domain only} with \textit{uniformly selected weights} \cite{klusowski2021sharp} is $\left(n_t(\log_2^{d-1}n_t)^{1/2}\right)^{-r(1/d)}$. We remind the reader that we do not assume sparse function in the target domain. Ignoring the log terms, then the rate for transfer learning is $O(n_t^{-1/l})$, where $l=|I_R^c|<<d$ by assumption, while the rate for target domain only CRF is  $O(n_t^{-1/d})$. Therefore with transfer learning we have a faster rate of convergence. Analyzing the condition $\lim_{n_s, n_t \to \infty}h(n_s,n_t) > 0$, and ignoring the log terms, we note a sufficient condition is $n_s^{1/d} \gtrsim n_t$, i.e., the source domain has a lot more data available than the target domain.

Another way to think of the benefit of transfer learning is in terms of the \textit{depth of the trees}. Note that the depth of the trees in the residual random forest in the target domain is $\log_2 k_{n_t}= O(\log_2n_t)$, which is much lower than the depth of the trees in the source domain, namely, $ O(\log_2 n_s)$, when $n_s >> n_t$. The ``implicit regularization'' in trees of the random forest is provided through depth $\log_2 K_n$. As can be seen from the bias and variance results, trees of lower depth have higher bias, while having a lower variance. In the target domain, due to a small sample size, we are forced to fit trees of smaller depth, and hence $K_n$ needs to be smaller. This could lead to high bias, resulting in deteriorating performance, unless the function being fit is sufficiently less complex, such as a sparse function. Our transfer learning procedure here transfers the knowledge of a larger depth source domain tree so that a smaller depth tree in the target domain suffices to fit a sparse function. However, there is still a problem of high variance due to having a large number of features and a small sample size in the residual random forest. Here, our distance covariance-based weight selection comes into play as it enables the estimation of a sparse difference function in the target domain without a high variance.

On the other hand, when $\lim_{n_s, n_t \to \infty}h(n_s,n_t) = 0$, then from Equation \ref{knt} in Corollary \ref{cormain}, we have,  
$k_{n_t} = C_t\left(n_s(\log_2^{d-1}n_s)^{1/2}\right)^{r_s(1-r_t)}$, and consequently, 
$$
 \mathbb{E}\left[\left(\hat{Y}(\mathbf{X},(D_s,\tilde{D}_t)) - f_t(\mathbf{X})\right)^2\right]\leq C\left(n_s(\log_2^{d-1}n_s)^{1/2}\right)^{-r_sr_t}.
$$
 
Under the simplified setup discussed above and once again ignoring log terms we have the MSE bound as $O(n_s^{-1/(dl)})$.  The condition on  $h(n_s,n_t) \to 0$, implies $n_s^{1/d} << n_t$ and therefore in this case source domain does not have vastly large amount of data. Therefore, this upper bound is bigger than $n_t^{-1/l}$, but might still be smaller than $n_t^{-1/d}$, especially if $n_s^{1/l} > n_t$. Hence, there is still an advantage over the vanilla CRF in the target domain if $n_s^{1/d} < n_t < n_s^{1/l} $.

This remark above shows the benefit of transfer learning clearly. We can compare it with the minimax optimal rate for nonparametric regression with target domain data alone for the class of functions assumed, namely $O(n_t^{-\frac{2}{d+2}})$ \citep{biau2012analysis}. As \cite{biau2012analysis, klusowski2021sharp} note, this rate is generally not achieved by CRF. However, if the target domain regression function is sparse (depends on only a subset of features and is conditionally independent of the other features), and if we know which features are relevant, then the rate for CRF can be better than the minimax rate over the original problem. However, in our problem setup, we did not assume sparse functions in the target domain, but only the difference function between source and target domains to be sparse. With this assumption and an assumption on availability of large database at the source domain, transfer learning procedure then has an upper bound which is better than the minimax rate for the target domain problem alone. This observation is similar to the findings of \cite{cai2024transfer}. The rates in \cite{cai2024transfer} are not directly comparable to our rates since \cite{cai2024transfer} considered a H\"{o}lder ($\beta,L$) smoothness class which is a stronger assumption than ours and assumed the difference function to be well approximated by a polynomial function which cannot be easily compared with our sparse difference function assumption.  Further, while \cite{biau2012analysis,klusowski2021sharp} did not provide a method for selecting the weights such that the upper bounds for sparse regression functions hold with high probability, we provide a distance covariance-based algorithm for obtaining the weights. In this aspect, our results are related to \cite{klusowski2021nonparametric} and provide an alternative to their decision stumps-based variable screening procedure.

\section{Takeaways for Standard random forest}
We can generalize our transfer learning algorithm to Breiman’s standard random forest \cite{breiman2001random}. In this section, we discuss the proposed method and provide simulation evidence suggesting it performs well. Recall that when building a tree in Breiman’s random forest, at each branch of the tree, we randomly select $m$ features with equal probability and choose the best split among the selected $m$ features that minimizes our objective function (typically $m$ is recommended to be $\sqrt{d}$). Our idea is to use distance covariance to conduct a \textit{soft feature screening} when we select $m$ features for the \textit{residual random forest}. That is, the probability of each feature being selected as one of the $m$ features is proportional to the sample distance covariance between the feature and the residuals. Intuitively, suppose the difference function between the source and the target functions only depends on a subset of the original feature set. In that case, the distance covariance feature screening method can help us select dependent features with high probability. The actual feature to split on out of this $m$ features and the location of the split is still data-dependent (using CART type criteria). This approach does not sacrifice the randomness of feature selection that is important for heterogeneity among the trees and partly responsible for random forest method's empirical success. The feature set is still randomly chosen, however, distance covariance guides which features are more likely to appear in the set. The proposed method of standard random forest with distance covariance weighting (RF-DCOV) is shown in Algorithm \ref{alg:srf}.

\begin{algorithm}[h]
\caption{Distance covariance weighted Standard Random Forest (RF-DCOV)}
\label{alg:srf}
\begin{longlist}

  \item[1.] \textbf{Training.}\quad 
    Given training data 
    $\displaystyle D=\{(X_i,Y_i)\}_{i=1}^n$ 
    with $d$ features, train $t=1,\dots,T$ trees:
    \begin{longlist}
    \item Calculate the empirical distance covariance $\widehat{{\rm dCov}}(\mathbf{X}_{.,j}, Y)$ between $j$-th feature and the response variable.
    \item Generate a bootstrap sample $D^{(b)}$ of size $n_{boot}$ from $D$.
    \item Grow a CART tree $h_t(\cdot)$ with maximum depth $l$:
        at each node, select a random subset $m \subseteq \{1,\dots,d\}$ with probabilities 

        $$
    \frac{\widehat{\mathrm{dCov}}(\mathbf{X}_{.,j}, Y)}
     {\sum_{k=1}^d \widehat{\mathrm{dCov}}(\mathbf{X}_{.,k}, Y)}
        $$
        for feature $j$, with $j=1,2,\cdots, d$,  
        then choose the best split among features in $m$.
      \item Collect $h_t$ into the ensemble $\mathcal{F}=\{h_t\}_{t=1}^T$.
    \end{longlist}

  \item[2.] \textbf{Prediction.}\quad 
    For a new input $\mathbf{X}$:
    \[
      \hat{Y}(x) \;=\ \dfrac{1}{T}\sum_{t=1}^T h_t(\mathbf{X});
    \]

\end{longlist}
\end{algorithm}

\begin{figure}[h]
    \centering    \includegraphics[width=0.6\linewidth]{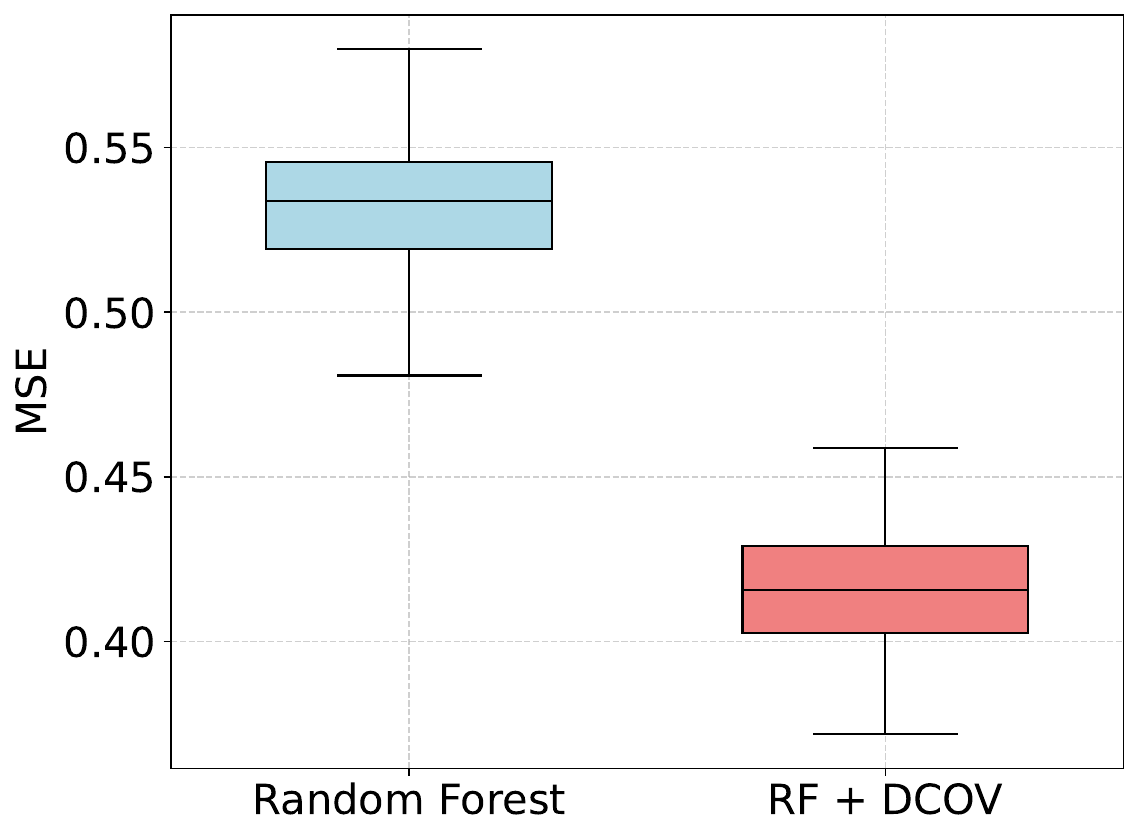}
    \caption{Distance covariance aided standard random forest for sparse functions}
    \label{fig:srf_dcov}
\end{figure}

This intuition can be confirmed by the simulation shown in Figure \ref{fig:srf_dcov}. We see from this simulation that this soft feature screening method can improve the performance of standard random forest when the model is sparse in the sense that the true data-generating function only depends on a few features.  In Figure \ref{fig:srf_dcov}, the sample size of the training data is $n=10000$ and the test data is $n_{test} = 100$. Assume there are 50 candidate features and only 25 of them contribute to the response variable. Specially, we generate data from,
\begin{align*}
    Y_i & = f(\textbf{X}_i) + \epsilon_i, \ i=1, 2, \cdots, n \\
    f(\bb x) & = \sum_{j=1}^{\frac{d}{4}} e^{- x_j} + \sum_{j=\frac{d}{4}+1}^{d/2} \tanh(x_j)
\end{align*}
and $\epsilon_i\overset{\text{i.i.d.}}{\sim} N(0,1)$. We randomly select $m=15$ features at each split in both RF and RF-DCOV algorithms. The only difference is that our new algorithm selects those features with probabilities proportional to the distance covariances between the feature and the response variable estimated by all the observations (in the training data). The number of trees in the forests is 100 and each tree gets a bootstrapped sample of size 1000. The maximum depth of the trees is set to $\log_2 n=10$. The simulation runs for 100 times with the same setting. This simulation shows the distance covariance aided random forest has significantly better out-of-sample MSE than the regular random forest (Figure \ref{fig:srf_dcov}).

\begin{figure}[h]
    \centering
    \begin{subfigure}{0.5\textwidth}
  \includegraphics[width=1\linewidth]{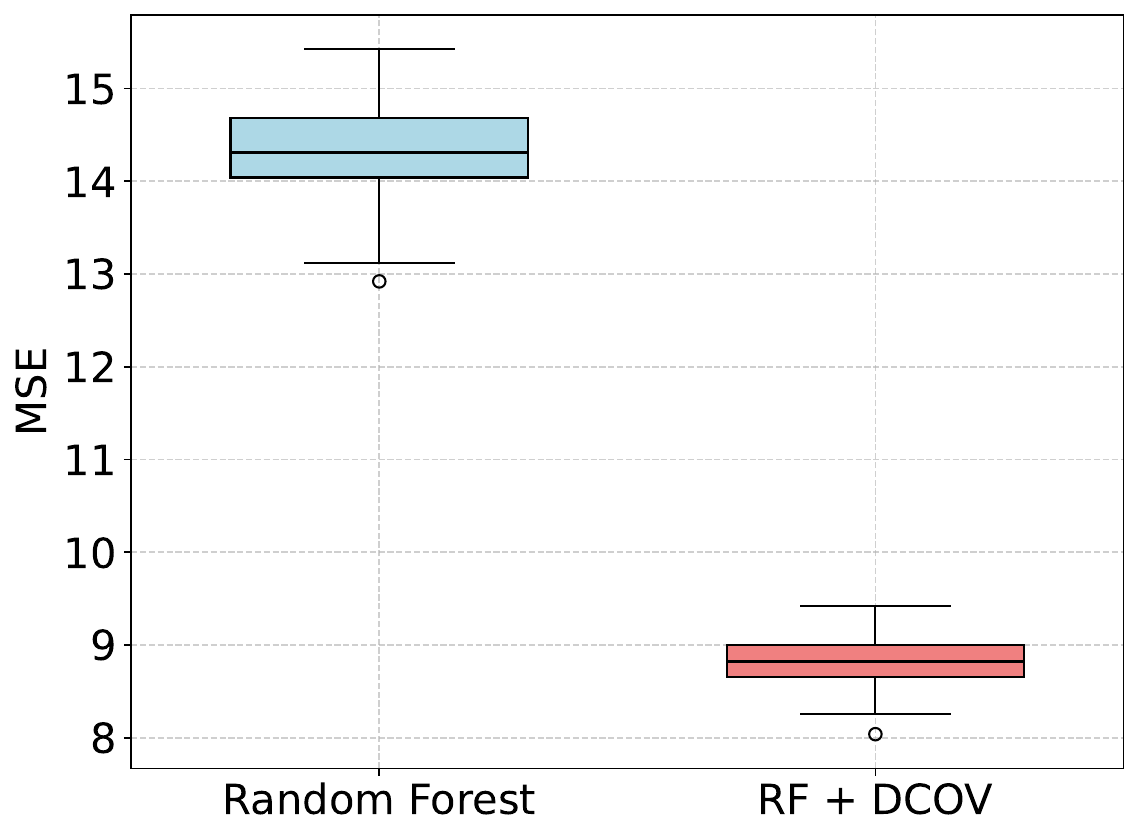}      
    \end{subfigure}%
    \begin{subfigure}{0.5\textwidth}
            \includegraphics[width=1\linewidth]{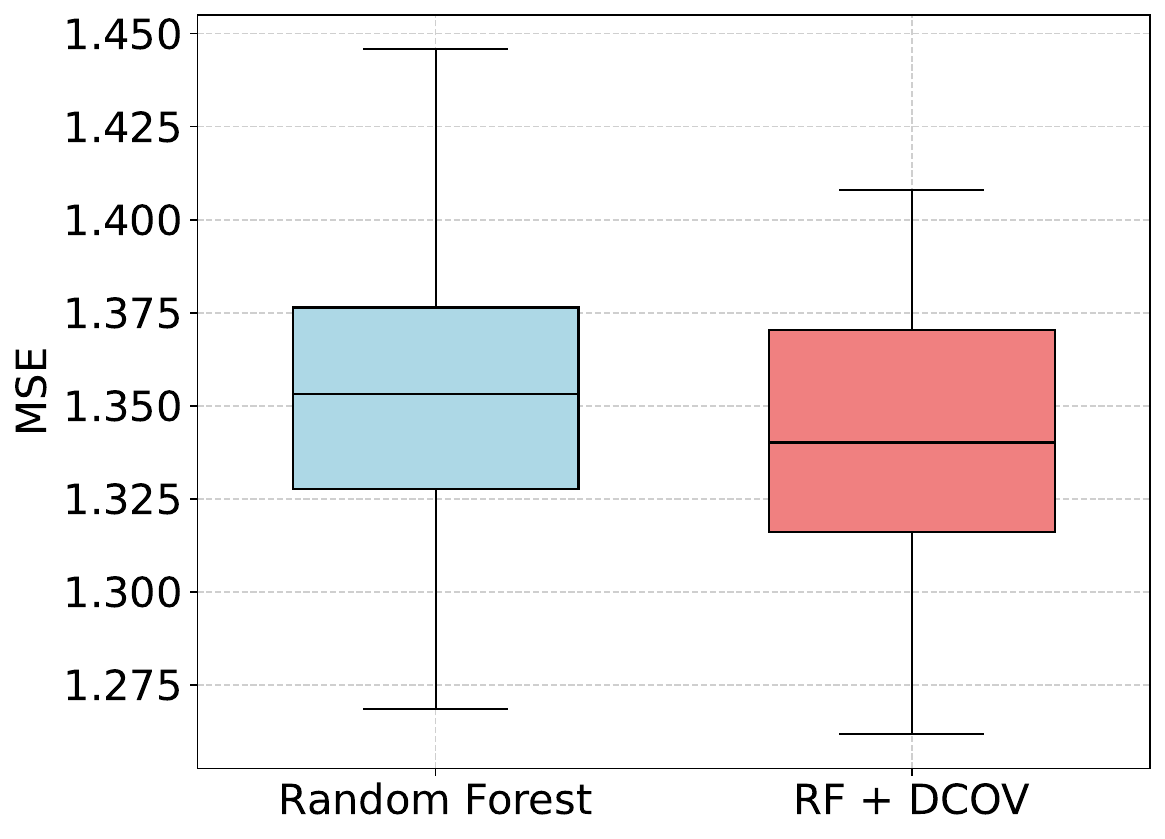}
    \end{subfigure}
    \caption{Benefits of distance covariance weighting when some features dominate (left) and no benefit when all features contribute similarly (right).}
    \label{fig:srf_dcov_strong}
\end{figure}

We further investigate this empirical observation by contrasting performance in two scenarios with two more simulations. Consider the following two data-generating functions.
\[
f_1(x) = \sum_{i=1}^{\frac{d}{2}} e^{- x_i } + \sum_{i=\frac{d}{2}+1}^{d_0} \tanh(x_i) + \sum_{i=d_0 + 1}^{d} 6\sin(2\pi x_i), \quad f_2(x) = \sum_{i=1}^{\frac{d}{2}} e^{- x_i} + \sum_{i=\frac{d}{2}+1}^{d} \tanh(x_i).
\]
The first function clearly has the last $(d-d_0)$ features, which dominate the function value. Intuitively, if one can estimate the functional relationship with those predictors well, then one would be able to achieve a reasonably low MSE. 
 This can be seen by noting that in the first function, we have the functions $6\sin (2\pi x)$ ranging within $[-6, 6]$, whereas other functions have ranges no larger than 1. The second function, on the other hand do not have any features that is a dominant determinant of the function value. What we can see from Figure \ref{fig:srf_dcov_strong} is that if there are particularly strong features that explain large variation in the response, then distance covariance weighting as proposed here is quite effective (Left figure). On the other hand it is not so effective if all features contribute similarly to the variation in the data.

With this new distance covariance weighted standard random forest, we propose the following method for transfer learning with a standard random forest. The procedure is similar to the procedure for centered random forest, except that we train standard random forests in the source domain and in the target domain, we train a distance covariance weighted standard random forest on the residuals using Algorithm \ref{alg:srf}. The method, which we call TLSRF is presented in Algorithm \ref{alg:transfer-rf-standard}.

\begin{algorithm}[h]
\caption{Transfer Learning with Standard Random Forest (TLSRF)}
\label{alg:transfer-rf-standard}
\begin{longlist}
  \item \textbf{Source‐domain modeling.}\quad 
    Given source data 
   $\displaystyle D_s= (Y^{(s)}, \mathbf{X}^{(s)})$, train 
    a SRF $\hat{Y}(\cdot;D_s)$  using uniform weights for all features.
  \item \textbf{Residualization in target.}\quad 
    For each target point $\mathbf{X}_{i}^{(t)}$, compute the source prediction $
      \hat{Y}_{i}^{(s,t)} = \hat{Y}(\mathbf{X}_{i}^{(t)},D_s)$, and then the residuals are
    \(\widetilde{Y}_{i}^{(t)}=Y_{i}^{(t)}-\hat{Y}_{i}^{(s,t)}\).
  \item \textbf{Target‐domain calibration.}\quad 
    Let 
    \(\widetilde{D}_t=\{\widetilde{Y}_{i}^{(t)},\mathbf{X}_{i}^{(t)}\}_{i=1}^{n_t}\).  Divide the data into two parts $\widetilde{D}_t',\widetilde{D}_t''$.
    Train a distance covariance weighted standard Random Forest $\hat{Y}(\cdot;\widetilde{D}_t'),$ using Algorithm~\ref{alg:srf}, with weights (distance covariance between residual and feature)
    \(\widehat{\mathrm{dCov}}(\mathbf{X}_{.,j},\tilde{Y})\) obtained from $\tilde{D}_t''$
    and the final prediction would be
    \[
      \hat{Y}_{i}^{(t)} 
      = \hat{Y} \bigl(\mathbf{X}_{i}^{(t)};D_s\bigr) \;+\; \hat{Y}\bigl(\mathbf{X}_{i}^{(t)};\widetilde{D}_t\bigr).
    \]
\end{longlist}
\end{algorithm}

\section{Simulation}
In this section, we conduct a comprehensive simulation study to illustrate the advantages of our Transfer RF methods over fitting a random forest in the target data alone. We consider both the Centered Random Forest (CRF) and the standard Breiman's Random Forest (SRF). While our theoretical results only apply to the centered random forest, our method of transferring knowledge is applicable for both data-driven and non-data-driven choices of splits. We will call the procedures with transfer learning as Transfer Learning Centered Random Forest (TLCRF) and Transfer Learning Standard Random Forest (TLSRF), respectively. 

The setting for the centered random forest will be introduced first. Recall our notation, $\bb X^{(s)}\in \mathbb{R}^{n_s\times d}, \bb X^{(t)} \in \mathbb{R}^{n_t\times d}, \bb X_{test} \in \mathbb{R}^{n_{test}\times d} $ are feature matrices of source data, target data and test data with $d$ features. We obtain these matrices by randomly sampling each entry from $\mathcal{U}(0,1)$ distribution. The response variables in the source and target domain are generated based on the following expressions:
\begin{align}
    Y_{i}^{(s)} = f_s(\bb X_{i}^{(s)}) + \epsilon_{i}^{(s)}, & \quad     Y_{i}^{(t)} = f_t(\bb X_{i}^{(t)}) + \epsilon_{i}^{(t)} \\
    f_s(\bb x) & = \sum_{i=1}^{\frac{d}{2}} e^{- x_i} + \sum_{i=\frac{d}{2}+1}^{d} \tanh(x_i),\\
    f_t(\bb x) & = \sum_{i=1}^{\frac{d}{2}} e^{- x_i } + \sum_{i=\frac{d}{2}+1}^{d_0} \tanh(x_i) + \sum_{i=d_0 + 1}^{d} 6\sin(2\pi x_i)
\end{align}
 and the error term $\epsilon_{i}^{(s)}, \epsilon_{i}^{(t)}\overset{\text{i.i.d.}}{\sim} N(0,1)$. Here $d_0:=d - \lfloor d \cdot r \rfloor$ and $r$ is what we call the discrepancy ratio. Note that this setup assumes $d/2$ features are related to $Y$ as negative exponential function and are the same for both domains, while the remaining features are related as $\tanh(\cdot)$ function. The discrepancy ratio $r$ is the fraction of the features that are related to $R(x):=f_t(x) - f_s(x)$. We increase this ratio from $0$ to $0.5$ to increase the number of features whose functional association with the response is different across domains. We can see that the difference function is then given by
 \[
 R(x) = \sum_{i=d_0 + 1}^{d} (\tanh(x_i) - 6\sin(2\pi x_i)),
 \]
 which is dependent on $d-d_0$ features. This setup, therefore, mimics our assumption of non-linear (and non-polynomial) discrepancy between target and source domains that is limited to a fraction of features.

\begin{figure}[h]
    \centering
    \includegraphics[width=0.6\textwidth]{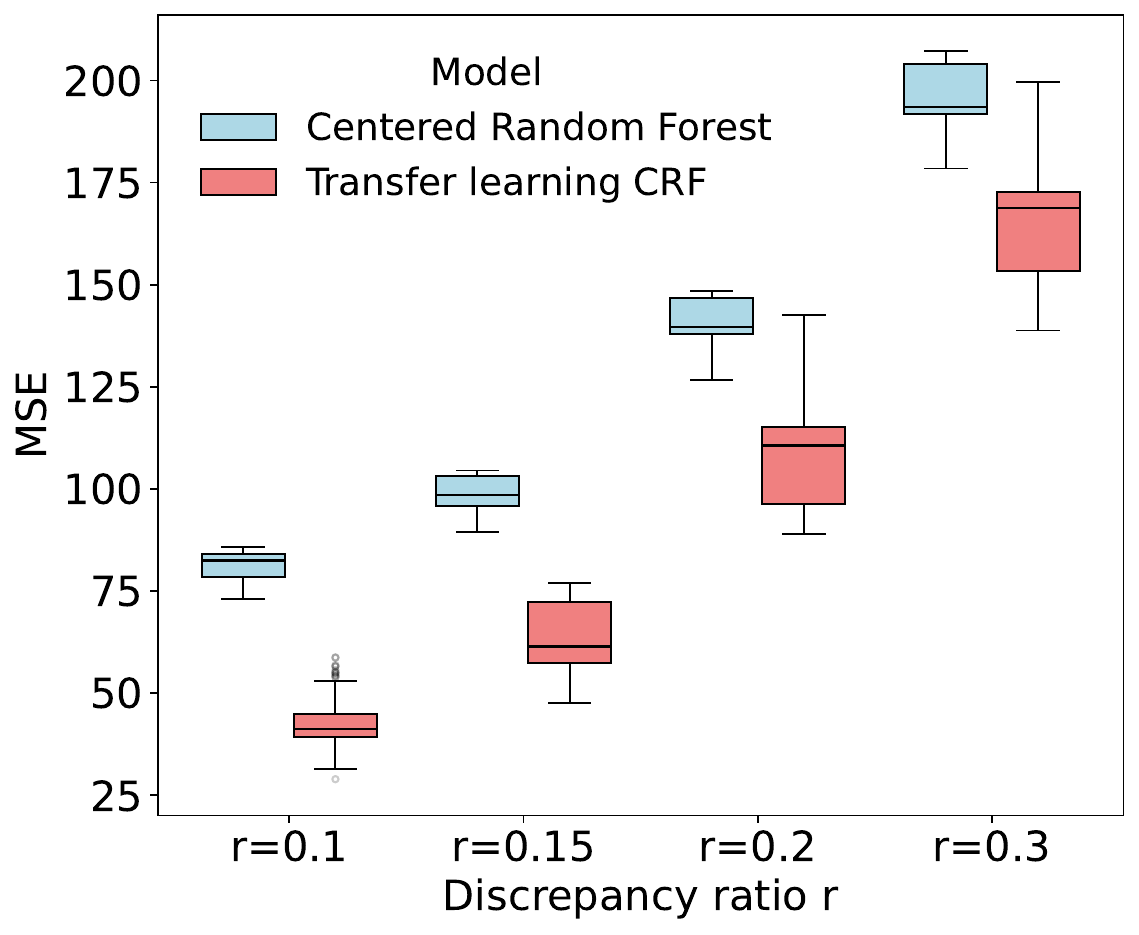}  % Adjust width
    \caption{Mean Square Error performance of TLCRF and CRF on test data, fixing $n_s = 20000, n_t = 500, n_{test} = 100, d = 50$, while increasing the discrepancy ratio $r$.}
    \label{tlcrf_discrepancy ratio}
\end{figure}

\subsection{Centered random forest}
We now describe our results for the centered random forest. Note that the split of each node is stopped only when the maximum depth is attained. Even if the node contains fewer than two data points, it continues to be split if the maximum depth is not attained. The maximum depth is determined by cross-validation with 5 partitions. The candidate depths are chosen from $\left\{h\leq \log_2 n: h = 2k+1, k=0,1,\cdots \right\}$ where $n$ is the training data size of the model. We always keep the number of trees in each forest as 100, and the number of replications as 100. We conduct two simulation studies for centered random forests. The first one varies the discrepancy ratio within $\left\{0.1, 0.15, 0.2, 0.3\right\}$, keeping the sample sizes $n_s=20000$, $n_t=500$, $n_{test}=100$, and the total number of features $d=50$ fixed. This therefore means that for $5,8,10,15$ features, the functional forms are different between the source and target domains. The results of this simulation is shown in Figure \ref{tlcrf_discrepancy ratio}. We see that the transfer learning method uniformly performs better than the model trained in target data, despite having large differences in the functional forms of some features between the domains. This shows the benefit of transfer learning from related yet different domains. As expected we see the performance advantage of transfer CRF over CRF reduces quite a bit as the number of features whose functional association is different across domains increases.

\begin{figure}[h]
    \centering
    \includegraphics[width=0.9\textwidth]{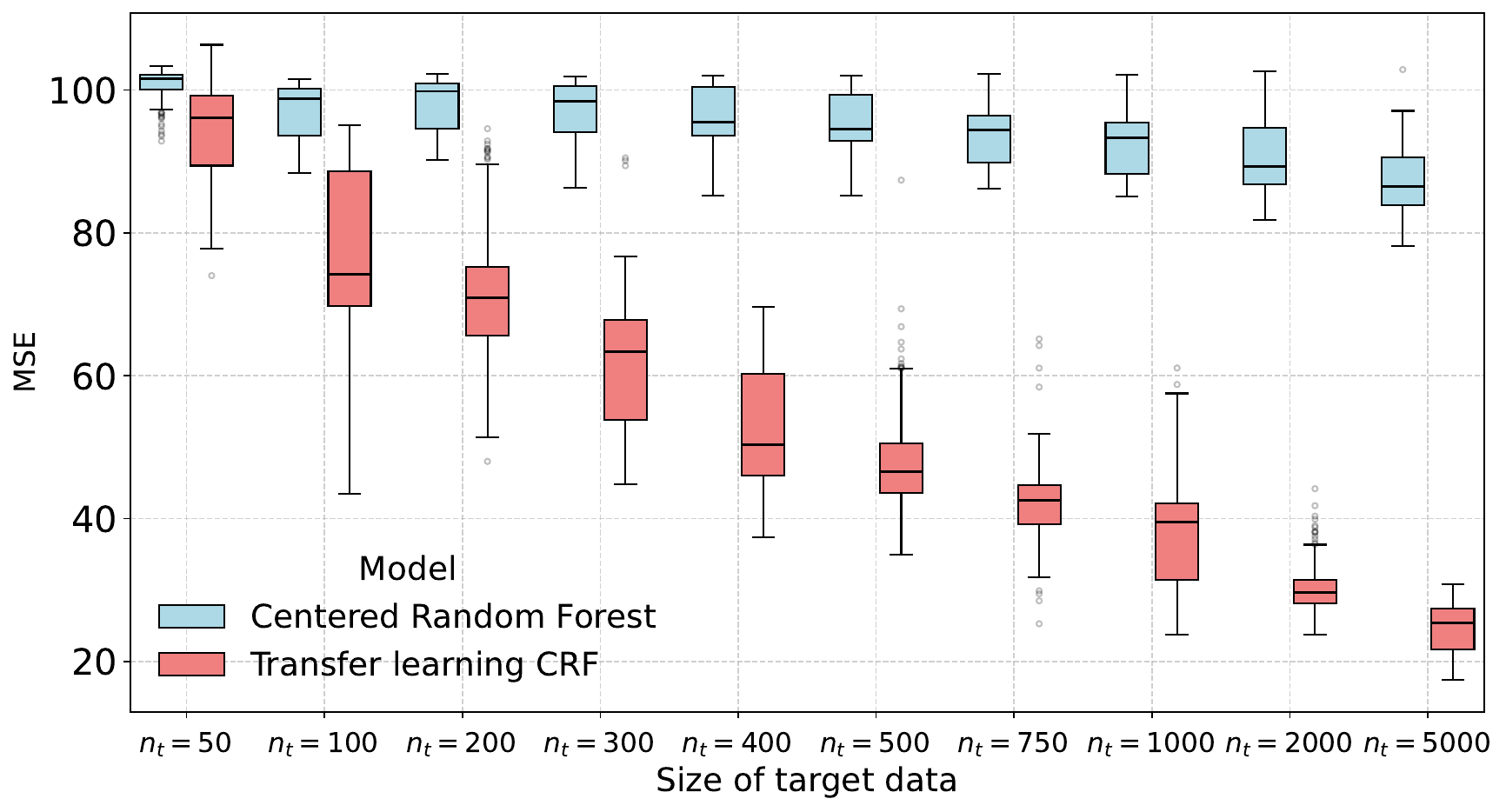}  % Adjust width
    \caption{Performance of TLCRF and CRF on test data as the size of target training data increases with $n_s = 10000, n_{test} = 100, d = 50, r = 0.1$.}
    \label{tlcrf_target sample size}
\end{figure}

Our second simulation varies the size of target data from $50$ to $5000$ but with unequal imcrements to map the entire range well. We fix the sample size of the source domain at 10000, the number of features at 50, and the discrepancy parameter at 0.1. The results are shown in Figure \ref{tlcrf_target sample size}. We see that as the target domain sample size increases the performance of both the TLCRF and CRF steadily improves. However, the decrease in MSE is substantially faster for TLCRF than for CRF. Therefore, we see high benefits of transfer learning when the target domain sample size is reasonably higher so that our method of bias correction works well. This makes sense since the discrepancy between the source and target domains is small (10\% of features), transfer learning is expected to have an advantage over target-only data. However, our procedure requires sample splitting and an estimation of distance covariance in the target data, and therefore needs the sample size in the target domain to be above a certain amount to be beneficial in comparison to target-only training.

Both of these methods perform better than source-trained CRF only due to the severe bias in the source only CRF as can be seen in the Supplementary Figures 1 and 2. The outperformance of TLCRF against the source-only CRF is uniform across all discrepency ratios in Supplementary Figure 1 and continues to hold as we increase target sample size in Supplementary Figure 2. This shows that when there is some discrepancy between the source and target domains, calibration using our methods improves transfer of knowledge. We also note that the variance in the source domain-trained CRF is much lower than both the target CRF and TLCRF due to having a much higher sample size in the source.

\subsection{Standard random forest} 
\begin{figure}[h]
    \centering
    \includegraphics[width=0.8\textwidth]{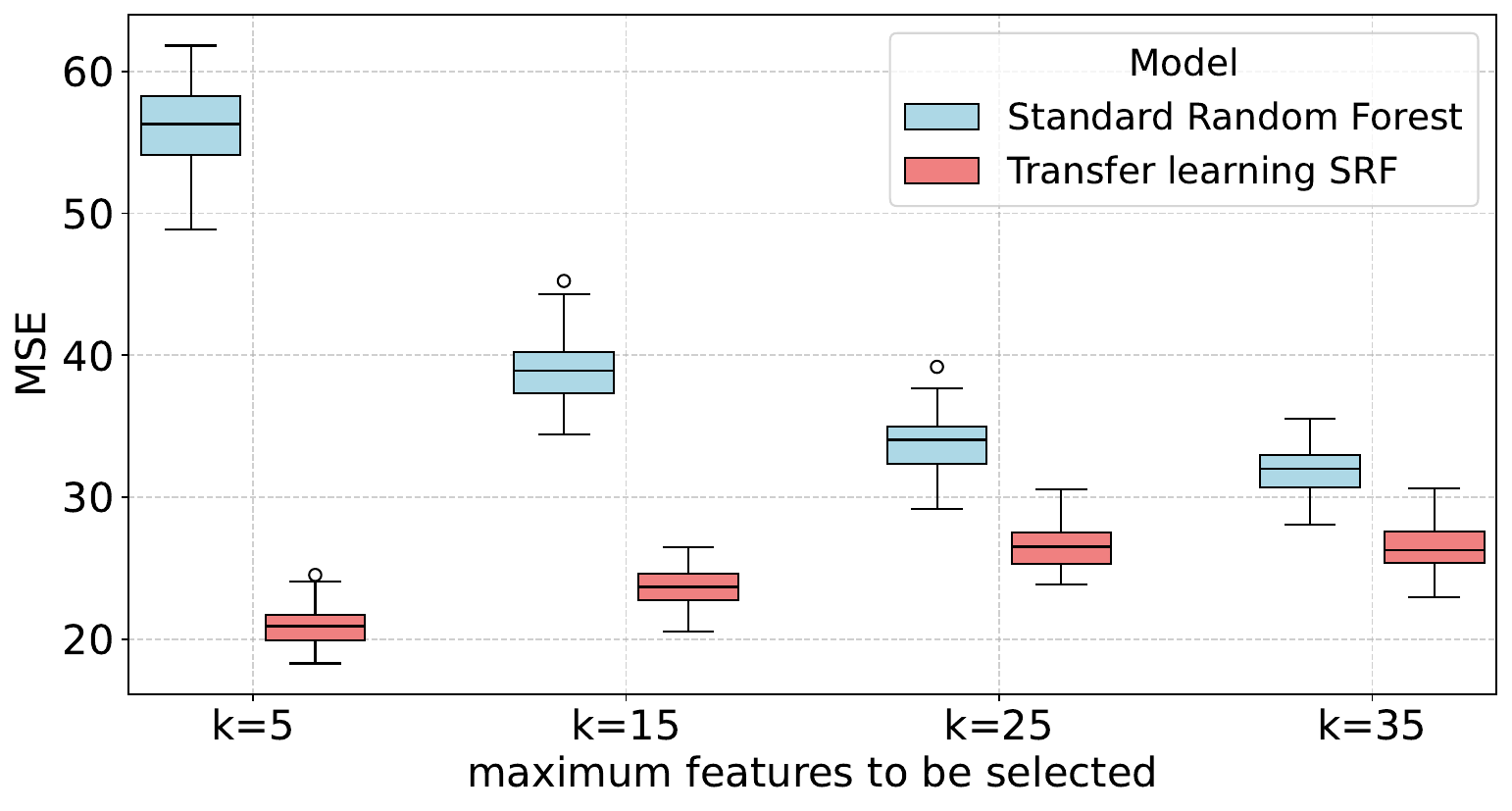}  % Adjust width
    \caption{Performance of SRF and TLSRF with differing number of features selected, with $n_s = 10000, n_t = 500, n_{test} = 100, d = 50, r = 0.1$}
    \label{srf_selected feature}
\end{figure}

Next, we study the performance of our transfer learning method with standard random forest in Algorithm \ref{alg:transfer-rf-standard}. We use the same source function and target function as those in the simulations of centered random forest. We run two simulation scenarios for SRF as well. The first one is to investigate the effect of the maximum features to be selected on the performance of our TLSRF algorithm and compare with SRF in target. The number of trees is set at 50 and the number of replications is set at 100. The bootstrapping size is set at 100 for both the target domain only SRF and the residual standard random forest we fit in our transfer learning method. The bootstrapping size is 1000 for the source model in the transfer learning. Therefore, the corresponding maximum depth is set at $\lceil\log_2 100 \rceil = 7$ and $\lceil\log_2 1000 \rceil = 10$ respectively. The number of features available to select from at each node split (``mtry'') is varied from $\left\{5, 15, 25, 35\right\}$. The result is shown in Figure \ref{srf_selected feature}.

We see from Figure \ref{srf_selected feature} that transfer learning uniformly performs better than the target data-only model. The performance of TLSRF is relatively stable as we change the number of features selected at each split, while the performance of the target data only SRF improves as the number of features made available to split increases. 

\begin{figure}[h]
    \centering
    \includegraphics[width=0.8\textwidth]{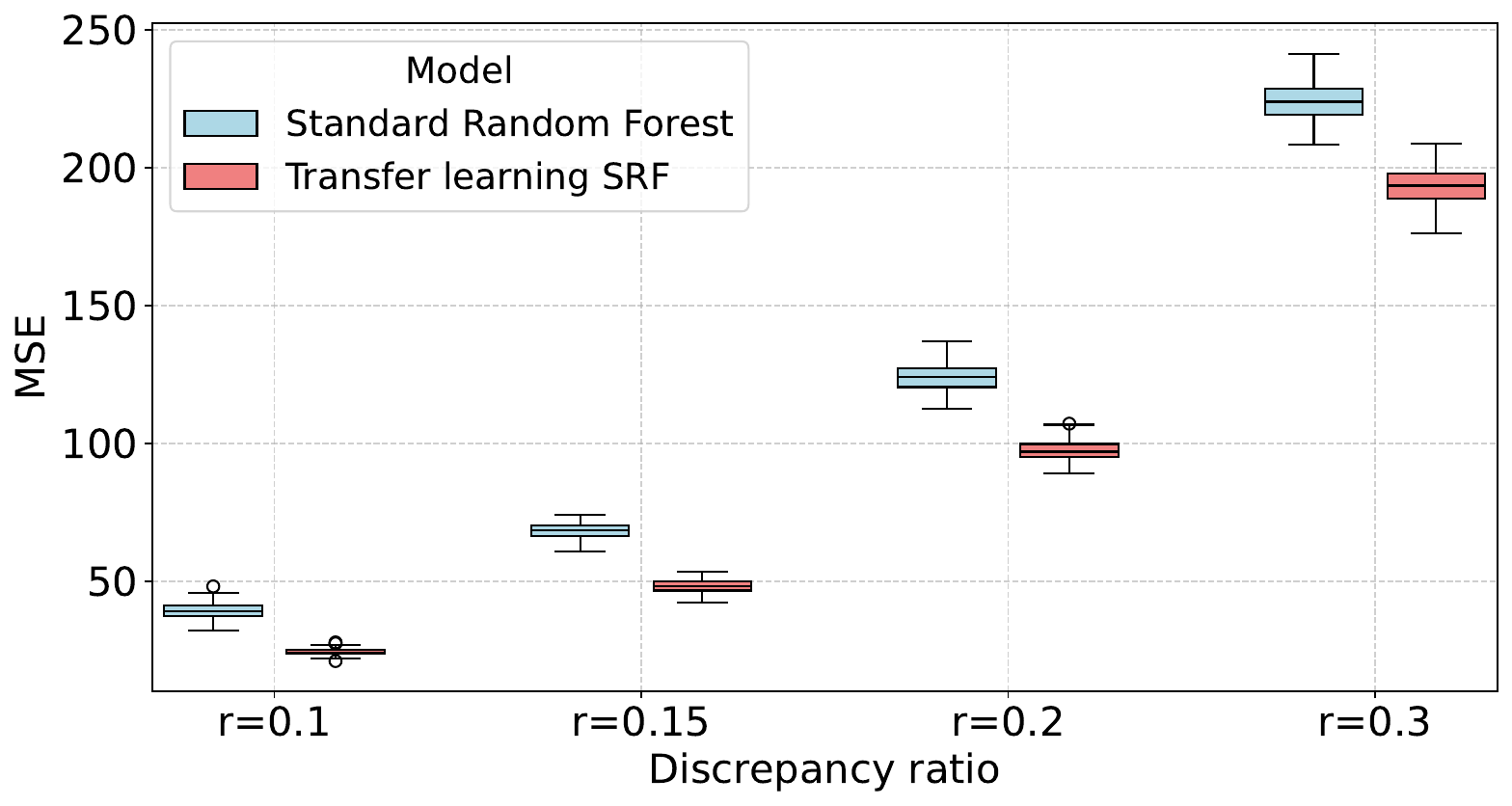}  % Adjust width
    \caption{Performance of SRF and TLSRF with increasing discrepancy ratio, fixing, $n_s = 10000, n_t = 500, n_{test} = 100, d = 50, r = 0.1$.}
    \label{srf_discrepancy ratio}
\end{figure}

The second simulation is to investigate the effect of discrepancy ratio on the performance of the proposed algorithm. We vary the discrepancy ratio from $0.1$ to $0.3$, and the number of selected features is fixed as 15. All the other settings are the same as those in the first simulation.
The result is shown in Figure \ref{srf_discrepancy ratio}. Here also we observe that the performance of TLSRF is uniformly better than the performance of the target only random forest. Note that because of the way the simulation is setup as discrepency increases, the magnitude of the response variable also increases and hence the MSE keeps climbing. However, we are primarily interested in assessing the relative performance of SRF vs TLSRF.

\subsection{An alternative simulation setup}

\begin{figure}[h]
    \centering
    \begin{subfigure}{0.5 \textwidth}
        \includegraphics[width=\textwidth]{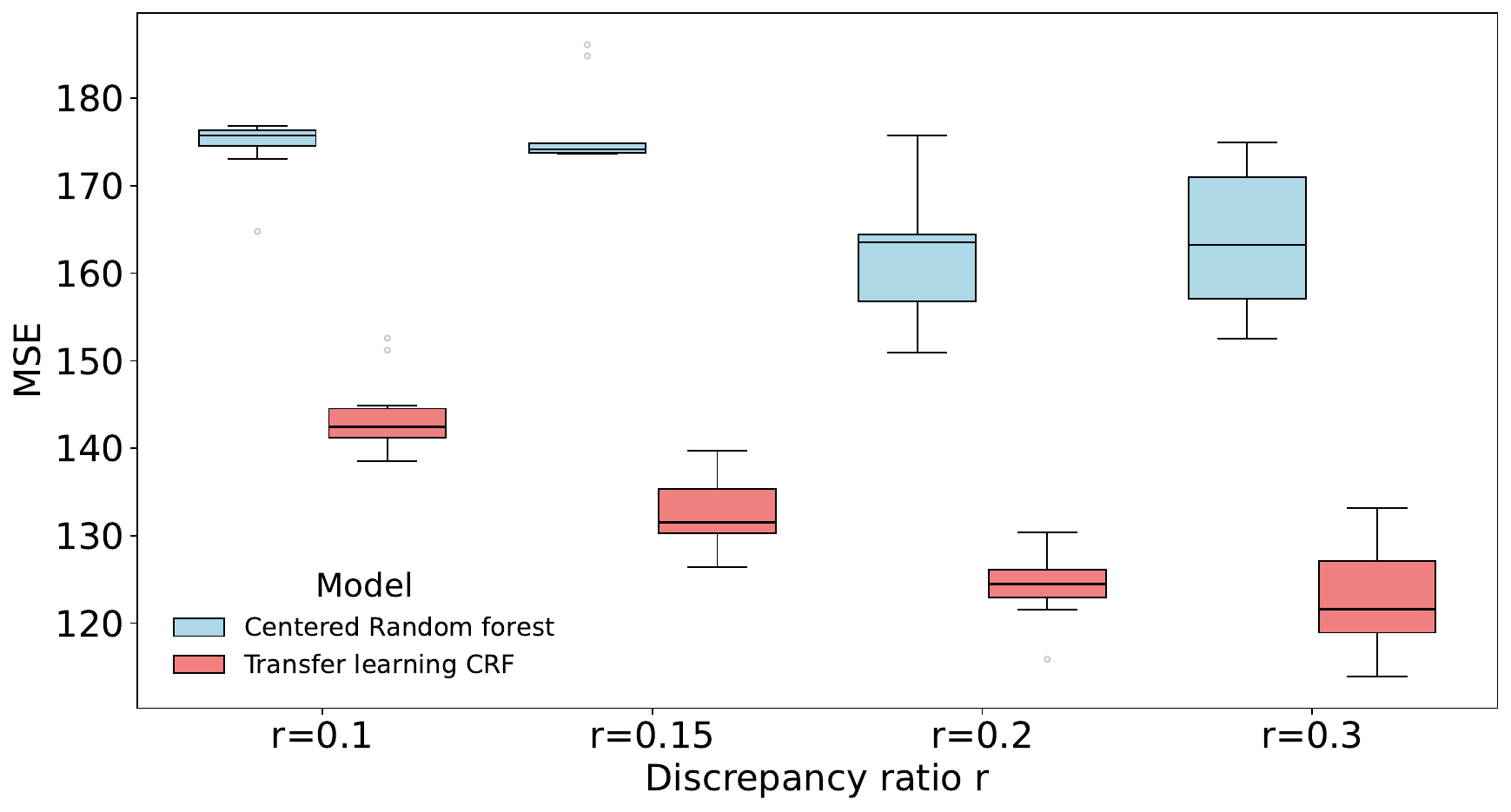} 
    \end{subfigure}%
     \begin{subfigure}{0.5 \textwidth}
        \includegraphics[width=\textwidth]{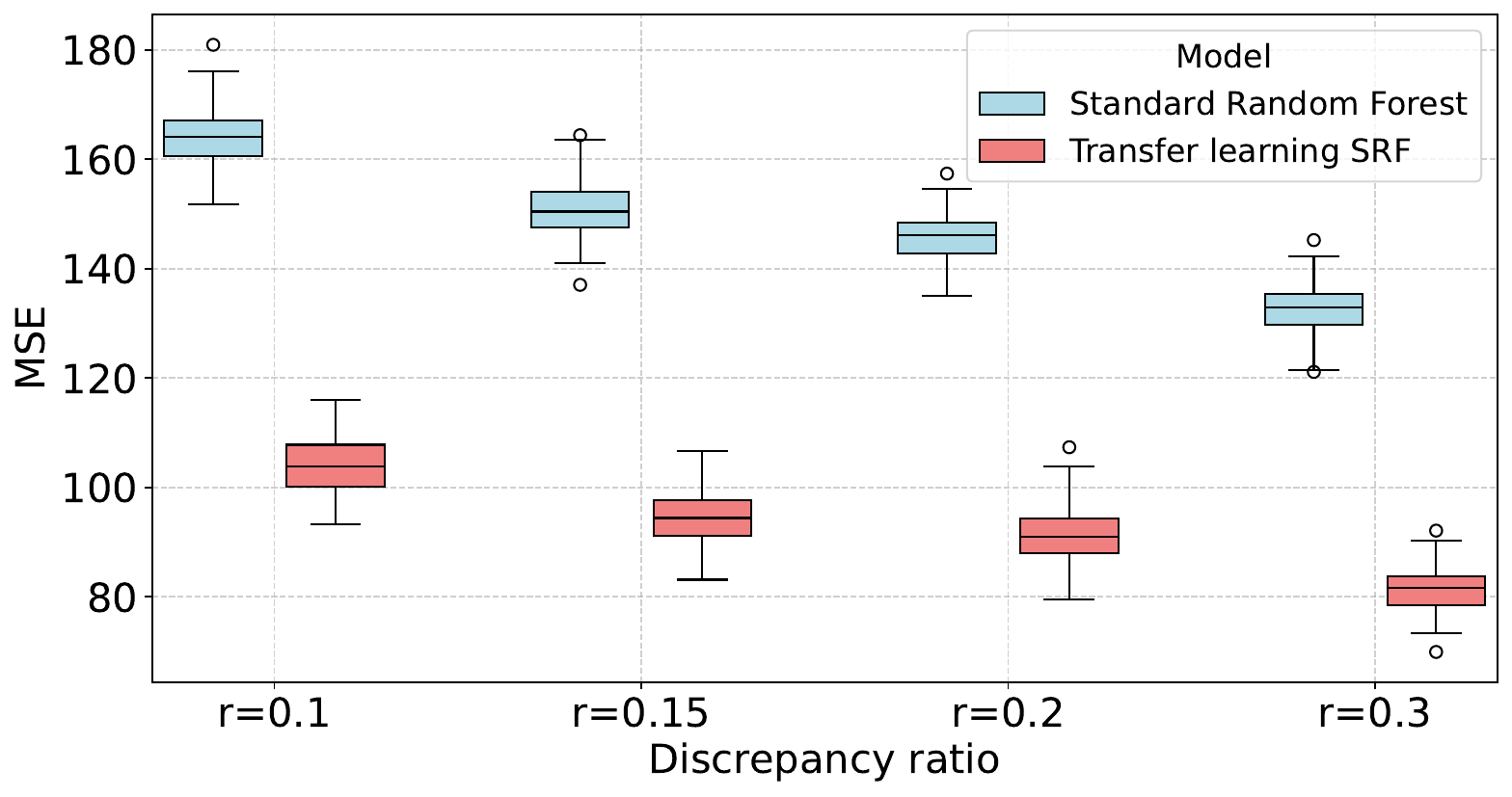}
        \end{subfigure}
    \caption{A comparison of performance of transfer learning with target only RF with increasing discrepancy ratio for the second simulation setting, (left) CRF vs TLCRF and (right) SRF vs TLSRF}
    \label{tl_discrepancy ratio_new}
\end{figure}

We end this section with simulation results under a different function discrepancy. Using the same notations as before, we generate data where the source and target domain functions are as follows,
\begin{align}
    f_s(x) & = \sum_{i=1}^{\frac{d}{2}} 3\cos\left(\pi x_i\right) + \sum_{i=\frac{d}{2}+1}^{d} e^{2.5-x_i},\\
    f_t(x) & = \sum_{i=1}^{\frac{d}{2}} 3\cos\left(\pi x_i\right) + \sum_{i=\frac{d}{2}+1}^{d_0} e^{2.5-x_i} + \sum_{i=d_0 + 1}^{d} e^{1.5-x_i}
\end{align}
 such that $R(x)=\sum_{i=d_0+1}^d (\exp(2.5-x_i)- \exp(1.5-x_i))$. As the reader would note in the previous simulation scenario, the MSE of the target domain CRF and SRF also increased with increasing discrepancy. This is because the target domain functions, which are different from the corresponding source domain functions, generally have a different scale, making the scale of the response higher. Consequently, the scale of MSE was also going higher with increasing discrepancy. In this new setup, we define the difference function so that the scale of the response does not increase with changing discrepancy. The results presented in Figure \ref{tl_discrepancy ratio_new} show that transfer learning outperforms target-only RF for both CRF and SRF under these simulation settings as well.

\section{eICU data application}

We apply our new transfer learning methods to a large-scale dataset on ICU patient admissions available from the eICU database \cite{pollard2018eicu}. The database contains detailed data on more than 200,000 patients admitted to Intensive Care Units (ICU) from 208 hospitals and medical centers in the United States. This is a comprehensive database covering hospitals from all regions of the United States and of different capacities (beds).  In particular, the database contains 200,859 stays across 335 ICU units in 208 hospitals, by 139,367 unique patients, between 2014 and 2015.

We consider the response variable as whether the patient died in the hospital. We follow the preprocessing pipelines and the predictors used in the previous benchmarking studies by \cite{sheikhalishahi2020benchmarking,zhao2023improving}. This allows us to compare the performance of our transfer learning CRF and SRF methods with not only the target CRF and SRF methods, but also other methods considered in the above papers. The features we use for our analysis are provided in Table \ref{tab:variables_list}, and are the same as in \cite{sheikhalishahi2020benchmarking}. These include numerical features such as Heart rate, Mean arterial pressure, Diastolic and Systolic blood pressures, Oxygen saturation level, Respiratory rate, body temperature, glucose level, FiO2\footnote{Fraction of Inspired Oxygen}, pH, height, weight, age, bmi, and categorical features such as admission diagnostics, gender, and Glasgow Coma Scores. These features are extracted from tables related to patient information, laboratory measurements, nurse charting, and diagnosis.
%After creating dummy variables to represent various categories of the categorical variables, we end up with 20 features. 

\begin{table}[htbp]
\centering
\caption{List of features}
\label{tab:variables_list}
\begin{tabular}{|l|p{10cm}|}
\hline
\textbf{Feature Group} & \textbf{Variables} \\ \hline
\textbf{Quantitative} & Heart rate, Mean arterial pressure, Diastolic blood pressure, Systolic blood pressure, O2, Respiratory rate, Temperature, Glucose, FiO2, pH, Height, Weight, Age, Bmi \\
 & \\ \hline
\textbf{Categorical} & Admission diagnosis, Ethnicity, Gender, Glasgow Coma Score Total, Glasgow Coma Score Eyes, Glasgow Coma Score Motor, Glasgow Coma Score Verbal \\
& \\ \hline
\end{tabular}
\end{table}

\subsection{Data preprocessing} We also followed \cite{sheikhalishahi2020benchmarking}'s preprocessing pipeline to clean the data. Starting from 200,859 ICU stays, patients younger than 18 or elder than 89 are excluded. Besides, patients with more than one ICU stay, and an abnormal number of records (more than 200 or fewer than 15) are dropped. Then, patients without gender or discharge status are also removed. Finally, only patients who stay in the ICU longer than 24 hours are retained.

Since various health monitoring measurements vary dynamically, we further refined the timeline of the training data on the ICU stays used for prediction following \cite{zhao2023improving}. 
\begin{itemize}
    \item 7-day window slice: we selected observations within a 7-day window ending at the moment of discharge (or death) from ICU. If the patients stay fewer than 7 days, the selection begins at the moment of admission to ICU.
    \item After the 7-day window start time, the clinical records within the first 24 hours of the 7-day window are extracted. 
\end{itemize}

Our selection of a 24-hour window 7 days ahead of response time allows the model to both capture the critical information needed to make the prediction, yet make the predictions early enough to remain useful for practitioners and decision makers. The authors in \cite{zhao2023improving} found that prediction accuracy, measured by out-of-sample AUC, decreases with the number of days ahead for which we want to predict mortality. With their XGBoost algorithm, they found the AUC of 0.84 using data from a 24-hour window, 7 days ahead of response.  After this preprocessing, we obtain 1,108,586 records corresponding to 55,965 patients. Since some of these features are time series observations of patients over 24 hours, we take the average of those dynamic features as a representation of status across those 24 hours. 

\subsection{Dealing with categorical variables}  For the binary categorical variables Ethnicity and Gender, we use the one-hot encoding or dummy variables to create the features for the random forest method. Although the Glasgow Coma Score components are ordinal variables, their time-averaged values are treated as continuous in the application. A major challenge is the Admission diagnosis variable. This corresponds to 385 categories, and when treated with one-hot encoding or dummy variables, the feature dimensionality will be quite high. Therefore, here we use a deep neural network-based embedding to reduce the dimension of this categorical feature in a supervised fashion. We use an embedding layer combined with an encoder and a decoder model to represent this multi-category variable using 8-dimensional vectors. Figure 9 in the Appendix graphically illustrates the method. In particular, the admission diagnosis is passed through an embedding layer and a sigmoid activation function to map into a 8-dimensional vector scaled between (0,1) for each dimension. Then a sequence of encoder and decoder neural networks uses these embedding vectors and other observed features to predict each of the observed features. The estimated embedding vectors, along with the observed features, are now our new set of predictors. Consequently, the final dimensions of our predictors are 33.

\subsection{Tuning the algorithms to boost performance}

Since the eICU data are large-scale and heterogeneous, we make some adjustments (tuning) to our method implementation to improve performance. The residual model is trained on the target data with a small sample size. Since half the training data from target are used to compute the empirical distance covariance and the other half to train, we observe greater variance in both the empirical distance covariance and the search for optimal depths.

\textit{Stability of Distance covariance (SRF and CRF):}
In the transfer learning algorithm of both CRF and SRF, we change the way to calculate the empirical distance covariance. In the half of the training data used to calculate empirical distance covariance, we bootstrap the data to create 10 bootstrap samples and take the average of distance covariance in those bootstrapped samples.
 
\textit{Depth of random forest (CRF):} We change the estimation of optimal depth for both small and large samples. 
If the sample size is too small (<500), we do the regularized Out-of-Bag grid search as follows,
\begin{itemize}
    \item Run multiple iterations to create the bootstrap samples and validation samples (Out-of-Bag samples)
    \item Calculate the average ${\rm 1-AUC}$ for each depth candidate.
    \item The metric for the optimal depth is 
    \[
    {\rm 1-AUC} +\lambda. {\rm depth}
    \]
    with $\lambda = 0.01$. The penalty added to $1-AUC$ efficiently reduces the risk of overfitting
\end{itemize}
If the sample size is medium (smaller or equal to 2000 but larger than 500), we directly use 3-fold cross validation to choose the optimal split with the lowest ${\rm 1-AUC}$ as in the simulations. In both cases, depths candidates are chosen as: $\left\{h\leq \log_2 n: h = 2k+1, k=0,1,\cdots \right\}$ where $n\in \left\{n_s, n_t^{(1)}, n_t^{(2)}\right\}$

When the sample size is larger than 2000, we use an approximation method. Since from \ref{cormain}, we know 
$
{\rm depth_{{\rm optimal}}} = \log_2 k_{n} \approx a \log_2n + b,
$
therefore, the optimal depth is approximately a linear function of $\log_2n$. 
Given that observation, we only need to estimate $a$ and $b$ to obtain optimal depth. We randomly choose  datasets of different sizes from the source data, use these chosen data to do cross-validation, and find the optimal depth. Then we use linear regression between optimal depth and $\log_2 n$ with the data points to estimate $a$ and $b$. This approach saves on computing cost needed to do a grid search with cross validation that can be quite computationally intensive for large datasets.

\textit{Terminal node size:} In the centered random forest, to align with the algorithm as much as we can, we fixed the terminal node size at 2. In contrast, in the standard random forest, the split is data-driven and overfitting is highly likely to happen. To reduce the effect of overfitting, the terminal node size in the source model will be prespecified at 50. For the target model, the terminal node size is set to be proportional to the initial target total observations (before train-test split) with a lower bound 5. This ratio is tuned with grid search. In all cases, we grow 100 trees in the forest.

\begin{figure}[h]
    \centering
    
    \begin{subfigure}{0.5\textwidth}
\includegraphics[width=\textwidth]{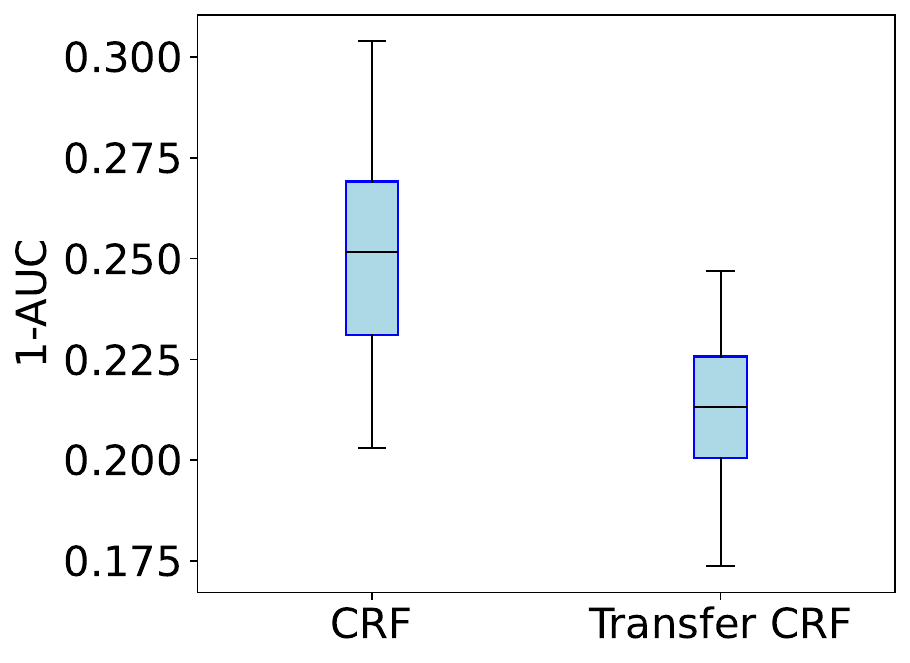}   
    \end{subfigure}%  
       \begin{subfigure}{0.5\textwidth}
  \includegraphics[width=\textwidth]{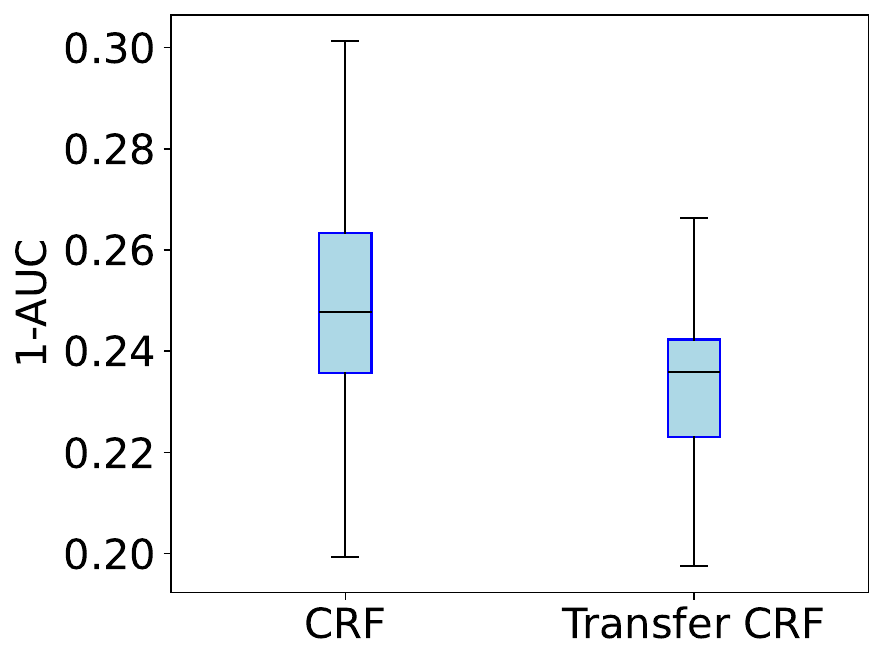}      
    \end{subfigure}
    \caption{Performance of TLCRF on Target 1 (left) and Target 2 (right) in terms of 1-AUC value. Both target data sets share the same settings: $30\%$ of target data is reserved for test, while the remaining is used in training. For the transfer learning procedure, $20\%$ of the data is used to calculate the empirical distance covariance, while the model is trained on the remaining portion.  
    }
    \label{fig:eICUTL}
\end{figure}

\subsection{Source and target data}

We consider large hospitals as source data and small hospitals as target data. We define the large and small hospitals based on the number of beds. The source data comprises hospitals with more than 250 beds and more than 150 recorded patients. We construct two target data sets. One cohort contains the hospitals with fewer than 100 beds but having at least 10 recorded patients. It also includes the hospitals with an unknown number of beds, but is limited to those with fewer than 100 recorded patients. The other target data set contains hospitals with beds from 100 to 249 and at least 10 recorded patients. 
The numbers of patients for source data and the two target datasets are 35737, 3428 (hospitals with fewer than 100 beds), and 10787 (hospitals with 100-249 beds), respectively. Therefore, the source dataset is about 10 times bigger than target dataset 1 and about 3.5 times bigger than the target dataset 2. This construction of source and target datasets makes sense since our motivation is to use rich datasets from large hospitals to improve prediction in small hospitals with fewer records. While we will use the target hospitals in aggregate, in the Appendix, we also describe prediction at each hospital within the target datasets.

After the split, we apply the 8-dimensional embedding to the feature admission diagnosis. The embedding model is trained on the source data and gets a mapping between each category from admission diagnosis and 8-dimensional vectors. Then, the admission diagnosis in the two target data sets is transformed based on the source mapping. If a new category is observed in target data sets, the global mean vector will be applied.

\textit{Averaging and aggregating performance metric over randomness:} We compare the performance of the target-only random forest and the transfer learning random forest methods by averaging the metric across multiple random trials, considering the randomness in the train-test split and the construction of the random forest. We conducted train-test splits 50 times with different random seeds, and for each train-test split, random forest and transfer learning algorithm are trained 5 times with different random seeds. The five 1-AUC for each train and test data are averaged and will serve as the metric to measure the performance of models. We will display our results as boxplots of the metric over these 50 random train-test splits.

\begin{figure}[h]
    \centering   \includegraphics[width=0.8\textwidth]{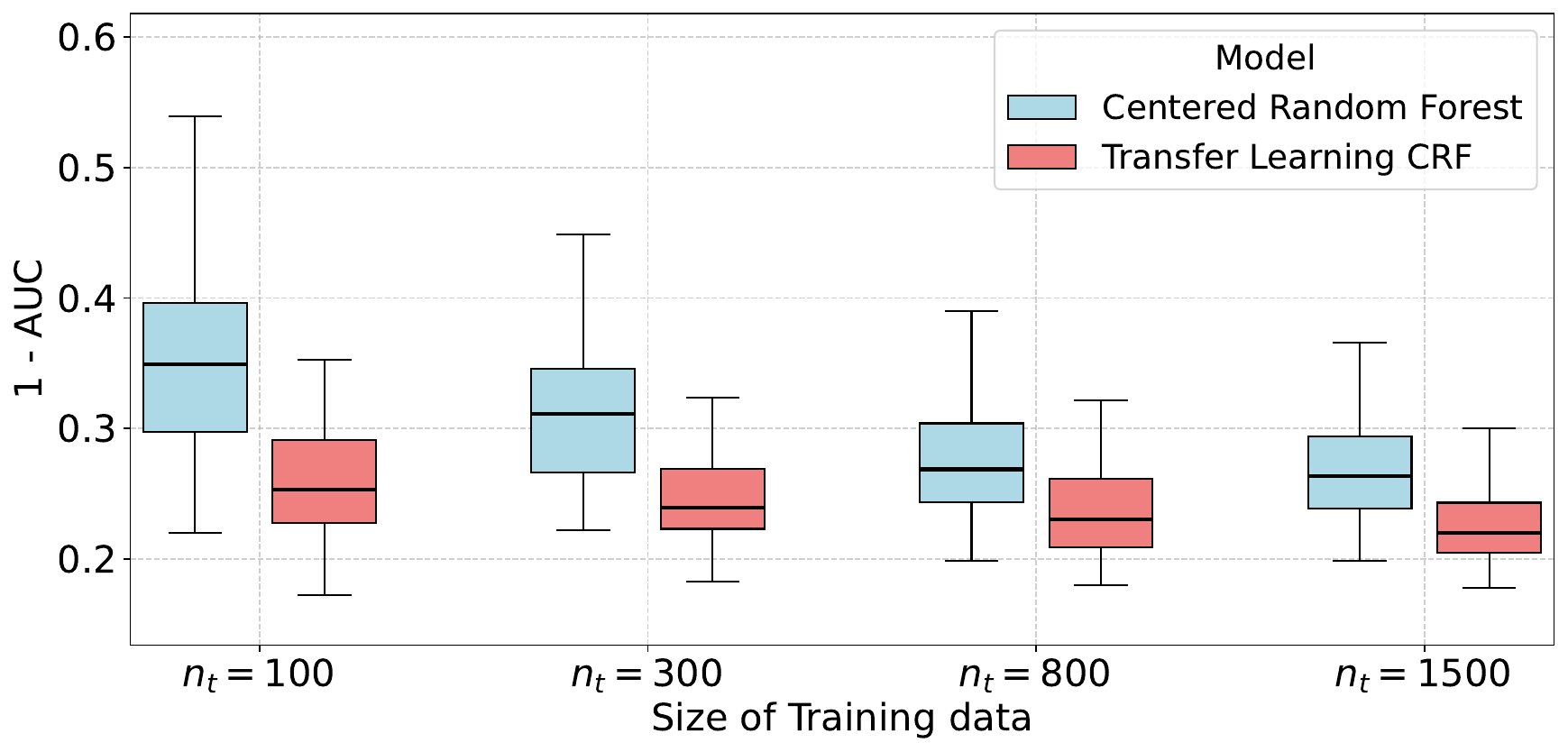}    
    \caption{Performance of TLCRF vs CRF in terms of 1-AUC values with varying size of the training data availability from the Target dataset 1. The test size is 900. The test data set is fixed to be the same for all sizes of training data. }
    \label{fig:eICUTL2CRF}
\end{figure}

\begin{figure}[h]
    \centering   \includegraphics[width=0.8\textwidth]{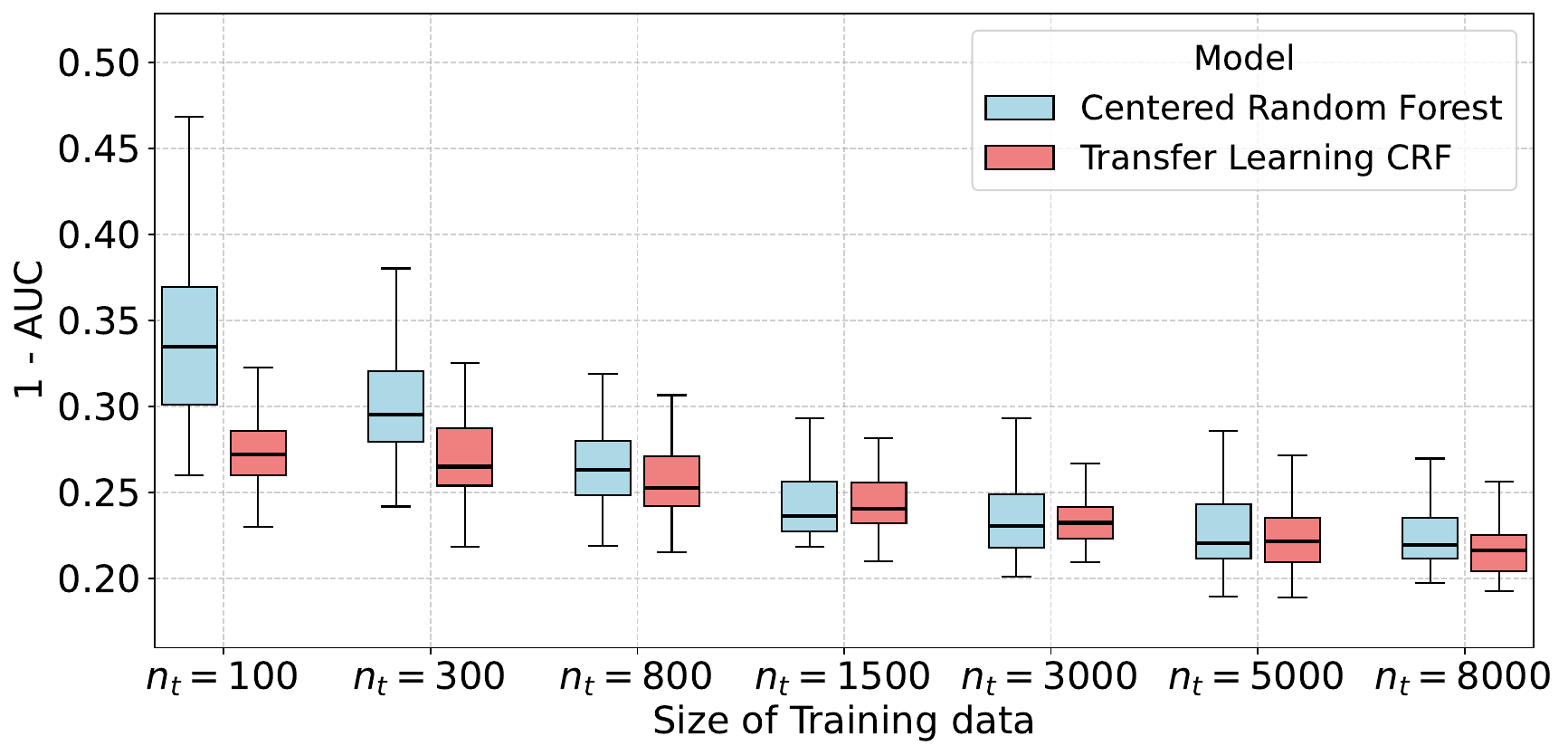}    
    \caption{Performance of TLCRF vs CRF in terms of 1-AUC value with varying size of the training data availability from Target 2. The test size is $20\%$ of Target 2 data (roughly 2000), while the training data is varied from 100 to 8000.}
    \label{fig:eICUTL2CRF2}
\end{figure}

\begin{figure}[h]
    \centering
\includegraphics[width=0.8\textwidth]{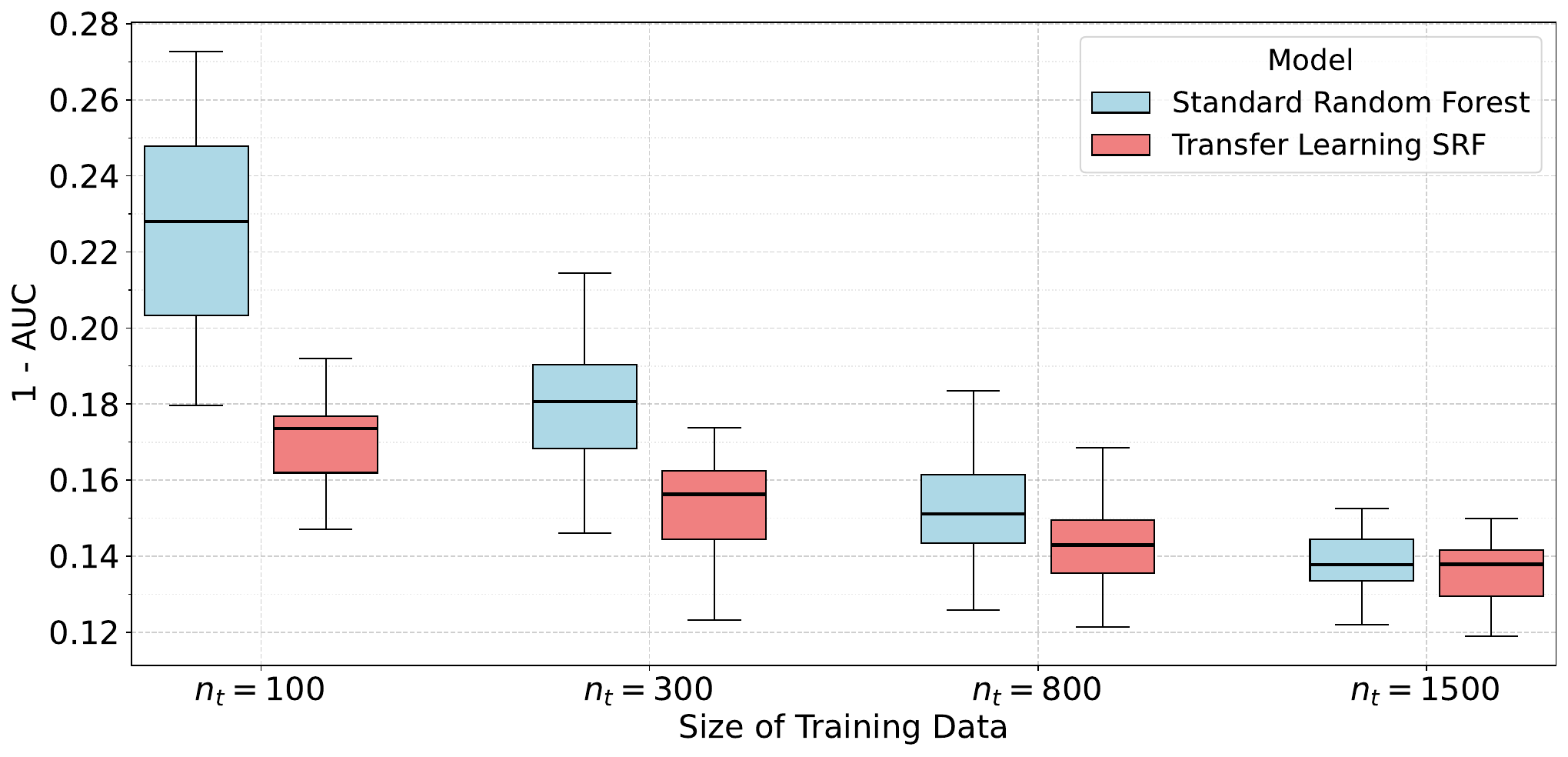} 
    \caption{Performance of  TLSRF vs SRF in terms of 1-AUC value with varying size of the training data availability from target 1. The test size is 900 and the same across all sizes of training data. Source model:  maximum depth: $\lfloor \log_2 n_s \rfloor + 1$; bootstrap size: $\lfloor 0.8\times n_s \rfloor$; terminal node size: 50; number of features to be selected: $\lfloor \sqrt{d} \rfloor$; Target models: SRF or residual model from TLSRF: number of trees: 100; maximum depth:\(\lfloor \log_2 (\lfloor 0.2 \cdot n_t \rfloor) \rfloor + 1\); bootstrap size: $\lfloor 0.8 n_t\rfloor$; terminal node size:$\max \left( 5, \left\lfloor \frac{n_t \times 0.02}{0.7} \right\rfloor \right)$.}
    \label{fig:eICUTL2SRF}
\end{figure}

\begin{figure}[h]
    \centering
\includegraphics[width=0.8\textwidth]{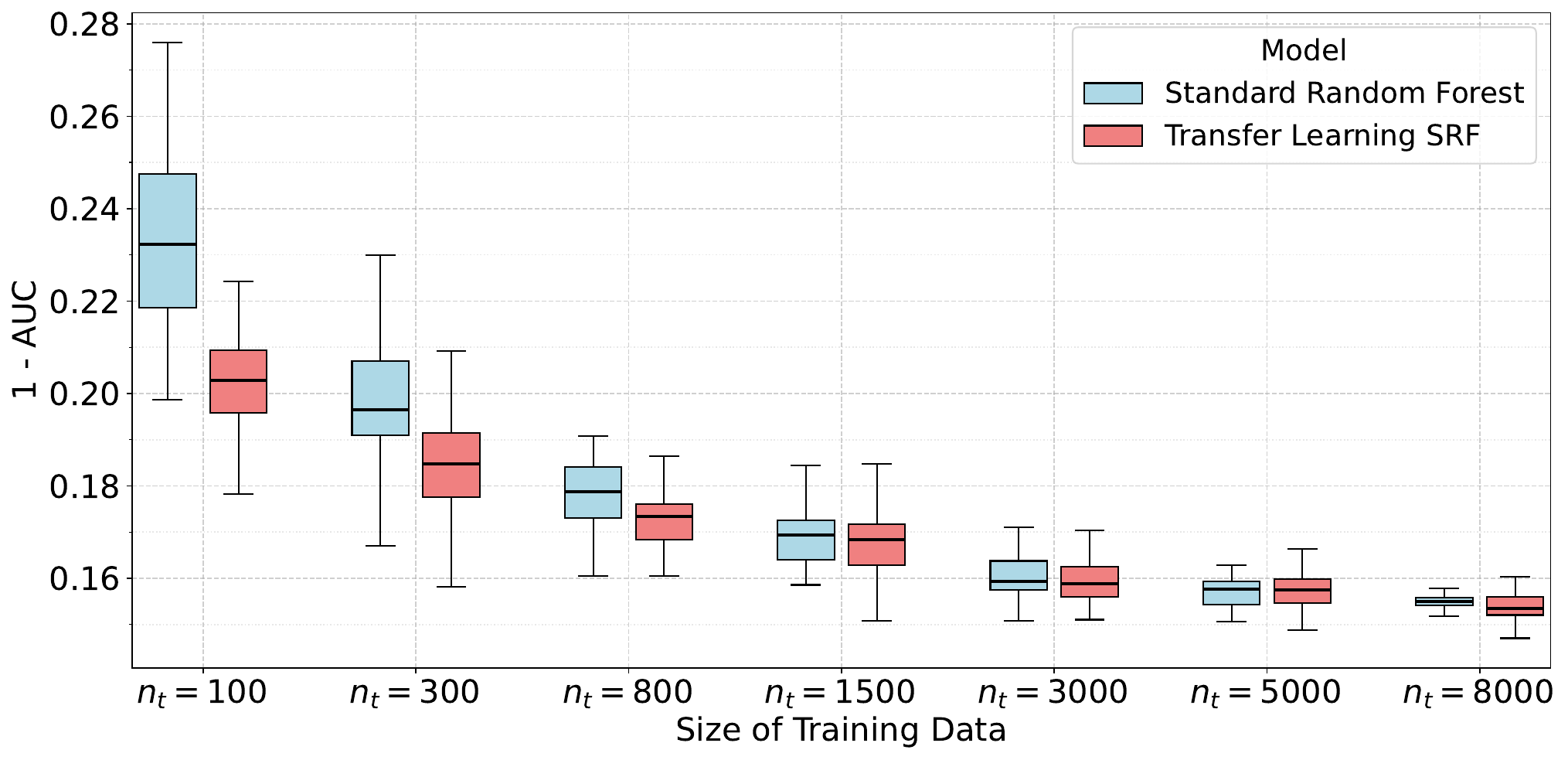} 
\caption{Performance of  TLSRF vs SRF in terms of 1-AUC value with varying size of the training data availability from target 2. The test size is $20\%$ of Target 2 data. The settings is the same as the previous simulation.}
\label{fig:eICUTL3SRF}
\end{figure}

\subsection{Results}
To thoroughly investigate the performance of transfer learning for predictive accuracy, we consider two setups: (1) Assume we have all the source data and 70\% of the target data available to train and calibrate models. The remaining 30\% of the target data is the test set, (2) Continue to assume all the source data is available, but the amount of target data available is varied in steps. We select 900 observations as the test set and  select $\left\{ 100, 300, 800, 1500\right\}$ observations as the training set from the target data 1. The test set is kept fixed when training sets in the target data vary. For target data 2, we randomly select 20\% of the observations or about 2000 observations as the test set and select $\left\{ 100, 300, 800, 1500, 3000, 5000, 8000 \right\}$ observations as the training set. These two data availability assumptions are denoted as scenario (1) and scenario (2).

The results from Scenario (1) is presented in Figure \ref{fig:eICUTL}. As we can see, for both target datasets, the proposed TLCRF method outperforms the  CRF, which is trained just on the target data. This shows that our transfer learning method is quite beneficial for predicting ICU outcomes in smaller hospitals leveraging data from larger hospitals. The improvement of performance is more for the target dataset 1, which consists of smaller hospitals (less than 100 beds).

The results from scenario (2) is presented in Figures \ref{fig:eICUTL2CRF} and \ref{fig:eICUTL2CRF2}. From both figures, we generally see the benefit of transfer learning for different sample sizes of the training data. With more training data from the target domain available, the performance of all methods improves, but the transfer learning methods continue to outperform their non-transfer counterparts. We make similar observations for the standard random forest in Figures \ref{fig:eICUTL2SRF} and \ref{fig:eICUTL3SRF}. In all four figures (Figures \ref{fig:eICUTL2CRF} and \ref{fig:eICUTL2CRF2} for CRF and Figures \ref{fig:eICUTL2SRF} and \ref{fig:eICUTL3SRF} for SRF), we see substantial benefit for transfer learning when the target training data is much smaller compared to both the complexity of the model and the source data size.

In the Appendix, we also investigate the utility of transfer learning for predictions at individual hospitals within these two target datasets. These results are noisier, and we do not see a clear trend. Generally the TLCRF method performs better than the CRF method in most of the hopsitals, but not all. Further, there is a large variation in the performance of both methods across hospitals. This is because due to heterogeneity among the hospitals, there is an imbalance in the covariates in the hospitals, especially for those with a small number of records.

Finally, we also note that with sufficient training data, the performance of our standard RF and transfer standard RF implementations often exceeds 0.86 (Figure \ref{fig:eICUTL3SRF}) and 0.85 (Figure \ref{fig:eICUTL2SRF}) in median AUC. This performance is substantially better than specialized time-series-based deep learning methods and other baselines employed in the benchmarking study of \cite{sheikhalishahi2020benchmarking} (We note the caveat that while \cite{sheikhalishahi2020benchmarking} used the first 24 hours of information from admission time, we used 24 hours of data 7 days prior to the response event, which may make our prediction task easier since patients may stay more than 7 days in the ICU). While our sample does not exactly match (and our predictors are different) with \cite{zhao2023improving}, we also note that our AUC values are higher than their XGBoost method's AUC of 0.84 for 7-day-ahead prediction. Therefore, our implementation of random forest is one of the best methods for this dataset. This case study clearly establishes the benefits of transfer learning, especially in small target datasets, even when a state-of-the-art method is employed.

\section{Conclusion}
In this article, we developed a transfer learning procedure for random forest using distance covariance to weigh features. We theoretically studied the procedure's accuracy in terms of mean squared error. Our results show that the accuracy of transfer learning depends on sample sizes in source and target domains and the sparsity of the difference function. In particular, our results show the benefit of transfer learning when the source domain sample size is much higher than the target domain sample size and the difference function only involves a subset of the features. We empirically verified the performance of our method through extensive simulations and an application to a multi-hospital EHR dataset. Together, our results expand the scope of transfer learning methodology with statistical error guarantees to include methods in modern machine learning, such as the random forest.

\begin{acks}[Acknowledgments]
The authors would like to thank the poster session participants at the Department of Statistics at OSU and at the ICSA Midwest conference for their feedback.
\end{acks}

%%%%%%%%%%%%%%%%%%%%%%%%%%%%%%%%%%%%%%%%%%%%%%
%% Funding information, if any,             %%
%% should be provided in the                %%
%% funding section.                         %%
%%%%%%%%%%%%%%%%%%%%%%%%%%%%%%%%%%%%%%%%%%%%%%
\begin{funding}
A grant from the NSF Mathematics of Digital Twins program (DMS 2529302) partially supported this research.
\end{funding}

%%%%%%%%%%%%%%%%%%%%%%%%%%%%%%%%%%%%%%%%%%%%%%
%% Supplementary Material, including data   %%
%% sets and code, should be provided in     %%
%% {supplement} environment with title      %%
%% and short description. It cannot be      %%
%% available exclusively as external link.  %%
%% All Supplementary Material must be       %%
%% available to the reader on Project       %%
%% Euclid with the published article.       %%
%%%%%%%%%%%%%%%%%%%%%%%%%%%%%%%%%%%%%%%%%%%%%%

\newpage 
\begin{supplement}
\stitle{Supplement to ``Transfer Learning with Distance Covariance for Random Forest: Error Bounds and an EHR Application''}
\sdescription{The supplementary file contains proofs of all lemmas, theorems and the corollary, some additional propositions, and some additional figures from the simulation and eICU application section}
\end{supplement}

\section{Proof of theoretical results}

\subsection{Additional lemmas and propositions}

We begin with the following proposition regarding population distance covariance.
\begin{prop}
\label{dcov}
     Let $\mathcal{V}^2\left(X, Y\right)$ denote the distance covariance between $X$ and $Y$. Suppose $X_1, X_2, Y \in \mathbb{R}$ are random variables with finite first moments. If the random vector $(X_1, Y_1)$ is independent of the random variable $X_2$, then
\begin{align}
    \mathcal{V}^2(X_1 + X_2, Y_1) \leq \mathcal{V}^2(X_1, Y_1).
\end{align}
\end{prop}

%Proof of Proposition \ref{dcov}

\begin{proof}

Let $\varphi_{X,Y}(t,s)=\mathbb E[e^{i(t X+s Y)}]$ denote the joint characteristic function and
$\varphi_X(t)=\mathbb E[e^{it X}]$ the marginal characteristic function of $X$. By definition,
\[
\mathcal V^2(X,Y)=\int
\bigl|\varphi_{X,Y}(t,s)-\varphi_X(t)\varphi_Y(s)\bigr|^2\,w(t,s)\,dt\,ds,
\]
with a suitable weight function $w$ designed for distance covariance \cite{szekely2007measuring}.
Because of independence between $X_2$ and $(X_1,Y)$, we have
\[
\varphi_{X_1+X_2,Y}(t,s)=\varphi_{X_1,Y}(t,s)\varphi_{X_2}(t),
\qquad
\varphi_{X_1+X_2}(t)=\varphi_{X_1}(t)\varphi_{X_2}(t).
\]
Hence
\[
\varphi_{X_1+X_2,Y}(t,s)-\varphi_{X_1+X_2}(t)\varphi_Y(s)
=\varphi_{X_2}(t)\bigl(\varphi_{X_1,Y}(t,s)-\varphi_{X_1}(t)\varphi_Y(s)\bigr).
\]
Therefore, using $|\varphi_{X_2}(t)|\le 1$ for all $t$,
\begin{align*}
\mathcal V^2(X_1+X_2,Y)
&=\int \Bigl|\varphi_{X_1+X_2,Y}(t,s)-\varphi_{X_1+X_2}(t)\varphi_Y(s) \Bigr|^2 w(t,s)\,dt\,ds \\
& = \int \Bigl| \varphi_{X_2}(t)\bigl(\varphi_{X_1,Y}(t,s)-\varphi_{X_1}(t)\varphi_Y(s)\bigr)\Bigr|^2 w(t,s)\,dt\,ds \\
&\le \int \bigl|\varphi_{X_1,Y}(t,s)-\varphi_{X_1}(t)\varphi_Y(s)\bigr|^2 w(t,s)\,dt\,ds \\
&=\mathcal V^2(X_1,Y).
\end{align*}
\end{proof}

The following proposition is an adaptation of the result in \cite{li2012feature}. 
\begin{prop}
\label{distance}
    Let $\mathbf{X} = \left(X_1, \cdots, X_d\right)^T$ follows the uniform distribution within $\left[0, 1\right]^d$ and $Y = f(\mathbf{X})+\xi$ where $\xi$ follows sub-Gaussian distribution and independent with $\mathbf{X}$ and $f$ is a Lipschitz function. Suppose $\omega_k$ is the distance covariance between $X_k$ and $Y$ and $\hat{\omega}_k$ is the sample (empirical) version of $\omega_k$.  Suppose $\mathcal{I}$ is the set of inactive features and $\mathcal{I}^C$ is the set of active features. Let $\min_{k \in \mathcal{I}^C} \omega_k \geq 2 c n^{-\alpha}$ for $c>0$ and $0 \leq \alpha<1/2$. Let $\epsilon_n = cn^{-\alpha}$. For any $0 < \eta < 1/2-\alpha$ , there exist positive constants $c_1 > 0$ and $c_2 > 0$, such that for a given $k$,
    $$
    \mathbb{P}\left\{ |\hat{\omega}_k - \omega_k|\geq \epsilon_n\right\} = O\left([\exp(-c_1\epsilon_n^2n^{1-2\eta})+n \exp(-c_2n^{\eta})]\right).
    $$
\end{prop}
We note that we stated the above proposition for a particular covariate $k$. We can make the statement uniform by multiplying a factor of $d$ in the right-hand side.

%\subsubsection{Proof of Proposition \ref{distance}}

\begin{proof}
    Note that since $X_k \in [0,1]$, then $E[\exp(s|X_k|^2)]<\infty$. Further, $\mathbf{X}$ is bounded (in $[0,1]^d$), and $f$ is a function bounded by a constant that does not depend on dimension $d$. Since $Y$ is then summation of a sub-Gaussian distribution (i.e., $E[\exp(s\xi^2)]<\infty$) and a bounded function, we have $E[\exp(sY^2)] < \infty$ for $s>0$. Therefore they satisfy the definition of sub-exponential tail probability uniformly in $d$. Then the conditions of Theorem 1 in the paper \cite{li2012feature} is satisfied and we can use the result.
\end{proof}

We prove the following lemma that shows the residual random variable obtained by predicting the target response using the source model on the target covariates is sub-exponential.
\begin{lemma}
   The residual random variable $\tilde{Y}_t(X_t, D_s)$ follows sub-Gaussian distribution.
   \label{subgaussian}
\end{lemma}

%\subsubsection{Proof of Lemma \ref{subgaussian}}
\begin{proof}
    Recall $\tilde{Y}_t(X_t, D_s) = Y_t - \hat{Y}(X_t, \mathcal{D}_s)$. If we can show $\hat{Y}(X_t, \mathcal{D}_s)$ is sub-Gaussian, then $\tilde{Y}_t(X_t, D_s)$ is sub-Gaussian since $Y_t$ is sub-Gaussian by the arguments in Proposition \ref{distance}. Therefore, we are going to show that $\hat{Y}(X_t, \mathcal{D}_s)$ is sub-Gaussian. We start from the moment generating function of $\hat{Y}(X_t, \mathcal{D}_s) -  \mathbb{E}[\hat{Y}(X_t, \mathcal{D}_s)]$. At first, for any $\lambda$, we calculate
\begin{equation}
     \mathbb{E} \left[ \exp\left( \lambda \left( \hat{Y}(X_t, \mathcal{D}_s) - \mathbb{E}[\hat{Y}(X_t, \mathcal{D}_s)] \right) \right) \right]
\end{equation}
Since $\hat{Y}(X_t, \mathcal{D}_s) - \mathbb{E}[\hat{Y}(X_t, \mathcal{D}_s)]$ can be decomposed into two parts, that is,
\begin{equation}
    \hat{Y}(X_t, \mathcal{D}_s) - \mathbb{E}[\hat{Y}(X_t, \mathcal{D}_s)] = Z_f + Z_\epsilon
\end{equation}
where 
\begin{align*}
    Z_f &= \sum_{i=1}^n \mathbb{E}_{\Theta}[W_i] f_s(X_i) - \mathbb{E}\left[ \sum_{i=1}^n \mathbb{E}_{\Theta}[W_i] f_s(X_i) \right] \\
    Z_\epsilon &= \sum_{i=1}^n \mathbb{E}_{\Theta}[W_i] \epsilon_i^{(s)} 
\end{align*}
For $Z_\epsilon$, 
\[
\mathbb{E}[Z_\epsilon] =  \sum_{i=1}^n \mathbb{E}\left[\mathbb{E}_{\Theta}[W_i] \epsilon_i^{(s)}\right] =  \sum_{i=1}^n \mathbb{E}\left[\mathbb{E}_{\Theta}[W_i]\right] \mathbb{E}\left[\epsilon_i^{(s)}\right] = 0
\]
By the Cauchy-Schwarz inequality, 
\[
 \mathbb{E} \left[ \exp\left( \lambda \left( \hat{Y}(X_t, \mathcal{D}_s) - \mathbb{E}[\hat{Y}(X_t, \mathcal{D}_s)] \right) \right) \right] =\mathbb{E} \left[ e^{\lambda(Z_f + Z_\epsilon)} \right] \leq \sqrt{\mathbb{E} \left[ e^{2\lambda Z_f} \right] \cdot \mathbb{E} \left[ e^{2\lambda Z_\epsilon} \right] }
\]
By the assumption $\|f_s\|_\infty \leq M_s$ as well as $\sum_{i=1}^n\mathbb{E}_{\Theta}[W_i]=1, \mathbb{E}_{\Theta}[W_i]\geq 0$,
\[\left|\sum_{i=1}^n\mathbb{E}_{\Theta}[W_i]f_s(X_i)\right|\leq M_s\]
Then from Hoeffding's lemma we have,
\begin{equation}
    \mathbb{E} \left[ \exp(2\lambda Z_f) \right] \leq \exp\left( \frac{(2\lambda)^2 (2M_s)^2}{8} \right) = \exp\left( 2\lambda^2 M_s^2 \right)
\end{equation}
We use the conditional expectation to analyze $\mathbb{E} \left[ e^{2\lambda Z_\epsilon} \right]$. We notice that $\mathbb{E}_{\Theta}[W_i]$ depends only on $X_{1,s},\ldots X_{n_s,s}$ and $X_t$ and therefore is independent of $\epsilon^{(s)}$,
\[
 \mathbb{E} \left[ \exp(2\lambda Z_\epsilon) \right] = \mathbb{E}_{D_s,X_t} \left[ \mathbb{E}_{\epsilon^{(s)}|D_s,X_t} \left[ \exp\left( \sum_{i=1}^n 2\lambda \mathbb{E}_{\Theta}[W_i] \epsilon_i^{(s)} \right) \right] \right]
\]
where $\mathbb{E}_{\epsilon^{(s)}|D_s,X_t}$ denotes taking the expectation with respect to $\epsilon^{(s)}$ conditional on $D_s,X_t$.
With the independence and sub-Gaussianity of $\epsilon^{(s)}$,
\begin{align*}
    \mathbb{E}_{\epsilon^{(s)}|D_s,X_t} \left[ \exp\left( \sum_{i=1}^n 2\lambda \mathbb{E}_{\Theta}[W_i] \epsilon_i^{(s)} \right) \right] &= \prod_{i=1}^n \mathbb{E}_{\epsilon_i^{(s)}|D_s,X_t} \left[ \exp\left( 2\lambda \mathbb{E}_{\Theta}[W_i] \epsilon_i^{(s)} \right) \right] \\
    &\leq \prod_{i=1}^n \exp\left( \frac{(2\lambda \mathbb{E}_{\Theta}[W_i])^2 \sigma_s^2}{2} \right) \\
    &= \exp\left( 2\lambda^2 \sigma^2 \sum_{i=1}^n (\mathbb{E}_{\Theta}[W_i])^2 \right)
\end{align*}
Combined with the fact that
\[
\sum_{i=1}^n (\mathbb{E}_{\Theta}[W_i])^2 \le \sum_{i=1}^n \mathbb{E}_{\Theta}[W_i] = 1,
\]
we have
\[
\mathbb{E} \left[ \exp(2\lambda Z_\epsilon) \right] \leq \mathbb{E}_{D_s,X_t} \left[ \exp(2\lambda^2 \sigma^2) \right] = \exp(2\lambda^2 \sigma^2)
\]
In summary, combining two terms of $Z_f$ and $Z_\epsilon$:
\begin{align*}
   \mathbb{E} \left[ \exp\left( \lambda \left( \hat{Y}(X_t, \mathcal{D}_s) - \mathbb{E}[\hat{Y}(X_t, \mathcal{D}_s)] \right) \right) \right] &\leq \left( \exp(2\lambda^2 M_s^2) \right)^{1/2} \cdot \left( \exp(2\lambda^2 \sigma^2) \right)^{1/2} \\
    &= \exp(\lambda^2 M_s^2) \cdot \exp(\lambda^2 \sigma^2) \\
    &= \exp\left( \lambda^2 (M_s^2 + \sigma^2) \right)
\end{align*}
This shows that $\hat{Y}(X_t, \mathcal{D}_s)$ is sub-Gaussian. 
\end{proof}

\subsection{Proof of Lemma 3.2}

\begin{proof}
    The proof makes an adjustment to the proof of a lemma in \cite{klusowski2021sharp}. Let 
$\mu(\mathbf{x})
=\binom{n}{x_1,\dots,x_k}\,(p_{n1})^{x_1}\cdots (p_{nk})^{x_k}$
denote the  PMF of the multinomial distribution and consider a mode \(\mathbf{x}^*\) of the distribution. Then \cite{klusowski2021sharp} have shown that 
$$
\mathbb{E}\bigl[2^{-\tfrac12\sum_{j=1}^k\lvert X_j - X'_j\rvert}\bigr]\leq (4 + 2\sqrt{2})^{\,k-1}\,\mu(\mathbf{x}^*).
$$
and
\begin{equation}
\label{S1}
    \mu(\mathbf{x}^*)
\le 
\frac{e^{k+1}}{(\sqrt{2\pi})^{\,k-1}}
\sqrt{\frac{n}{(x_1^*+1)\cdots(x_k^*+1)}}
\biggl(\frac{n p_{n1}}{x_1^* + 1}\biggr)^{x_1^*}
\cdots
\biggl(\frac{n p_{nk}}{x_k^* + 1}\biggr)^{x_k^*}\,.
\end{equation}
Then, (\cite{feller1968introduction}, page 171, Exercise 28, Equation 10.1) 
states that any mode \(\mathbf{x}^*\) of the multinomial distribution satisfies $np_{nj} -1 \leq x_j^* \leq  \left(n+k-1\right)\,p_{nj}$. This means that since $x_j^*$ is an integer, when $np_{nj}$ is small enough, $x_j^*=0$
and hence from (\ref{S1}),
\begin{equation}
\label{S2}
    \mu(\mathbf{x}^*)
\le 
\frac{e^{k+1}}{(\sqrt{2\pi})^{\,k-1}}
\sqrt{\frac{n}{(x_1^*+1)\cdots(x_{k_1}^*+1)}}
\biggl(\frac{n p_{n1}}{x_1^* + 1}\biggr)^{x_1^*}
\cdots
\biggl(\frac{n p_{nk_1}}{x_{k_1}^* + 1}\biggr)^{x_{k_1}^*}\,.
\end{equation}
Hence,
$$\mathbb{E}\bigl[2^{-\tfrac12\sum_{j=1}^k\lvert X_j - X'_j\rvert}\bigr]< \frac{8^k}{\sqrt{n^{k_1-1} p_{n1} \cdots p_{nk_1}}}$$
\end{proof}

\subsection{Proof of Theorem 3.3}

\begin{proof}
We have the bias-variance decomposition inequality
\begin{align}
\label{mainineq}
    \mathbb{E}\left[(\hat{f}_{Y}(X; D) - f(X))^2\right]\leq \mathbb{E}\left[( \bar{f}_{Y}(X; D) - f(\mathbf{X}))^2 \right] +  \mathbb{E} \left[ ( \hat{f}_{Y}(X; D) -  \bar{f}_{Y}(X; D) )^2 \right]
\end{align}

We modify the proofs of Theorems 1 and 2 in \cite{klusowski2021sharp} to establish our result.

Recall that,
\[
\hat{f}_Y(X,D) = \sum_{i=1}^n \mathbb{E}_{\Theta} [\frac{1(X_i \in t(X)}{\sum_i 1(X_i \in t(X) } 1(\xi)] Y_i
\]
Here $X$ is a new data point (a draw from the distribution of predictors) and $X_1, \ldots, X_n \in D$. Let us denote $W_i = \frac{1(X_i \in t(X)}{\sum_i 1(X_i \in t(X) } 1(\xi)$. 
Then we define, 
\[
\bar{f}_Y(X,D) = \mathbb{E} [\hat{f}_Y(X,D)| X_1, \ldots, X_n,X] = \sum_{i=1}^n \mathbb{E}_{\Theta} [\frac{1(X_i \in t(X)}{\sum_i 1(X_i \in t(X) } 1(\xi)] f(X_i)
\]
Let $X_{i,S^C}$ denote the restriction of $X_i$ to the ``active set'' $S^C$. By assumption we have $f(X_i) = f^*(X_{i,S^C})$, and the function $f^{*}$ is $L$-Lipschitz as a function of $X_{S^C}$.

Then, following \cite{klusowski2021sharp}, we can write the first part of the above decomposition (that corresponds to the approximation error or bias term) as

\begin{align*}
& \mathbb{E}\left[(\bar{f}_Y(X,D)-f(\mathbf{X}))^{2}\right]  \\
& =\mathbb{E}\left[\left(\sum_{i=1}^{n} \mathbb{E}_{\Theta}\left[W_{i}\left(f\left(\mathbf{X}_{i}\right)-f(\mathbf{X})\right)\right]-\mathbf{1}\left(\mathcal{E}^{c}\right) f(\mathbf{X})\right)^{2}\right] \\
& =\mathbb{E}\left[\left(\sum_{i=1}^{n} \mathbb{E}_{\Theta}\left[W_{i}\left(f^*(X_{i,S^C})-f^*(X_{i,S^C})\right)\right]\right)^{2}\right]+\mathbb{E}\left[\mathbf{1}\left(\mathcal{E}^{c}\right)|f(\mathbf{X})|^{2}\right] \\
& \leq \mathbb{E}\left[\left(\sum_{i=1}^{n} \mathbb{E}_{\Theta}\left[W_{i}\left(f^*(X_{i,S^C})-f^*(X_{i,S^C})\right)\right]\right)^{2}\right]+M^{2} \mathbb{P}\left(\mathcal{E}^{c}\right)
\end{align*}

Now we compute
\[
W_i\left|f^*(X_{i,S^C})-f^*(X_{i,S^C})\right| \leq W_i\sum_{j=1}^{S^C}\left\|\partial_{j} f\right\|_{\infty}\left|X_{i,S^C}^j-X_{S^C}^j\right| \leq W_{i} \sum_{j=1}^{S^C}\left\|\partial_{j} f\right\|_{\infty}\left(b_{j}(\mathbf{X})-a_{j}(\mathbf{X})\right).
\]
The last inequality follows since the $j$th coordinate of $X$ is at most $b_j(X)-a_j(X)$ distance away from the $j$th coordinate of any $X_i \in D$ that falls in the leaf $t$.
Summing over all data points in data $D$, we have
\begin{align*}
\sum_{i=1}^nW_i\left|f^*(X_{i,S^C})-f^*(X_{i,S^C})\right| & \leq \sum_{i=1}^n W_{i} \sum_{j=1}^{S^C}\left\|\partial_{j} f\right\|_{\infty}\left(b_{j}(\mathbf{X})-a_{j}(\mathbf{X})\right) \\
& \leq \sum_{j=1}^{S^C}\left\|\partial_{j} f\right\|_{\infty}\left(b_{j}(\mathbf{X})-a_{j}(\mathbf{X})\right), 
\end{align*}
since $\sum_{i=1}^n W_i=1$.

Then we have
$$
\mathbb{E}\left[\left(\sum_{i=1}^{n} \mathbb{E}_{\Theta}\left[W_{i}\left(f^*(X_{i,S^C})-f^*(X_{i,S^C})\right)\right]\right)^{2}\right] \leq |S^C| \sum_{j=1}^{|S^C|}\left\|\partial_{j} f\right\|_{\infty}^{2} \mathbb{E}\left[\left(\mathbb{E}_{\Theta}\left[b_{j}(\mathbf{X})- a_{j}(\mathbf{X})\right]\right)^{2}\right]
$$

Since we split the $j$th feature $\log_2 k_n$ times with probability $p_j$, from the arguments in \cite{klusowski2021sharp} we have,

\begin{align*}
    \mathbb{E}\left[(\bar{f}_{Y}(X; D)  - f(X))^2 \right]\leq |\mathcal{S}^C| \sum_{j \in \mathcal{S}^C} \left\| \partial_j f \right\|_{\infty}^2 k_n^{2 \log_2 (1 - p_{nj} / 2)} + M^2 e^{-n/(2 k_n)}.
\end{align*}

For the second term, we have
\begin{align}
\label{estierror}
\mathbb{E} \left[ \left( \hat{f}_{Y}(X; D) - \bar{f}_{Y}(X; D) \right)^2 \right]
&\leq \frac{12\sigma^2 k_n^2}{n} \, \mathbb{E}_{\boldsymbol{\Theta}, \boldsymbol{\Theta'}} \left[ 2^{-\frac{1}{2} \sum_{j=1}^d |K_j - K_j'|} \right]
\end{align}
    where $(K_1, \ldots, K_d)$ is a random vector following the multinomial distribution with $\lceil \log_2 k_{n} \rceil$ trials and probabilities $(p_{nj})_{1 \leq j \leq d}$ given $X$. Using Lemma 3.2, we can upper bound $\mathbb{E}_{\boldsymbol{\Theta}, \boldsymbol{\Theta'}} \left[ 2^{-\frac{1}{2} \sum_{j=1}^d |K_j - K_j'|} \right]$ with $m=\lceil \log_2 k_{n} \rceil$. Since when $j \in \mathcal{S}_\alpha$, we have $p_{nj}\lceil \log_2 k_{n} \rceil\leq n^{-\alpha}\lceil \log_2 k_{n} \rceil\to 0$ as $n\to \infty$. Then using Lemma 3.2 in inequality \ref{estierror}, we would get inequality 
\begin{equation}
\label{6.4}
\mathbb{E} \left[ \left( \hat{f}_{Y}(X; D) - \bar{f}_{Y}(X; D) \right)^2 \right] \;\le\;
\frac{12\sigma^2 k_{n}\,8^{d}}{n\,\sqrt{\displaystyle\prod_{j\in\mathcal{S}_\alpha^C}p_{nj}\;\bigl(\log_2 k_{n}\bigr)^{|\mathcal{S}_\alpha^C|-1}}}\,.
\end{equation}
Combine these two together and we get the result.
\end{proof}

\subsection{Proof of Lemma 3.4}

   \begin{proof}
    Since $X_t^{(j)}$ is independent to $R(X_t)$, then 
    \begin{align*}
        \mathcal{V}^2\left(\tilde{Y}_t(X_t, D_s), X_t^{(j)}\right)
        &=\mathcal{V}^2\left(R(X_t) +\epsilon_t + \delta_s(X_t), X_t^{(j)}\right)\\
        &\leq \mathcal{V}^2\left(R(X_t)  + \delta_s(X_t), X_t^{(j)}\right)
    \end{align*}
    where $\delta_s(X) = f_s(X) - \hat{Y}(X, D_s)$ and the second inequality follows from the Proposition \ref{dcov} by noting that $\left(R(X_t) + \delta_s(X_t), X_t^{(j)}\right)$ is independent of $\epsilon_t$.
    From (\cite{szekely2007measuring} , Remark 3), we have
\begin{align*}
    \mathcal{V}^2\left(R(X_t)  + \delta_s(X_t), X_t^{(j)}\right)= P_1 + P_2 - 2P_3
\end{align*}
where 
\begin{align*}
&P_1 = \mathbb{E}\Bigl[
   \bigl|X_t^{(j)} - \bigl(X_t^{(j)}\bigr)'\bigr|
   \,\bigl|\,
   R(X_t) - R(X_t)' + \delta_s(X_t) - \delta_s(X_t)'
   \bigr|
\Bigr]
\\
&P_2 = \mathbb{E}\bigl|
   X_t^{(j)} - \bigl(X_t^{(j)}\bigr)'
\bigr|
\mathbb{E}\bigl|\,
  R(X_t) - R(X_t)' + \delta_s(X_t) - \delta_s(X_t)'
\bigr|
\\
&P_3 = \mathbb{E}\Bigl[
   \bigl|X_t^{(j)} - \bigl(X_t^{(j)}\bigr)'\bigr|
   \,\bigl|\,
   R(X_t) - R(X_t)'' + \delta_s(X_t) - \delta_s(X_t)''
   \bigr|
\Bigr].
\end{align*}
where
$\left((X_t^{(j)})', R(X_t)', \delta_s(X_t)'\right)$ and $\left((X_t^{(j)})'',R(X_t)'',\delta_s(X_t)'' \right)$ are independent copies of $\left(X_t^{(j)}, R(X_t), \delta_s(X_t)\right)$.

For $P_1$ and $P_2$, we have
\begin{align*}
    &P_1 = \mathbb{E}\Bigl[
   \bigl|X_t^{(j)} - \bigl(X_t^{(j)}\bigr)'\bigr|
   \,\bigl|\,
   R(X_t) - R(X_t)' + \delta_s(X_t) - \delta_s(X_t)'
   \bigr|
\Bigr]\\
&\quad \leq \mathbb{E}\Bigl[
   \bigl|X_t^{(j)} - \bigl(X_t^{(j)}\bigr)'\bigr|
   \,\bigl|\,
  R(X_t) - R(X_t)'
   \bigr|
\Bigr] + \mathbb{E}\Bigl[ \bigl|X_t^{(j)} - \bigl(X_t^{(j)}\bigr)'\bigr|\bigl| \delta_s(X_t) - \delta_s(X_t)'\bigr|\Bigr]
\end{align*}

\begin{align*}
    &P_2 =\mathbb{E}\bigl|
   X_t^{(j)} - \bigl(X_t^{(j)}\bigr)'
\bigr|
\mathbb{E}\bigl|\,
   \bigl(R(X_t) - R(X_t)' + \delta_s(X_t) - \delta_s(X_t)'\bigr)
\bigr|\\
&\quad \leq \mathbb{E}
   \bigl|X_t^{(j)} - \bigl(X_t^{(j)}\bigr)'\bigr|
   \,\mathbb{E}\bigl|\,
  R(X_t) - R(X_t)'
   \bigr|
 + \mathbb{E}\bigl|X_t^{(j)} - \bigl(X_t^{(j)}\bigr)'\bigr|\mathbb{E}\bigl| \delta_s(X_t) - \delta_s(X_t)'\bigr|
\end{align*}
For $P_3$, we have
\begin{align*}
    &P_3 = \mathbb{E}\Bigl[
   \bigl|X_t^{(j)} - \bigl(X_t^{(j)}\bigr)'\bigr|
   \,\bigl|\,
   R(X_t) - R(X_t)'' + \delta_s(X_t) - \delta_s(X_t)''
   \bigr|
\Bigr]\\
& \geq \mathbb{E}\Bigl[
   \bigl|X_t^{(j)} - \bigl(X_t^{(j)}\bigr)'\bigr|
   \,\bigl|\,
  R(X_t) - R(X_t)''
   \bigr|
\Bigr] - \mathbb{E}\Bigl[ \bigl|X_t^{(j)} - \bigl(X_t^{(j)}\bigr)'\bigr|\bigl| \delta_s(X_t) - \delta_s(X_t)''\bigr|\Bigr]
\end{align*}
Conclusively, we have
\begin{align*}
    &\mathcal{V}^2\left(R(X_t)  + \delta_s(X_t), X_t^{(j)}\right)\\
    &\qquad\leq  
    \mathbb{E}\Bigl[ \bigl|X_t^{(j)} - \bigl(X_t^{(j)}\bigr)'\bigr|\bigl| \delta_s(X_t) - \delta_s(X_t)'\bigr|\Bigr]+\mathbb{E}\bigl|X_t^{(j)} - \bigl(X_t^{(j)}\bigr)'\bigr|\mathbb{E}\bigl| \delta_s(X_t) - \delta_s(X_t)'\bigr|\\
    &\qquad\quad + 2\mathbb{E}\Bigl[ \bigl|X_t^{(j)} - \bigl(X_t^{(j)}\bigr)'\bigr|\bigl| \delta_s(X_t) - \delta_s(X_t)''\bigr|\Bigr]\\
    &\qquad \leq 4\mathbb{E}\bigl| \delta_s(X_t) - \delta_s(X_t)'\bigr|\\
    &\qquad \leq 4\sqrt{2}\sqrt{\mathbb{E}\bigl| \delta_s(X_t)\bigr|^2}.
\end{align*}
The penultimate inequality follows since $\bigl|X_t^{(j)} - \bigl(X_t^{(j)}\bigr)'\bigr|$ is bounded by 1 and therefore 
$$
 \mathbb{E}\Bigl[ \bigl|X_t^{(j)} - \bigl(X_t^{(j)}\bigr)'\bigr|\bigl| \delta_s(X_t) - \delta_s(X_t)'\bigr|\Bigr]
 \leq \mathbb{E}\Bigl[ \bigl| \delta_s(X_t) - \delta_s(X_t)'\bigr|\Bigr]$$

From proposition 3.1 (in main text), we see that $\mathbb{E}[\delta_s(X_t)^2]$ is bounded by 
\[
\mathbb{E}[\delta_s(X_t)^2] \leq C_s\left(n_s(\log_2^{d-1}n_s)^{\frac{1}{2}}\right)^{-z_s},
\]
with appropriate choices of $p^s$ and $k_{n_s} = C_s\left(n_s (\log_2^{d-1} n_s)^{1/2}\right)^{1-z_s}$ as outlined in the proposition.
Therefore, the result in the lemma follows
\begin{align*}    \mathcal{V}^2\left(\tilde{Y}(X_t, D_s), X_t^{(j)}\right)\leq \tilde{C}_s\left(n_s(\log_2^{d-1}n_s)^{\frac{1}{2}}\right)^{-z_s/2},
\end{align*}
for some constant $C_s > 0$ independent of $n_s$.
\end{proof}

\subsection{Proof of Theorem 3.6}

\begin{proof}
 We will use the result in Proposition \ref{distance}. We note that both $\Omega$ and $\hat{\Omega}$ are strictly positive numbers. First, for notational convenience, let $\epsilon_n = cn_t^{-\alpha}$. Then we note that,
    \begin{align*}
\mathbb{P}\left(\max_{j\in \mathcal{I}_R^c}\left| \frac{\hat{\omega}_j}{\hat{\Omega}} - \frac{\omega_j}{\Omega} \right| \geq \epsilon_n\right)
&\leq \mathbb{P}\left( \max_{j\in \mathcal{I}_R^c}\left| \frac{\hat{\omega}_j}{\hat{\Omega}} - \frac{\omega_j}{\hat{\Omega}} \right| + \max_{j\in \mathcal{I}_R^c}\left| \frac{\omega_j}{\hat{\Omega}} - \frac{\omega_j}{\Omega} \right| \geq \epsilon_n \right)\\
&\leq \mathbb{P}\left( \max_{j\in \mathcal{I}_R^c}\left| \frac{\hat{\omega}_j}{\hat{\Omega}} - \frac{\omega_j}{\hat{\Omega}} \right| \geq \frac{\epsilon_n}{2} \right)
+ \mathbb{P}\left( \max_{j\in \mathcal{I}_R^c}\left| \frac{\omega_j}{\hat{\Omega}} - \frac{\omega_j}{\Omega} \right| \geq \frac{\epsilon_n}{2} \right)\\
& \leq \mathbb{P}\left( \max_{j\in \mathcal{I}_R^c}\left| \hat{\omega}_j - \omega_j \right| \geq \frac{\epsilon_n}{2} \hat{\Omega} \right)
+ \mathbb{P}\left( \omega_{max}\cdot \left| \frac{1}{\hat{\Omega}} - \frac{1}{\Omega} \right| \geq \frac{\epsilon_n}{2} \right)
\end{align*}

The first term on the right hand side can be bounded as follows. For any $0<\epsilon_1 < \Omega$,
we have,
\begin{align*}
\mathbb{P}\left( \max_{j\in \mathcal{I}_R^c}\left| \hat{\omega}_j - \omega_j \right| \geq \frac{\epsilon_n}{2} \hat{\Omega} \right)
&= \mathbb{P} \left( \max_{j\in \mathcal{I}_R^c}|\hat{\omega}_j - \omega_j| \geq \frac{\epsilon_n}{2} \hat{\Omega},\, |\hat{\Omega} - \Omega| < \epsilon_1 \right) \\
&\quad + \ \mathbb{P} \left( \max_{j\in \mathcal{I}_R^c}|\hat{\omega}_j - \omega_j| \geq \frac{\epsilon_n}{2} \hat{\Omega},\, |\hat{\Omega} - \Omega| \geq \epsilon_1 \right) \\
&\leq \mathbb{P}\left( \max_{j\in \mathcal{I}_R^c}|\hat{\omega}_j - \omega_j| \geq \frac{\epsilon_n}{2} (\Omega - \epsilon_1),\, |\hat{\Omega} - \Omega| < \epsilon_1 \right) \\
&\quad + \ \mathbb{P}\left(\max_{j\in \mathcal{I}_R^c}|\hat{\omega}_j - \omega_j| \geq \frac{\epsilon_n}{2} \hat{\Omega},\, |\hat{\Omega} - \Omega| \geq \epsilon_1 \right) \\
&\leq \mathbb{P}\left( \max_{j\in \mathcal{I}_R^c}|\hat{\omega}_j - \omega_j| \geq \frac{\epsilon_n}{2} (\Omega - \epsilon_1) \right)
+ \mathbb{P}(|\hat{\Omega} - \Omega| \geq \epsilon_1) \\
\end{align*}

Now, we analyze the second term. Since $\Omega>0$, define the event $E=\{\hat\Omega \ge \Omega/2\}$. On $E$ we have
$\Omega\hat\Omega \ge \Omega^2/2$, hence
\[
\left|\frac{1}{\hat\Omega}-\frac{1}{\Omega}\right|
= \frac{|\hat\Omega-\Omega|}{\Omega\hat\Omega}
\le \frac{2}{\Omega^2}|\hat\Omega-\Omega|.
\]
Therefore,
\[
\left\{\omega_{\max}\left|\frac{1}{\hat\Omega}-\frac{1}{\Omega}\right|\ge \frac{\epsilon_n}{2}\right\}\cap E
\subseteq
\left\{|\hat\Omega-\Omega|\ge \frac{\epsilon_n\Omega^2}{4\omega_{\max}}\right\}.
\]
It follows that
\begin{align*}
\mathbb P\left(\omega_{\max}\left|\frac{1}{\hat\Omega}-\frac{1}{\Omega}\right|\ge \frac{\epsilon_n}{2}\right)
&\le 
\mathbb P\left(|\hat\Omega-\Omega|\ge \frac{\epsilon_n\Omega^2}{4\omega_{\max}}\right)
+\mathbb P(E^c)\\
&\le 
\mathbb P\left(|\hat\Omega-\Omega|\ge \frac{\epsilon_n\Omega^2}{4\omega_{\max}}\right)
+\mathbb P\left(|\hat\Omega-\Omega|\ge \Omega/2\right).
\end{align*}

Now consider for a given $\alpha^\prime$ and $c^\prime$, set $\epsilon_n^\prime = c^\prime n_t^{-\alpha^\prime}$, with $0\leq \alpha'<1/2.$
\begin{align*}
\mathbb{P}(|\hat{\Omega} - \Omega| \geq \epsilon_n^\prime) 
&\leq \mathbb{P}(|\hat{\Omega}_{\mathcal{I}_R^c} - \Omega_{\mathcal{I}_R^c}| + |\hat{\Omega}_{\mathcal{I}_R} - \Omega_{\mathcal{I}_R}| \geq \epsilon_n^\prime) \\
&\leq \mathbb{P}(|\hat{\Omega}_{\mathcal{I}_R^c} - \Omega_{\mathcal{I}_R^c}| \geq \tfrac{\epsilon_n^\prime}{2}) + \mathbb{P}(|\hat{\Omega}_{\mathcal{I}_R} - \Omega_{\mathcal{I}_R}| \geq \tfrac{\epsilon_n^\prime}{2}) 
\end{align*}

From the Proposition \ref{distance}, we have for any $0<\eta^\prime<1/2 - \alpha^\prime$, there exists positive constants $c_1$, $c_1^\prime$, $c_2$, $c_2^\prime$ such that
\begin{align*}
    \mathbb{P}(|\hat{\Omega}_{\mathcal{I}_R^c} - \Omega_{\mathcal{I}_R^c}| \geq \tfrac{\epsilon_n^\prime}{2})&\leq \sum_{j\in \mathcal{I}_R^c}\mathbb{P}(\left|\hat{\omega}_j-\omega_j\right|\geq \tfrac{\epsilon_n^\prime}{2|\mathcal{I}_R^c|})\\
    &\leq O\left(|\mathcal{I}_R^c|\left[\exp(-c_1\left(\tfrac{\epsilon_n^\prime}{|\mathcal{I}_R^c|}\right)^2n_t^{1-2\eta^\prime})+n_t \exp(-c_2n_t^{\eta^\prime})\right]\right)\\
    \mathbb{P}(|\hat{\Omega}_{\mathcal{I}_R} - \Omega_{\mathcal{I}_R}| \geq \tfrac{\epsilon_n^\prime}{2})&\leq \sum_{j\in \mathcal{I}_R}\mathbb{P}(\left|\hat{\omega}_j-\omega_j\right|\geq \tfrac{\epsilon_n^\prime}{2|\mathcal{I}_R|})\\
    &\leq O\left(|\mathcal{I}_R|\left[\exp(-c_1^\prime\left(\tfrac{\epsilon_n^\prime}{|\mathcal{I}_R|}\right)^2n_t^{1-2\eta^\prime})+n_t \exp(-c_2^\prime n_t^{\eta^\prime})\right]\right).
\end{align*}
Then, there exists $\tilde{c}_1$, $\tilde{c}_2$ depending on $c_1$, $c_1^\prime$, $c_2$, $c_2^\prime$, such that
\begin{align*}
    \mathbb{P}(|\hat{\Omega} - \Omega| \geq \epsilon_n^\prime)  \leq O\left(|\mathcal{I}_R|\left[\exp(-\tilde{c}_1\left(\tfrac{\epsilon_n^\prime}{|\mathcal{I}_R|}\right)^2n_t^{1-2\eta^\prime })+n_t \exp(-\tilde{c}_2 n_t^{\eta^\prime})\right]\right)
\end{align*}  

The last inequality holds because $\left|\mathcal{I}_R\right|>\left|\mathcal{I}_R^c\right|$.

In summary, 
\begin{align*}
    \mathbb{P}\left(\max_{j\in \mathcal{I}_R^c}\left| \frac{\hat{\omega}_j}{\hat{\Omega}} - \frac{\omega_j}{\Omega} \right|  \geq \epsilon_n\right)&\leq \mathbb{P}\left(\max_{j\in \mathcal{I}_R^c} \left| \hat{\omega}_j - \omega_j \right| \geq \frac{\epsilon_n}{2} |\hat{\Omega}| \right)
+ \mathbb{P}\left( \omega_{max} \cdot \left| \frac{1}{\hat{\Omega}} - \frac{1}{\Omega} \right| \geq \frac{\epsilon_n}{2} \right)\\
&\leq \mathbb{P}\left( \max_{j\in \mathcal{I}_R^c}|\hat{\omega}_j - \omega_j| \geq \frac{\epsilon_n}{2} (\Omega - \epsilon_1) \right)
+ \mathbb{P}(|\hat{\Omega} - \Omega| \geq \epsilon_1) \\
&\quad + \mathbb P\left(|\hat\Omega-\Omega|\ge \frac{\epsilon_n\Omega^2}{4\omega_{\max}}\right)
+\mathbb P\left(|\hat\Omega-\Omega|\ge \Omega/2\right)
\end{align*}

For a given $\epsilon_n = cn_t^{-\alpha}$, let $\epsilon_1 = \Omega/2$, for any $0 < \eta < 1/2 - \alpha$, there exists $c_1, c_2, c_3, c_4$, s.t.

\[
 \mathbb{P}\left( \max_{j\in \mathcal{I}_R^c}|\hat{\omega}_j - \omega_j| \geq \frac{\epsilon_n\Omega}{4} \right)\leq O\left(|\mathcal{I}^c_R|\left[\exp(-c_1\left(\epsilon_n\Omega\right)^2n_t^{1-2\eta})+n_t \exp(-c_2 n_t^{\eta})\right]\right)
\]
\[
\mathbb P\left(|\hat\Omega-\Omega|\ge \frac{\epsilon_n\Omega^2}{4\omega_{\max}}\right)\leq O\left(|\mathcal{I}_R|\left[\exp(-c_3\left(\tfrac{\epsilon_n\Omega^2}{|\mathcal{I}_R|\omega_{\max}}\right)^2n_t^{1-2\eta})+n_t \exp(-c_4 n_t^{\eta})\right]\right)
\]
and for any $0 < \eta_1 < 1/2$, there exists $c_5, c_6$, s.t.
\[
\mathbb P\left(|\hat\Omega-\Omega|\ge\Omega/2\right)\leq O\left(|\mathcal{I}_R|\left[\exp(-c_5\left(\tfrac{\Omega}{|\mathcal{I}_R|}\right)^2n_t^{1-2\eta_1})+n_t \exp(-c_6 n_t^{\eta_1})\right]\right)
\]

Consequently, for any $0 < \eta < 1/2 - \alpha$, there exists $c_7, c_8$, s.t.

\begin{align*}
    \mathbb{P}\left(\max_{j\in \mathcal{I}_R^c}\left| \frac{\hat{\omega}_j}{\hat{\Omega}} - \frac{\omega_j}{\Omega} \right|  \geq cn_t^{-\alpha}\right)&\leq \mathbb{P}\left( \max_{j\in \mathcal{I}_R^c}|\hat{\omega}_j - \omega_j| \geq \frac{\epsilon_n\Omega}{4} \right)
+ \mathbb{P}\left(|\hat{\Omega} - \Omega| \geq \frac{\epsilon_n\Omega^2}{4\omega_{\max}}\right)\\
&\quad + 2\mathbb P\left(|\hat\Omega-\Omega|\ge \Omega/2\right)\\
&\leq O\left(|\mathcal{I}_R|\left[\exp(-c_7\left(\tfrac{\Omega}{|\mathcal{I}_R|}\right)^2n_t^{1-2(\eta + \alpha)}+n_t \exp(-c_8 n_t^{\eta})\right]\right)
\end{align*}
\end{proof}

\subsection{Proof of Theorem 3.7}

    \begin{proof}
Following the similar procedure in Theorem 3.6, for any given $ \epsilon_n = cn_t^{-\alpha}$,
\begin{align*}
    \mathbb{P}\left\{\max_{j\in \mathcal{I}_R}\left|\frac{\hat{\omega}_k}{\hat{\Omega}}-\frac{\omega_j}{\Omega} \right|\geq \epsilon_n\right\}  &\leq \mathbb{P}\left( \max_{j\in \mathcal{I}_R}\left| \frac{\hat{\omega}_j}{\hat{\Omega}} - \frac{\omega_j}{\hat{\Omega}} \right| + \max_{j\in \mathcal{I}_R}\left| \frac{\omega_j}{\hat{\Omega}} - \frac{\omega_j}{\Omega} \right| \geq \epsilon_n \right)\\
&\leq \mathbb{P}\left( \max_{j\in \mathcal{I}_R}\left| \frac{\hat{\omega}_j}{\hat{\Omega}} - \frac{\omega_j}{\hat{\Omega}} \right| \geq \frac{\epsilon_n}{2} \right)
+ \mathbb{P}\left( \max_{j\in \mathcal{I}_R}\left| \frac{\omega_j}{\hat{\Omega}} - \frac{\omega_j}{\Omega} \right| \geq \frac{\epsilon_n}{2} \right)\\
& \leq \mathbb{P}\left( \max_{j\in \mathcal{I}_R}\left| \hat{\omega}_j - \omega_j \right| \geq \frac{\epsilon_n}{2} \hat{\Omega} \right)
+ \mathbb{P}\left( \omega_{max}^R\cdot \left| \frac{1}{\hat{\Omega}} - \frac{1}{\Omega} \right| \geq \frac{\epsilon_n}{2} \right)\\
&\leq \mathbb{P}\left( \max_{j\in \mathcal{I}_R}|\hat{\omega}_j - \omega_j| \geq \frac{\epsilon_n\Omega}{4} \right)
+ \mathbb{P}\left(|\hat{\Omega} - \Omega| \geq \frac{\epsilon_n\Omega^2}{4\omega_{\max}^R}\right)\\
&\quad + 2\mathbb P\left(|\hat\Omega-\Omega|\ge \Omega/2\right)
\end{align*}
From Lemma 3.4, we know 
$$
\omega_{max}^R\leq \tilde{C}_s\left(n_s(\log_2^{d-1}n_s)^{\frac{1}{2}}\right)^{-z_s/2}
$$
Set $\delta_{n_s} = \tilde{C}_s\left(n_s(\log_2^{d-1}n_s)^{\frac{1}{2}}\right)^{-z_s/2}$, then 
\[
\mathbb{P}\left(|\hat{\Omega} - \Omega| \geq \frac{\epsilon_n\Omega^2}{4\omega_{\max}^R}\right)\leq \mathbb{P}\left(|\hat{\Omega} - \Omega| \geq \frac{\epsilon_n\Omega^2}{4\delta_{n_s}}\right)
\]

For a given $\epsilon_n = cn_t^{-\alpha}$, for any $0 < \eta < 1/2 - \alpha$, there exists $c_1, c_2, c_3, c_4$, s.t.

\[
 \mathbb{P}\left( \max_{j\in \mathcal{I}_R}|\hat{\omega}_j - \omega_j| \geq \frac{\epsilon_n\Omega}{4} \right)\leq O\left(|\mathcal{I}_R|\left[\exp(-c_1\left(\epsilon_n\Omega\right)^2n_t^{1-2\eta})+n_t \exp(-c_2 n_t^{\eta})\right]\right)
\]
\[
\mathbb P\left(|\hat\Omega-\Omega|\ge \frac{\epsilon_n\Omega^2}{4\delta_{n_s}}\right)\leq O\left(|\mathcal{I}_R|\left[\exp(-c_3\left(\tfrac{\epsilon_n\Omega^2}{4|\mathcal{I}_R|\delta_{n_s}}\right)^2n_t^{1-2\eta})+n_t \exp(-c_4 n_t^{\eta})\right]\right)
\]
and for any $0 < \eta_1 < 1/2$, there exists $c_5, c_6$, s.t.
\[
\mathbb P\left(|\hat\Omega-\Omega|\ge\Omega/2\right)\leq O\left(|\mathcal{I}_R|\left[\exp(-c_5\left(\tfrac{\Omega}{|\mathcal{I}_R|}\right)^2n_t^{1-2\eta_1})+n_t \exp(-c_6 n_t^{\eta_1})\right]\right)
\]

Since $\tfrac{\epsilon_n\Omega^2}{4|\mathcal{I}_R|\delta_{n_s}}$ is much larger than $\epsilon_n\Omega$, we have for any $0 < \eta < 1/2 - \alpha$, there exists $c_7, c_8$, s.t.
\begin{align*}
    \mathbb{P}\left\{\max_{j\in \mathcal{I}_R}\left|\frac{\hat{\omega}_k}{\hat{\Omega}}-\frac{\omega_j}{\Omega} \right|\geq \epsilon_n\right\} 
&\leq \mathbb{P}\left( \max_{j\in \mathcal{I}_R}|\hat{\omega}_j - \omega_j| \geq \frac{\epsilon_n\Omega}{4} \right)
+ \mathbb{P}\left(|\hat{\Omega} - \Omega| \geq \frac{\epsilon_n\Omega^2}{4\delta_{n_s}}\right)\\
&\quad + 2\mathbb P\left(|\hat\Omega-\Omega|\ge \Omega/2\right)\\
&\leq O\left(|\mathcal{I}_R|\left[\exp(-c_7\left(\tfrac{\Omega}{|\mathcal{I}_R|}\right)^2n_t^{1-2(\eta + \alpha)}+n_t \exp(-c_8 n_t^{\eta})\right]\right)
\end{align*}

$$
\mathbb{P}\left(\max_{j\in \mathcal{I}_R}\left|\frac{\hat{\omega}_j}{\hat{\Omega}}-\frac{\omega_j}{\Omega} \right|\geq \epsilon_n\right)\geq\mathbb{P}\left\{\max_{j\in \mathcal{I}_R}\left|\frac{\hat{\omega}_j}{\hat{\Omega}} \right|\geq \epsilon_n + \frac{\omega_{max}^R}{\Omega}\right\} \geq \mathbb{P}\left\{\max_{j\in \mathcal{I}_R}\left|\frac{\hat{\omega}_k}{\hat{\Omega}} \right|\geq c n_t^{-\alpha} + \delta_{n_s}/\Omega\right\}
$$

Find $\tilde{c}$ and $\tilde{\alpha}$ such that $\tilde{c}n_t^{-\tilde{\alpha}} = c n_t^{-\alpha} + \delta_{n_s}/\Omega$, then we have
\begin{align*}
      \mathbb{P}\left(\max_{j\in \mathcal{I}_R}\left|\frac{\hat{\omega}_k}{\hat{\Omega}} \right|\geq \tilde{c} n_t^{-\tilde{\alpha}}\right) &\leq \mathbb{P}\left(\max_{j\in \mathcal{I}_R}\left|\frac{\hat{\omega}_k}{\hat{\Omega}}-\frac{\omega_k}{\Omega} \right|\geq \epsilon_n\right)\\
      &\leq  O\left(|\mathcal{I}_R|\left[\exp(-c_7\left(\tfrac{\Omega}{|\mathcal{I}_R|}\right)^2n_t^{1-2(\eta + \alpha)}+n_t \exp(-c_8 n_t^{\eta})\right]\right)
\end{align*}

Consequently, 
\begin{align*}
    \mathbb{P}\left\{\max_{j\in \mathcal{I}_R}\left|\frac{\hat{\omega}_j}{\hat{\Omega}} \right|\geq \tilde{c}n_t^{-\tilde{\alpha}}\right\}
    & = 1- \mathbb{P}\left\{\max_{j\in \mathcal{I}_R}\left|\frac{\hat{\omega}_j}{\hat{\Omega}} \right|\geq \tilde{c}n_t^{-\tilde{\alpha}}\right\}\\
    &\geq 1-O\left(|\mathcal{I}_R| \left[\exp\left(-c_7 \left(\tfrac{\Omega}{|\mathcal{I}_R|}\right)^2 n_t^{1-2(\alpha+\eta)}\right) + n_t \exp(-c_8 n_t^{\eta})\right]\right)
\end{align*}

    \end{proof}

\subsection{Proof of Theorem 3.8}

\begin{proof}

From step 1, we got the expectation of predictions from the random forest trained on the source data
$$\hat{Y}(X, D_s) = \mathbb{E}_{\Theta_s}\left[\hat{Y}(X;\Theta_s,D_s)\right].$$ The residuals are calculated through target data in step 2 yielding $\tilde{Y}_i(X_{it}, D_s):=Y_{it} - \hat{Y}_i(X_{it}, D_s), i = 1, \cdots, n_t$. Then step 3 trained the random forest using data set $\tilde{D}_t = (\tilde{Y}(X_t, D_s),X_t)$ yielding the expectation of predictions $$\hat{Y}(X, \tilde{D}_t) = \mathbb{E}_{\Theta_t}\left[\hat{Y}(X;\Theta_t,\tilde{D}_t)\right].$$
Our predictions of new data $X$ would be $\hat{Y}(X,(D_s,\tilde{D}_t)) = \hat{Y}(X,D_s) + \hat{Y}(X,\tilde{D}_t)$. The goal is to upper bound 
the MSE error $\mathbb{E}\left[\left(\hat{Y}(X,(D_s,\tilde{D}_t)) - f_t(X)\right)^2\right].$
We note that 
\begin{align*}
&\mathbb{E}\left[\left(\hat{Y}(X,(D_s,\tilde{D}_t)) - f_t(X)\right)^2\right] \notag \\
&= \mathbb{E}\left[\left(\left(\hat{Y}(X,D_s) - f_s(X)\right) + \left(\hat{Y}(X,\tilde{D}_t) - \left(f_t(X) - f_s(X)\right)\right)\right)^2\right] \notag \\
&\leq 2\mathbb{E}\left[\left(\hat{Y}(X,D_s) - f_s(X)\right)^2\right] + 2\mathbb{E}\left[\left(\hat{Y}(X,\tilde{D}_t) - \left(f_t(X) - f_s(X)\right)\right)^2\right]\\
&\leq 2\mathbb{E}\left[
            \delta_s^2(X)
 \right] + 2\mathbb{E}\left[\left(\hat{Y}(X,\tilde{D}_t) - \left(f_t(X) - f_s(X)\right)\right)^2\right],
\end{align*}
where $\delta_s(X) = \hat{Y}(X, D_s) - f_s(X)$.

Now we have the following decomposition
\begin{align*}
\tilde{Y}(X_t, D_s) &= Y_t - \hat{Y}(X_t, D_s)\\
& = f_t(X_t) - f_s(X_t) +\epsilon_t - \left(\hat{Y}(X_t, D_s) - f_s(X_t)\right)
\end{align*}
Recalling the definition that $R(X)=f_t(X) - f_s(X)$, we have
$\tilde{Y}(X_t, D_s)-\delta_s(X_t) = R(X_t) + \epsilon_t,$ for any arbitrary data point in the target domain $Y_t, X_t$. Recall $\hat{Y}(X,\tilde{D}_t)$ is the prediction of the test data $X$ under the model trained by $\tilde{D}_t = \left(\tilde{Y}(X_t, D_s),X_t\right)$. Since
\begin{align*}
   \hat{Y}(X,\tilde{D}_t) &= \mathbb{E}_{\tilde{\Theta}}\left[\hat{Y}(X,\tilde{\Theta},\tilde{D}_t)\right]\\
    &= \sum_{i=1}^{n_t}\mathbb{E}_{\tilde{\Theta}}\left[\tilde{W}_i\right]\tilde{Y}(X_i, D_s)\\
    &=\sum_{i=1}^{n_t}\mathbb{E}_{\tilde{\Theta}}\left[\tilde{W}_i\right]\left(\delta_s(X_i)+R(X_i) + \epsilon_i\right)\\
\end{align*}
where $\tilde{W}_i = \frac{ \mathbf{1}(X \in \tilde{t})}{\sum_{i=1}^{n_t} \mathbf{1} (X \in \tilde{t})} \mathbf{1} (\sum_{i=1}^{n_t} \mathbf{1} (X \in \tilde{t}) \neq 0)$, and $\tilde{t}:=t(X,\tilde{\Theta},\tilde{D}_t)$ is the leaf node of a tree in the residual random forest that contains $X$. We note that $\sum_{i=1}^{n_t}\tilde{W}_i = 1$ a.s.

Then, 
\begin{align}
 &\mathbb{E}\left[\left(\hat{Y}(X,\tilde{D}_t)  - R(X)\right)^2\right] \notag\\
    &\qquad = \mathbb{E}\left[ \sum_{i=1}^{n_t}\mathbb{E}_{\tilde{\Theta}}\left[\tilde{W}_i\right]\left(\delta_s(X_i)+R(X_i) + \epsilon_i \right) - R(X) \right]^2 \notag\\
    &\qquad =\mathbb{E}\left[{\rm E}_{\tilde{\Theta}} \left[\sum_{i=1}^{n_t}\tilde{W}_i\left(\delta_s(X_i)+R(X_i) + \epsilon_{it}\right)   - R(X)\right]\right]^2 \notag\\
    &\qquad \leq \mathbb{E}\left[\sum_{i=1}^{n_t}\tilde{W}_i\left(R(X_i) + \epsilon_{it} \right) - R(X)+\sum_{i=1}^{n_t}\tilde{W}_i\delta_s(X_i)\right]^2 \notag\\
    &\qquad\leq 2\mathbb{E}\left[\sum_{i=1}^{n_t}\tilde{W}_i\left(R(X_i) + \epsilon_{it} \right)- R(X) \right]^2 + 2\mathbb{E}\left[\sum_{i=1}^{n_t}\tilde{W}_i\delta_s(X_i) \right]^2 \label{s11}
\end{align}

In the first term in Equation \ref{s11}, we can think of it as the MSE from the same centered random forest but with different response variables. That is, we build the centered random forest with $p_{nj}^t \;=\;\frac{\hat{\omega}_j^t}{\hat{\Omega}_t},\ j=1,\cdots, d$ and the depth $\lceil \log_2k_{n_t}\rceil$. Then we use the model to predict $R(X)$ using the data set $\left\{X_i, Y_R(X_i)\right\}$. Here we only change the response variables as $Y_R(X_i) = R(X_i) + \epsilon_{it}$. This new response variable is unobserved. The reason that we can do this is that $\tilde{W}_i$ is only determined by the $X_i$, $p_{nj}^t \;=\;\frac{\hat{\omega}_j^t}{\hat{\Omega}_t},\ j=1,\cdots, d$ and the depth $\lceil \log_2k_{n_t}\rceil$, and does not depend on the response.

Since the function $R(X)$ is ``low-dimensional'', i.e., only depends on the set $\mathcal{I}_R^C$, we wish to apply the result of Theorem 3.3 to the first term in Equation \ref{s11}. We verify the conditions of the theorem. In the notation of the thereom our set $S=\mathcal{I}_R$ and $S^C =\mathcal{I}_R^C$. We choose $S_{\alpha}$ to be equal to $S$.

Denote the event $E_1 = \left\{ p_{nj}^t\leq \tilde{c}n_t^{-\tilde{\alpha}}, \, \forall j\in \mathcal{I}_R \right\}$, and $E_2 = \{\frac{\hat{\omega}_j}{\hat{\Omega}_j} > \frac{\omega_j}{\Omega_j} -  cn_t^{-\alpha}, \, \forall j\in \mathcal{I}_R^C \}$

From Theorem 3.7, we know,
\begin{align*}
\mathbb{P}\left(E_1\right)\geq 1-O\left(|\mathcal{I}_R| \left[\exp\left(-c_1 \left(\tfrac{\Omega}{|\mathcal{I}_R|}\right)^2 n_t^{1-2(\alpha+\eta)}\right) + n_t \exp(-c_2 n_t^{\eta})\right]\right),
\end{align*}
and from Theorem 3.6 we have,
\[
\mathbb{P}\left(E_2\right)\geq 1-O\left(|\mathcal{I}_R|\left[\exp(-c_1\left(\tfrac{\Omega}{|\mathcal{I}_R|}\right)^2n_t^{1-2(\eta + \alpha)}+n_t \exp(-c_2 n_t^{\eta})\right]\right).
\]
Note that since by assumption $\min_{j \in \mathcal{I}_R^C} \omega_j \geq cn_t^{-\alpha}$, under the event $E_2$, $p_{nj}^t >0$ for all $j \in \mathcal{I}_R^C$. Then consider the event $E= E_1 \cap E_2$. Therefore, under the event $E$, the conditions of Theorem 3.3 are satisfied with $S_{\alpha}=S=\mathcal{I}_R$, and we get the following upper bound of the first term in Equation \ref{s11}. 
\begin{align*}   &2\mathbb{E}\left[\sum_{i=1}^{n_t}\tilde{W}_i\left(R(X_i) + \epsilon_{it} \right)- R(X) \right]^2\\
    &\qquad\leq  C_R\Bigg(|\mathcal{I}_R^c| \sum_{j\in\mathcal{I}_R^c} \left\| \partial_j R \right\|_{\infty}^2 k_{n_t}^{2 \log_2 (1 - p_{nj}^t / 2)} + M_t^2 e^{-n/(2 k_{n_t})}\\
&\qquad \quad + \frac{12 \sigma_t^2 k_{n_t}}{n_t} \frac{8^{d}}{
    \sqrt{
        \prod_{j \in \mathcal{I}_R^c} p_{nj}^t \times \log_2^{|\mathcal{I}_R^c| - 1}(k_{n_t})
    }
}\Bigg).
\end{align*}

What is left is the second term in Equation \ref{s11}. We note,

\begin{align*}
    2\mathbb{E}\left[\sum_{i=1}^{n_t}\tilde{W}_i\delta_s(X_i) \right]^2&\leq  2\,\mathbb{E}\left[
    \textstyle\sum_{i=1}^{n_t}\sqrt{\tilde{W}_i } \sqrt{\tilde{W}_i}
            \delta_s(X_i)
 \right]^2\\
& \leq 2\,\mathbb{E}\left[\Bigg(
        \textstyle\sum_{i=1}^{n_t} \tilde{W}_i \Bigg)\Bigg(\textstyle\sum_{i=1}^{n_t}  \tilde{W}_i 
            \delta_s^2(X_i)\Bigg)
 \right]\\
 &\leq 2\,\mathbb{E}\left[\textstyle\sum_{i=1}^{n_t}  \tilde{W}_i 
            \delta_s^2(X_i)
 \right]\\
 &\leq 2\,\textstyle\sum_{i=1}^{n_t} \mathbb{E}\left[ \tilde{W}_i 
            \delta_s^2(X_i)
 \right]\\
 &\leq 2\,n_t \mathbb{E}\left[ \tilde{W}_1 
            \delta_s^2(X_1)
 \right]
\end{align*}
From \cite{biau2012analysis}, it can be shown that
\begin{align*}
    &\mathbb{E}\left[ \tilde{W}_1 
            \delta_s^2(X_1)
 \right]\\
 &\qquad = \mathbb{E} \left[  \delta_s^2(X_1)\, \mathbf{1}_{\{\mathbf{X}_1 \in A_{n_t}(\mathbf{X}, \tilde{\Theta})\}} 
  \, \mathbb{E} \left[ \left. \frac{1}{1 + \sum_{i=2}^{n_t} \mathbf{1}_{\{\mathbf{X}_i \in A_{n_t}(\mathbf{X}, \tilde{\Theta})\}}} \right|\, \mathbf{X}, \tilde{\Theta} \right] \right]
\end{align*}
 and
 \begin{align*}
\mathbb{E} \left[
  \left. \frac{1}{1 + \sum_{i=2}^{n_t} \mathbf{1}_{\{\mathbf{X}_i \in A_{n_t}(\mathbf{X}, \tilde{\Theta})\}}}
  \,\right|\, \mathbf{X}, \tilde{\Theta}
\right]
\leq \frac{2^{\lceil \log_2 k_{n_t} \rceil}}{n}
\leq \frac{2^{2^{\log_2 k_{n_t} + 1}}}{n_t} \leq \frac{2k_{n_t}}{{n_t}}.
 \end{align*}
Then,
\begin{align*}
     2{\rm E}\left[\sum_{i=1}^{n_t}\tilde{W}_i\delta_s(X_i) \right]^2\leq 4k_{n_t}\mathbb{E}\left[
            \delta_s^2(X_1)
 \right]
\end{align*}

We have
\begin{align*}
    &\mathbb{E}\left[\left(\hat{Y}(X,\tilde{D}_t) - \left(f_t(X) - f_s(X)\right)\right)^2\right]\\
    &\qquad \leq  C_R\Bigg(|\mathcal{I}_R^c| \sum_{j\in\mathcal{I}_R^c} \left\| \partial_j R \right\|_{\infty}^2 k_{n_t}^{2 \log_2 (1 - p_{nj}^t / 2)} + B_t^2 e^{-n/(2 k_{n_t})}\\
&\qquad \quad + \frac{12 \sigma_t^2 k_{n_t}}{n_t} \frac{8^{d}}{
    \sqrt{
        \prod_{j \in \mathcal{I}_R^c} p_{nj}^t \times \log_2^{|\mathcal{I}_R^c| - 1}(k_{n_t})
    }
}\Bigg) + 4k_{n_t}\mathbb{E}\left[
            \delta_s^2(X_1)
 \right]
\end{align*}

Conclusively, when the event $E$ holds and $n_t$ is large enough, the MSE error 
\begin{align*}
    &\mathbb{E}\left[\left(\hat{Y}(X,(D_s,\tilde{D}_t)) - f_t(X)\right)^2\right]\\
    &\qquad \leq C_R\Bigg(|\mathcal{I}_R^c| \sum_{j\in\mathcal{I}_R^c} \left\| \partial_j R \right\|_{\infty}^2 k_{n_t}^{2 \log_2 (1 - p_{nj}^t / 2)} + M_t^2 e^{-n/(2 k_{n_t})}\\
&\qquad \quad + \frac{12 \sigma_t^2 k_{n_t}}{n_t} \frac{8^{d}}{
    \sqrt{
        \prod_{j \in \mathcal{I}_R^c} p_{nj}^t \times \log_2^{|\mathcal{I}_R^c| - 1}(k_{n_t})
    }
}\Bigg) + (4k_{n_t}+2)\mathbb{E}\left[
            \delta_s^2(X)
 \right] 
\end{align*}

We have also shown that,
\begin{align*}
    \mathbb{E}\left[
            \delta_s^2(X)
 \right]&\leq d \sum_{j=1}^d \|\partial_j f_s\|_{\infty}^2 \, k_{n_s}^{2 \log_2 (1 - 1/2d)}
+
\frac{12 \sigma^2 k_{n_s}}{n_s}
\frac{(8d^{1/2})^d}{\sqrt{\log_2^{\,d - 1}(k_{n_s}) }}
+
M_s^2 \, e^{ - n_s / (2 k_{n_s} ) }
\end{align*}

Denote the loss functions as $L = \left(\hat{Y}(X,(D_s,\tilde{D}_t)) - f_t(X)\right)^2$. Then we have shown that a bound holds on $E[L]$ under the good event $E$.

When the event $E$ does not hold, we want to show $E[L^2]$, which is the $4$th order moment of $\left(\hat{Y}(X,(D_s,\tilde{D}_t)) - f_t(X)\right)$ is bounded.

We first note,
\[
\mathbb{E}\left[\left(\hat{Y}(X,(D_s,\tilde{D}_t)) - f_t(X)\right)^4\right]\leq 8\left(\mathbb{E}\left[\delta_s(X)^4\right] + \mathbb{E}\left[\left(\hat{Y}(X,\tilde{D}_t) - \left(f_t(X) - f_s(X)\right)\right)^4\right]\right)
\]
And for the second term,
\[
\mathbb{E}\left[\left(\hat{Y}(X,\tilde{D}_t) - \left(f_t(X) - f_s(X)\right)\right)^4\right]\leq 8\mathbb{E}\left[\sum_{i=1}^{n_t}\tilde{W}_i\left(R(X_i) + \epsilon_{it} \right)- R(X) \right]^4 + 8\mathbb{E}\left[\sum_{i=1}^{n_t}\tilde{W}_i\delta_s(X_i) \right]^4 
\]
If we can show  $\sum_{i=1}^{n_t}\tilde{W}_i\left(R(X_i) + \epsilon_{it} \right)- R(X)$ and $\delta_s(X) $ are sub-Gaussian and the fourth moment of $\sum_{i=1}^{n_t}\tilde{W}_i\delta_s(X_i)$ is bounded, then $\mathbb{E}\left[\left(\hat{Y}(X,(D_s,\tilde{D}_t)) - f_t(X)\right)^4\right]$ is bounded. We have already show that $\hat{Y}(X, D_s)$ is sub-Gaussian. Since $\delta_s(X) = \hat{Y}(X, D_s) - f_s(X)$ and $f_s(X)$ is bounded, $\delta_s(X)$ is sub-Gaussian. 
Let $Z = \sum_{i=1}^{n_t} \tilde{W}_i \delta_s(X_i)$. We are going to show that $\mathbb{E}[Z^4]$ is bounded.
We decompose $Z= \sum_{i=1}^{n_t} \tilde{W}_i^{3/4} \tilde{W}_i^{1/4}\delta_s(X_i)$.
By Holder inequality with $p=4/3, q=4$,
\begin{equation*}
Z^4 \le \left( \sum_{i=1}^{n_t} \tilde{W}_i \right)^3 \left( \sum_{i=1}^{n_t}  \tilde{W}_i \delta_s^4(X_i) \right) \le  \left( \sum_{i=1}^{n_t}  \tilde{W}_i\delta_s^4(X_i) \right) .
\end{equation*}
The above inequality holds since $\sum_{i=1}^{n_t}  \tilde{W}_i=1$.  Now, taking the expectation on both sides and applying the tower property with respect to $\mathbb{X}_t = \left(X_{1t}, \cdots, X_{n_t t}\right)$:
\begin{equation*}
\mathbb{E}[Z^4] \le \mathbb{E} \left[ \sum_{i=1}^{n_t} \tilde{W}_i \delta_s^4(X_i) \right] = \sum_{i=1}^{n_t} \mathbb{E}\left[\mathbb{E} \left[ \tilde{W}_i \mid \mathbb{X}_t\right]\mathbb{E}[\delta_s^4(X_i) \mid \mathbb{X}_t] \right]
\end{equation*}
The second equality follows from conditional independence between $\tilde{W}_i$ and $\delta_s(X_i)$. This is because $\tilde{W}_i$ depends on $X, \tilde{\Theta}, \mathbb{X}_t$, and $\delta_s(X_i)$ depends on $\mathbb{X}_t$ and $D_s$. Then conditional on $\mathbb{X}_t$, $\delta_s(X_{it})$ is independent of $\tilde{W}_i$. Following the similar statement in lemma \ref{subgaussian}, $\delta_s(X_i)|\mathbb{X}_t$ is sub-Gaussian with the parameter $\Delta$. Notice that we can show that $\Delta$ is a constant depending only on $M_s$ and $\sigma_s^2$ and independent of $\mathbb{X}_t$. Therefore $\mathbb{E}[\delta_s(X_i)^4 \mid \mathbb{X}_t]$ is bounded by a constant $K$ depending only on $M_s$ and $\sigma^2$.
Thus:
\begin{equation*}
\mathbb{E}[Z^4] \le \sum_{i=1}^{n_t} \mathbb{E}[\tilde{W}_i] \cdot K = K\sum_{i=1}^{n_t} \mathbb{E}[\tilde{W}_i] = K
\end{equation*}
Therefore, the fourth moment is bounded by a constant.
\newline
Consequently, we can show the following expression is bounded
\[
\mathbb{E}\left[\left(\hat{Y}(X,(D_s,\tilde{D}_t)) - f_t(X)\right)^4\right]
\]

Therefore, $E[L^2]< \infty$. Then, using Cauchy-Schwarz inequality, we may upper bound the MSE by noting that
\[E[L] \leq  E[L|E] + E[L^2]^{1/2} P(E^C)^{1/2}.
\]
Now for given $\alpha > 0$, for any $0 < \eta<1/2-\alpha$, $\exists \ \tilde{\alpha}>0$ and a constant $\tilde{c}$, s.t.
\begin{align*}
\mathbb{P}\left(E^C\right)\leq O\left(|\mathcal{I}_R| \left[\exp\left(-c_1 \left(\tfrac{\Omega}{|\mathcal{I}_R|}\right)^2 n_t^{1-2(\alpha+\eta)}\right) + n_t \exp(-c_2 n_t^{\eta})\right]\right).
\end{align*}
Therefore $E[L^2]^{1/2}  P(E^C)^{1/2} \to 0$, when $n_t$ is large enough. Combining the results, we get the stated conclusion.

\end{proof}

\subsection{Proof of Corollary 3.9}

\begin{proof}
From Theorem 3.6, we know $\forall \ j \in \mathcal{I}_R^C$, the weights for the residual random forest $p_{nj}^t$ is close to $\frac{\omega^t_j}{\Omega_t}$ with high probability. By assumption 3.5, we also have $\omega_t:=\min_{j\in \mathcal{I}_R^c}\omega_j^t > 0$. From Theorem 3.6, we have, $p_{nj}^t\geq \frac{\omega_j^t}{\Omega} - cn_t^{-\alpha}, \forall \ j \in \mathcal{I}_R^C$ with high probability. That induces $p_{nj}^t\geq \frac{\omega_t}{\Omega} - cn_t^{-\alpha}, \forall \ j \in \mathcal{I}_R^C$ with high probability, since $\omega^t$ is the smallest of the the $\omega^t_j$ s. When $n_t$ is large enough, we can select a $\epsilon_t>0$ independent of $n_t$ and make sure $cn_t^{-\alpha}\leq \epsilon_t$, then
$$
p_{\epsilon_t}:=\frac{\omega_t}{\Omega_t}-\epsilon_t\leq \frac{\omega_t}{\Omega} - cn_t^{-\alpha}\leq p_{nj}^t,\ \forall \ j \in \mathcal{I}_R^C
$$ with probability
\begin{equation}
1-O\left(|\mathcal{I}_R|\left[\exp(-c_1\left(\tfrac{\Omega}{|\mathcal{I}_R|}\right)^2n_t^{1-2(\eta + \alpha)}+n_t \exp(-c_2 n_t^{\eta})\right]\right)
\label{prob1}
\end{equation}
This proves the first part of the Corollary.

 Next, we will optimize the conclusion from Theorem 3.8 with respect to $k_{n_t}$ and $k_{n_s}$

\begin{align*}
        \mathbb{E}\left[(\widehat{Y}(\mathbf{X}) - f_t(\mathbf{X}))^2\right] &\leq C_R\Bigg(|\mathcal{I}_R^c| \sum_{j\in\mathcal{I}_R^c  } \left\| \partial_j R \right\|_{\infty}^2 k_{n_t}^{2 \log_2 (1 - p_{nj}^t / 2)} + M_t^2 e^{-n_t/(2 k_{n_t})}\\
&\qquad \quad + \frac{12 \sigma_t^2 k_{n_t}}{n_t} \frac{8^{d}}{
    \sqrt{
        \prod_{j \in \mathcal{I}_R^c} p_{nj}^t \times \log_2^{|\mathcal{I}_R^c| - 1}(k_{n_t})
    }
}\Bigg)\\ 
&+ C_s(4k_{n_t}+2)\Bigg(d \sum_{j=1}^d \left\| \partial_j f_s \right\|_{\infty}^2 k_{n_s}^{2 \log_2 (1 - p_{nj}^s / 2)} + M_s^2 e^{-n_s/(2 k_{n_s})}.\\
&\quad  + \frac{12 \sigma_s^2 k_{n_s}}{n_s} \frac{8^{d}}{
    \sqrt{
        \prod_{j=1}^d p_{nj}^s \times \log_2^{d - 1}(k_{n_s})
    }
}\Bigg)
    \end{align*}

The first term $$C_R\Bigg(|\mathcal{I}_R^c| \sum_{j\in\mathcal{I}_R^c  } \left\| \partial_j R \right\|_{\infty}^2 k_{n_t}^{2 \log_2 (1 - p_{nj}^t / 2)} + M_t^2 e^{-n_t/(2 k_{n_t})}+ \frac{12 \sigma_t^2 k_{n_t}}{n_t} \frac{8^{d}}{
    \sqrt{
        \prod_{j \in \mathcal{I}_R^c} p_{nj}^t \times \log_2^{|\mathcal{I}_R^c| - 1}(k_{n_t})
    }
}\Bigg)$$
is asymptotically of the order 
$$
O\left(k_{n_t}^{2 \log_2 (1 - p_{\epsilon_t} / 2)} + \frac{k_{n_t}}{n_t\sqrt{\log_2^{|\mathcal{I}_R^c| - 1}n_t}}\right)
$$
where $M_t^2 e^{-n_t/(2 k_{n_t})}$ is ignorable as $k_{n_t} = O\left(n_t^c\right)$ with $c<1$ and then $M_t^2 e^{-n_t/(2 k_{n_t})} = O\left(e^{-n_t^{1-c}}\right)$ which is much smaller than $\frac{k_{n_t}}{n_t\log_2^{|\mathcal{I}_R^c| - 1}(k_{n_t})}$. Also, since we assume $\log_2 k_{n_t} = c_t \log_2n_t$ and $\log_2 k_{n_s} = c_s \log_2n_s$ with $c_t$ and $c_s$ as constants smaller than 1, then we can replace the term $\log_2^{|\mathcal{I}_R^c| - 1}(k_{n_t})$ with $\log_2^{|\mathcal{I}_R^c| - 1}(n_t)$. 
Similarly, the second term is at the order
$$
O\left(k_{n_t}\left(k_{n_s}^{2 \log_2 (1 - 1 / 2d)} + \frac{k_{n_s}}{n_s\sqrt{\log_2^{d - 1}n_s}}\right)\right)
$$
At first, optimizing the second term with respect to $k_{n_s}$, that is, when $k_{n_s} = C_s\left(n_s (\log_2^{d-1} n_s)^{1/2}\right)^{1-r_s}$, the order of the second term would be
$$
O\left(k_{n_t}\left(n_s(\log_2^{d-1}n_s)^{1/2}\right)^{-r_s}\right).
$$
Then we combine the above result and the first term to obtain the asymptotic order of the MSE to be
$$
O\left(k_{n_t}^{2 \log_2 (1 - p_{\epsilon_t} / 2)} + \frac{k_{n_t}}{n_t\sqrt{\log_2^{|\mathcal{I}_R^c| - 1}n_t}} + k_{n_t}\left(n_s(\log_2^{d-1}n_s)^{1/2}\right)^{-r_s}\right).
$$
By optimizing $k_{n_t}$, we get that when $$k_{n_t} = \left(\left(n_s(\log_2^{d-1}n_s)^{1/2}\right)^{-r_s} + C_t\left(n_t\left(\log_2^{|\mathcal{I}_R^c|-1}n_t\right)^{1/2}\right)^{-1}\right)^{r_t-1},$$ the order of MSE would be 
$$
O\left(\left(n_s(\log_2^{d-1}n_s)^{1/2}\right)^{-r_s} + C_t\left(n_t\left(\log_2^{|\mathcal{I}_R^c|-1}n_t\right)^{1/2}\right)^{-1}\right)^{r_t}.
$$

Define 
$$
h(n_s, n_t) = \frac{\left(n_s(\log_2^{d-1}n_s)^{1/2}\right)^{r_s}}{n_t(\log_2^{|\mathcal{I}_R^c|-1}n_t)^{1/2}}.
$$

When $\lim_{n_s, n_t \to \infty}h(n_s,n_t) = 0$, then 
$k_{n_t} = C_t\left(n_s(\log_2^{d-1}n_s)^{1/2}\right)^{r_s(1-r_t)}$, and 
$$
 \mathbb{E}\left[\left(\hat{Y}(X,(D_s,\tilde{D}_t)) - f_t(X)\right)^2\right]\leq C\left(n_s(\log_2^{d-1}n_s)^{1/2}\right)^{-r_sr_t}.
$$

When $\lim_{n_s, n_t \to \infty}h(n_s,n_t) > 0$, then $k_{n_t} = C_t\left(n_t(\log_2^{|\mathcal{I}_R^c|-1}n_t)^{1/2}\right)^{1-r_t}$, and
$$
\mathbb{E}\left[\left(\hat{Y}(X,(D_s,\tilde{D}_t)) - f_t(X)\right)^2\right]\leq C\left(n_t(\log_2^{|\mathcal{I}_R^c|-1}n_t)^{1/2}\right)^{-r_t}.
$$
 
\end{proof}

\section{ Additional Simulation and Real Data Results}

In this section we describe additional results, figures, and tables from Simulation and eICU eeal data analysis sections.

\subsection{Centered random forest additional simulations}

\begin{figure}[h]
    \centering
    \includegraphics[width=0.8\textwidth]{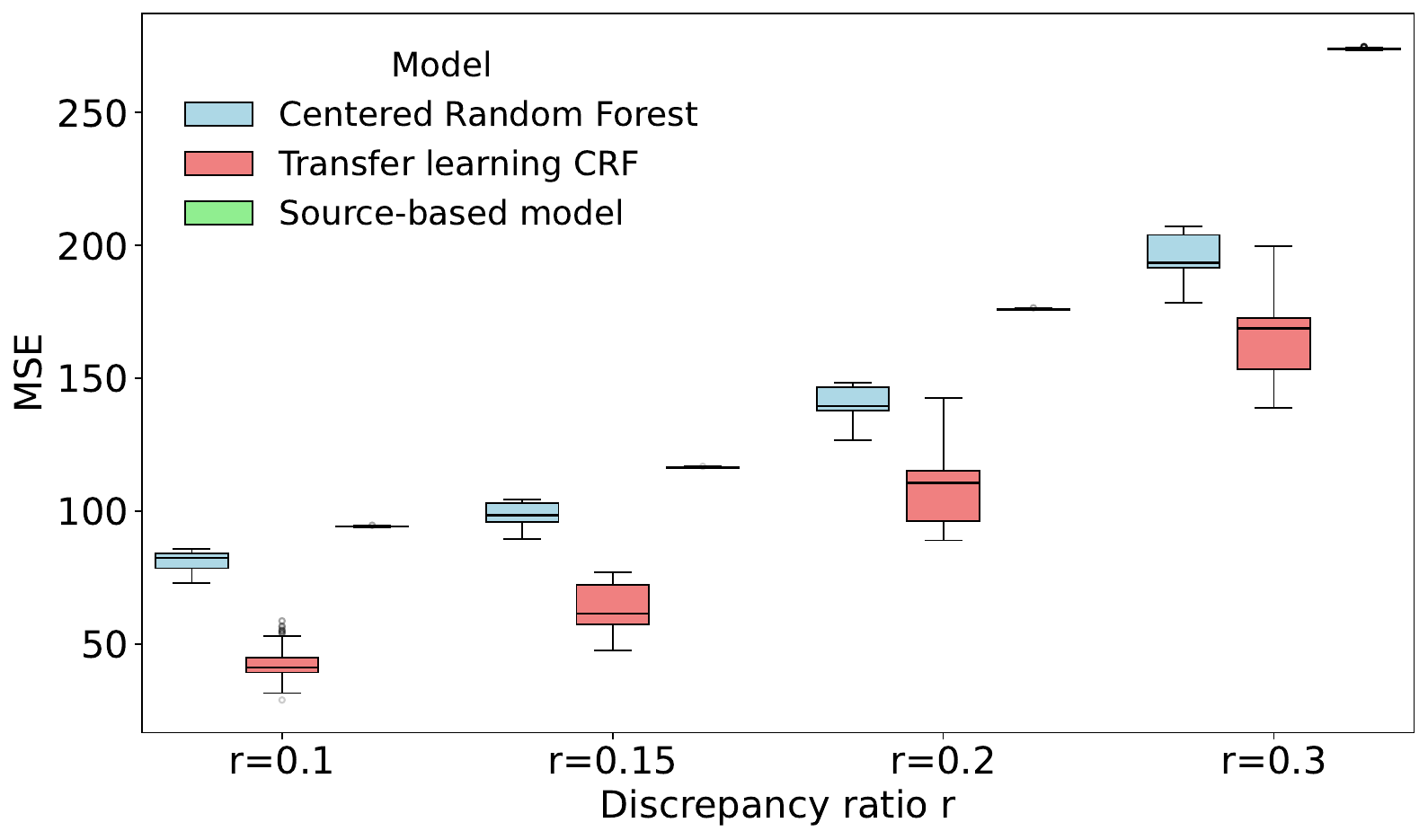}  % Adjust width
    \caption{Mean Square Error performance of TLCRF and CRF trained in source and target data only on test data fixing $n_s = 20000, n_t = 500, n_{test} = 100, d = 50$}
    \label{tlcrf_discrepancy ratio source}
\end{figure}

\begin{figure}[h]
    \centering
    \includegraphics[width=0.8\textwidth]{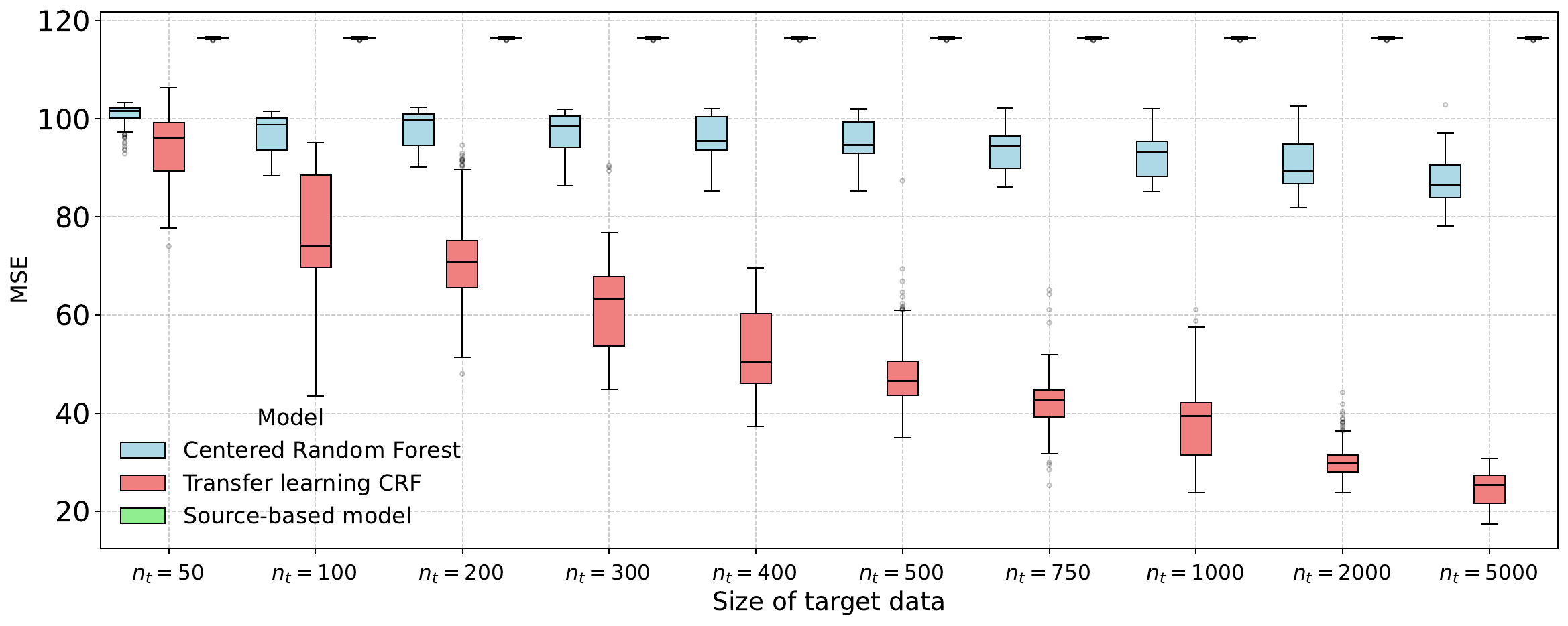}  % Adjust width
    \caption{The result shows the performance of the transfer learning algorithm and centered random forest on test data set. $n_s = 10000, n_{test} = 100, d = 50, r = 0.1$.}
    \label{tlcrf target sample size source}
\end{figure}

Figures \ref{tlcrf_discrepancy ratio source} and \ref{tlcrf target sample size source} present results for CRF simulations that include the results from the source based model. This shows that source based model does not have good predictive performance in the target data due to differences across the domains. The TLCRF method outperforms both target only method and the source only method.

\subsection{eICU data application additional figures}

Here we describe per-hospital results from the eICU data application. For these analysis, every hospital from target 1 and target 2 are considered as the ``target data'', while the source data remains the same (all hospitals with more than 250 beds).

\begin{figure}[H]
    \centering
    \includegraphics[width=0.8\textwidth]{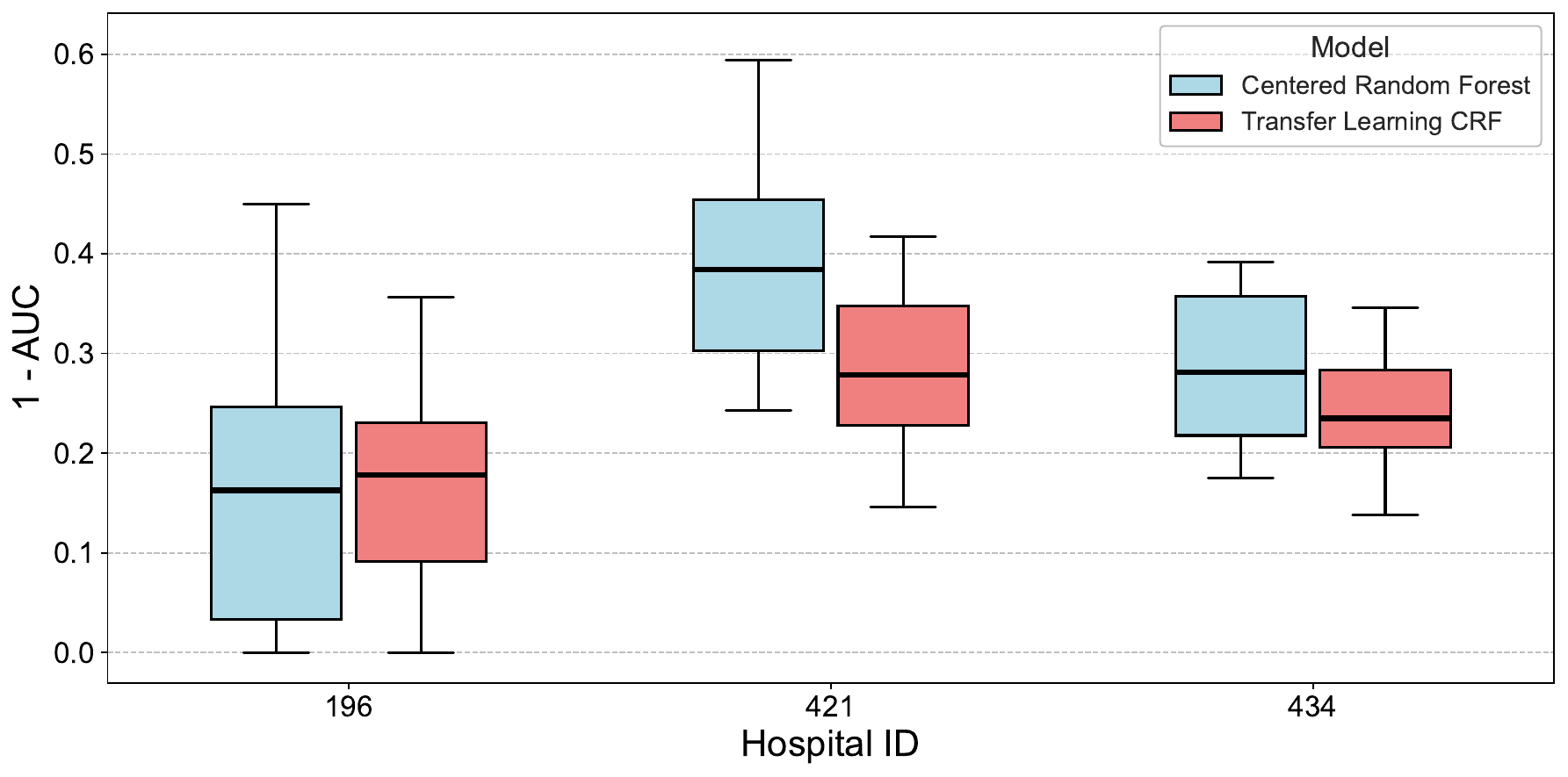}
\caption{Performance of TLCRF vs.\ CRF in terms of 1-AUC for hospitals from Target 1. The hospitals are limited to those with at least 10 deaths. The test size for each hospital is $30\%$ of the number of patients in each hospital. $30\%$ of the training data is used to calculate the distance covariance. Other settings are the same as the previous one.}
\label{fig:eICU_small_hospitals_CRF_1}
\end{figure}

For hospitals in target 1, in order to have sufficient calibration data from target tasks, we further limited our analysis to hospitals with at least 10 deaths. This resulted in only 3 hospitals. The performances of TLCRF and CRF at target are presented in Figure \ref{fig:eICU_small_hospitals_CRF_1}.

\begin{figure}[H]
    \centering
    \includegraphics[width=0.8\textwidth]{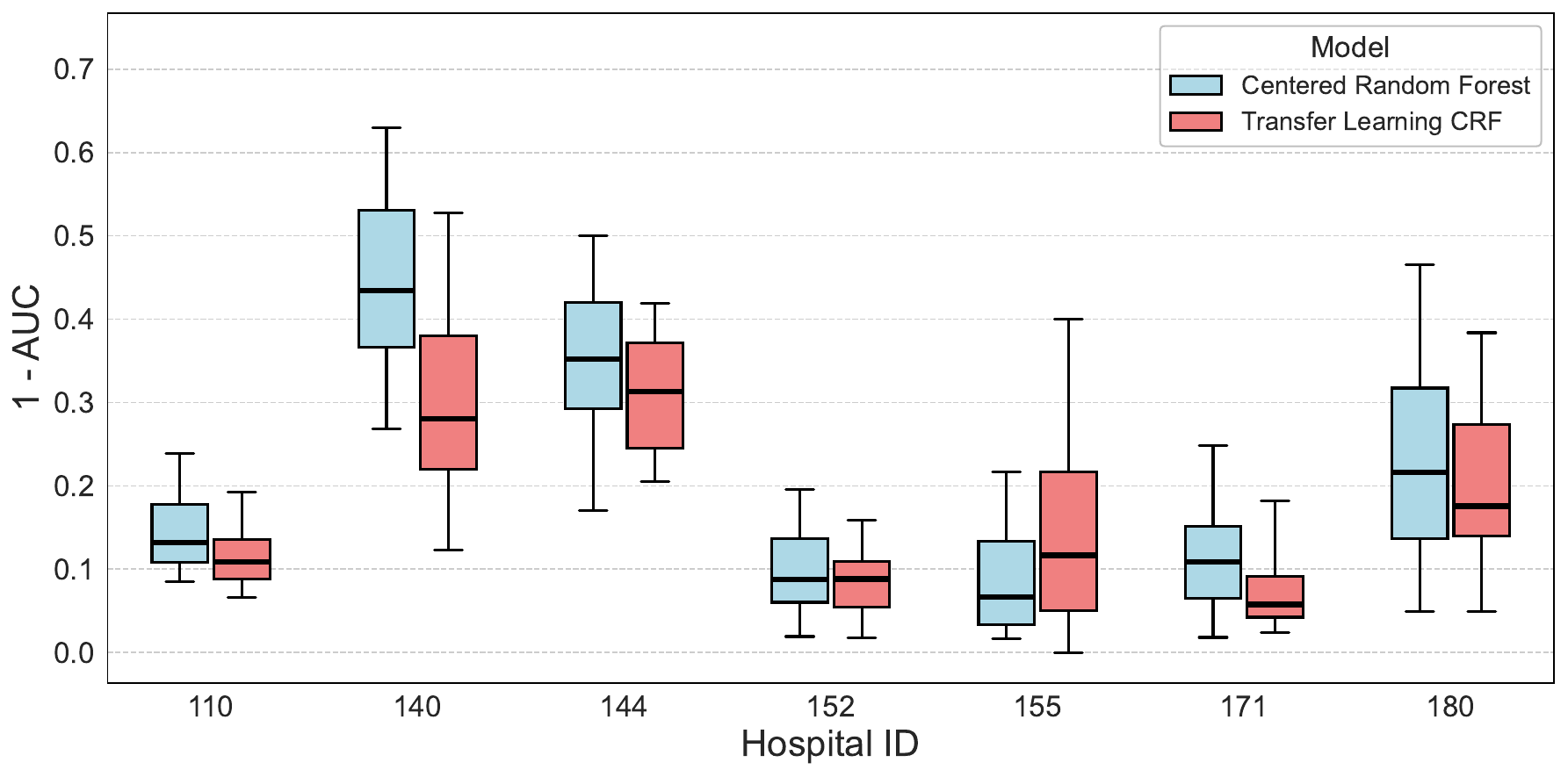}

    \vspace{0.5em} 

    \includegraphics[width=0.8\textwidth]{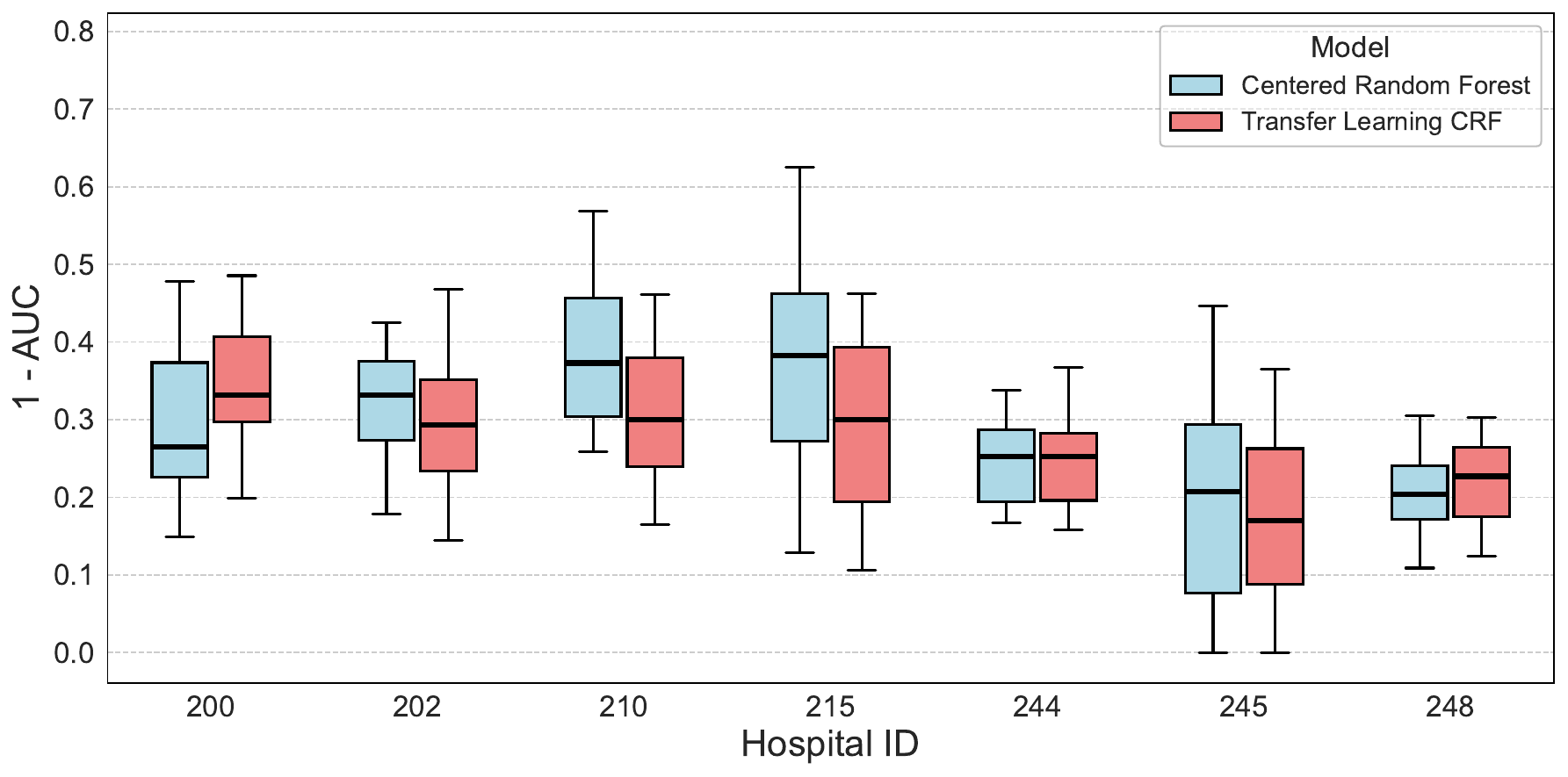}

    \vspace{0.5em}

    \includegraphics[width=0.8\textwidth]{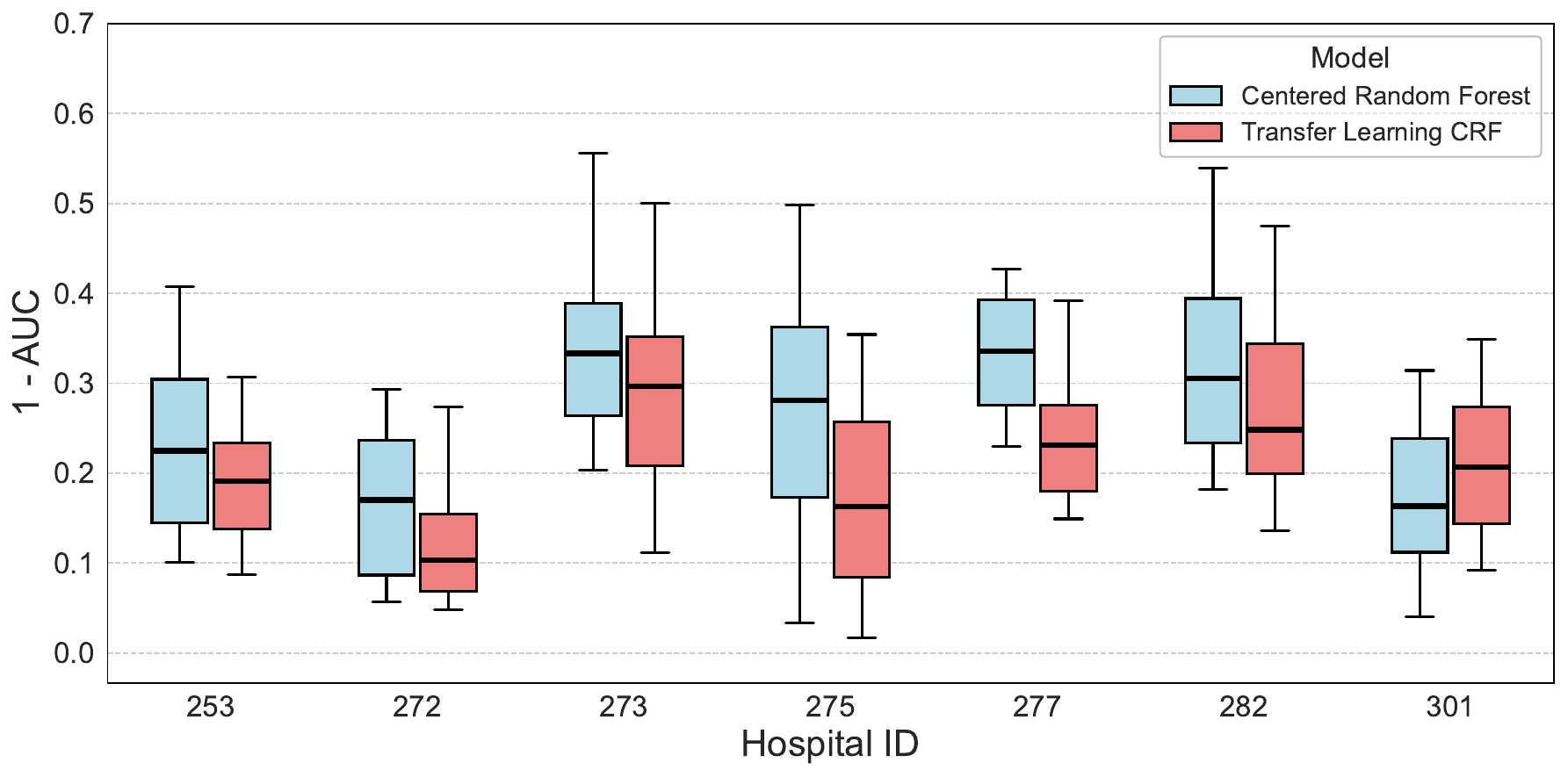}

    \caption{Performance of TLCRF vs.\ CRF in terms of 1-AUC for hospitals from target 2. The settings for these results are the same as those in target 1.}
\label{fig:eICU_small_hospitals_CRF_2_1}
\end{figure}

   \begin{figure}[H]
    \centering

    \includegraphics[width=0.8\textwidth]{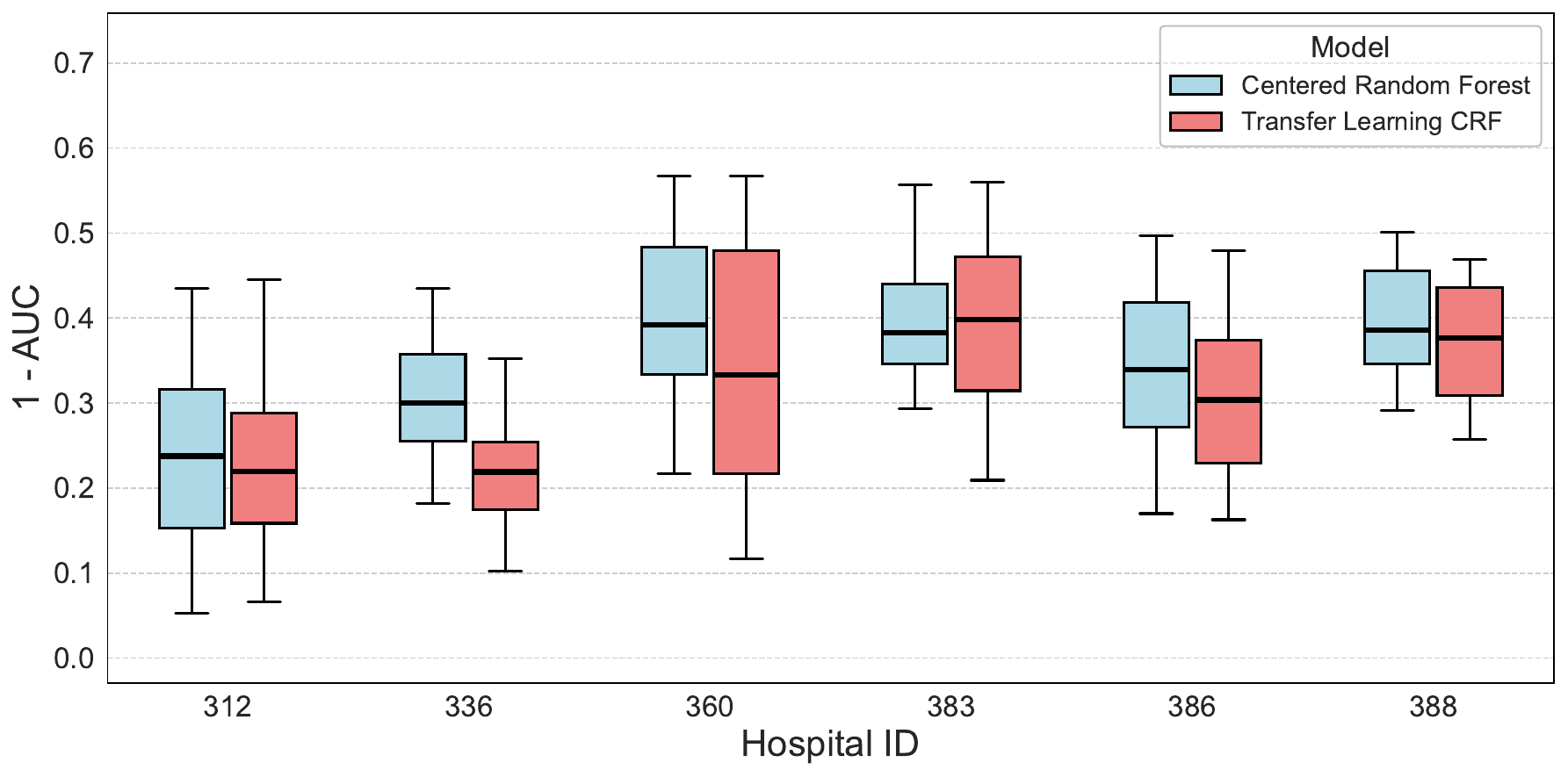}

    \vspace{0.5em}

    \includegraphics[width=0.8\textwidth]{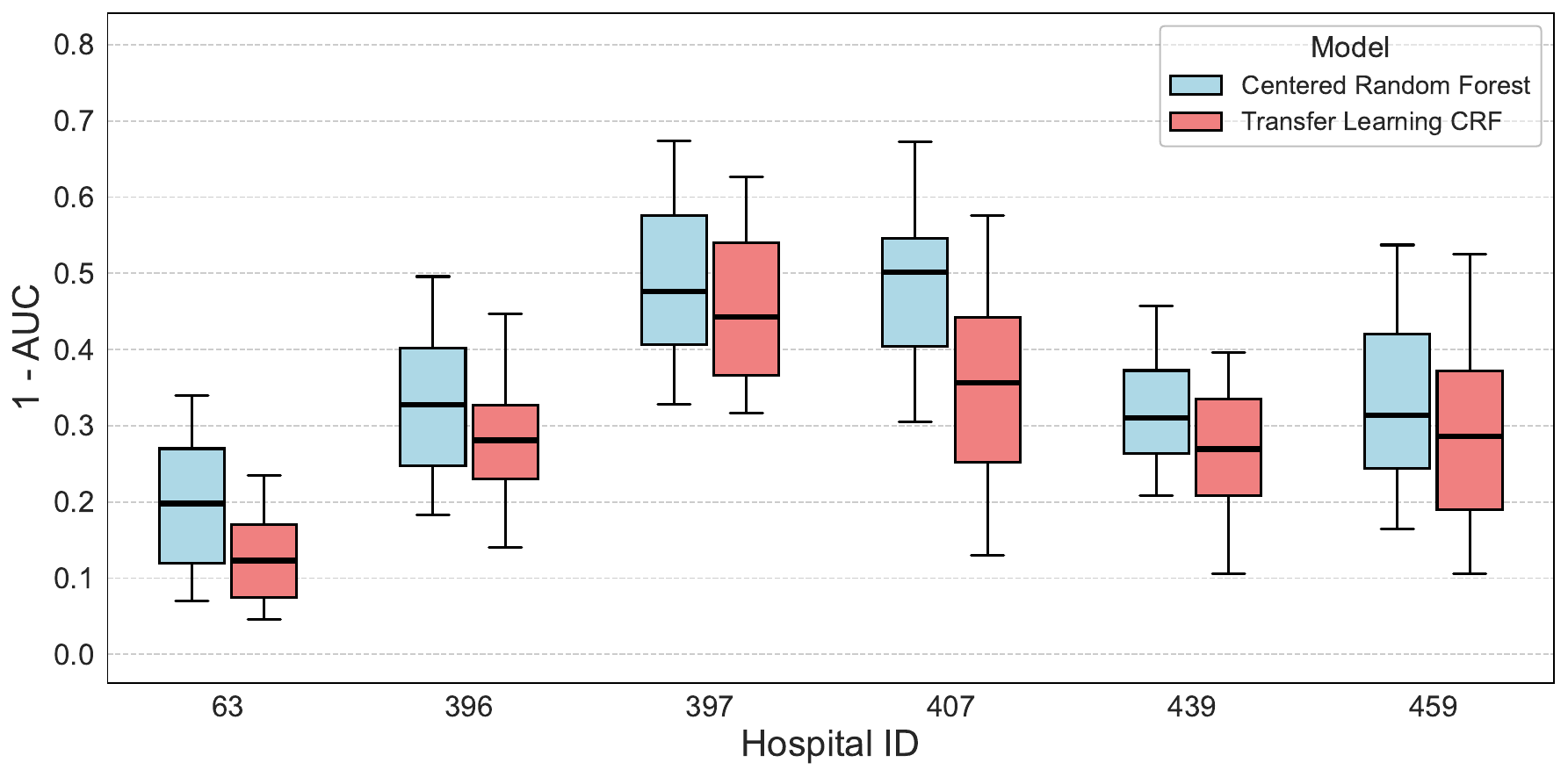}
\caption{Performance of TLCRF vs.\ CRF in terms of 1-AUC for hospitals from target 2. The settings for these results are the same as those in target 1.}
\label{fig:eICU_small_hospitals_CRF_2_2}
\end{figure}

For hospitals in target 2, since they are bigger hospitals, we have sufficient target data available for training in almost all hospitals. The results for the per hospital analysis are presented over multiple figures in Figures \ref{fig:eICU_small_hospitals_CRF_2_1} and  \ref{fig:eICU_small_hospitals_CRF_2_2}. We see that for most hospitals, TLCRF performs either better than or similar to the target only CRF despite the well-known heterogeneity in data across hospitals. 

\subsection{Details of the embedding method}

\begin{figure}[h]
    \centering
    \includegraphics[width=0.9\textwidth]{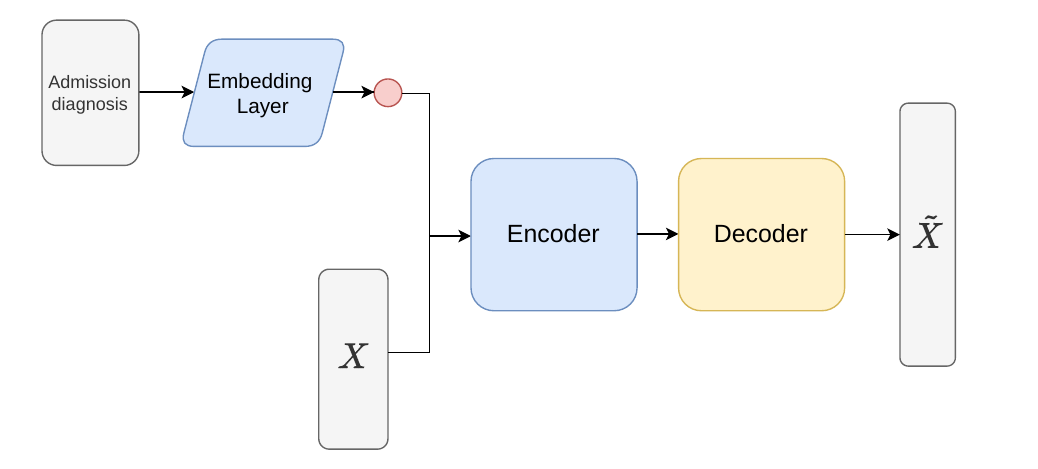}  % Adjust width
    \caption{Illustration of the embedding model}
    \label{embedding}
\end{figure}

The structure of the deep neural network-based embedding is shown in Figure \ref{embedding}. Each category of Admission diagnosis will be labeled and passed through the embedding layer to find the corresponding 8-dimensional vector representation. The embedding layer stores the mapping between labeled categories and 8-dimensional vectors. The 8-dimensional vector is then scaled within 0 and 1 by Sigmoid activation function(red circle), since CRF requires the input to be between 0 and 1. As shown in Figure \ref{embedding}, $X$ represents all the features except Admission diagnosis. The combination of 8-dimensional vector and all the other features are imported into the encoder. Then the decoder reconstructs the output of the encoder into the vectors with the same dimensions as $X$. The new vectors denoted as $\tilde{X}$ contain the information of $X$ and the admission diagnosis. The loss function of the system is the mean squared error between $\tilde{X}$ and $X$, along with variance across all elements in the embedding matrix. Each row of the embedding matrix is an 8-dimensional vector corresponding to each category of Admission diagnosis.

%%  MR numbers will be added by VTeX.                      %%
%%                                                         %%
%%  Use \cite{...} to cite references in text.             %%
%%                                                         %%
%%%%%%%%%%%%%%%%%%%%%%%%%%%%%%%%%%%%%%%%%%%%%%%%%%%%%%%%%%%%%

%% if your bibliography is in bibtex format, uncomment commands:
\bibliographystyle{imsart-nameyear} % Style BST file (imsart-number.bst or imsart-nameyear.bst)
\bibliography{references}       % Bibliography file (usually '*.bib')

%% or include bibliography directly:
%%\begin{thebibliography}{4}
%%

%%\end{thebibliography}

\end{document}